\documentclass{article}

 \usepackage[main, final]{neurips_2026}
\usepackage[utf8]{inputenc} % allow utf-8 input
\usepackage[T1]{fontenc}    % use 8-bit T1 fonts
\usepackage{hyperref}       % hyperlinks
\usepackage{url}            % simple URL typesetting
\usepackage{booktabs}       % professional-quality tables
\usepackage{amsfonts}       % blackboard math symbols
\usepackage{nicefrac}       % compact symbols for 1/2, etc.
\usepackage{microtype}      % microtypography
\usepackage{xcolor}         % colors

\usepackage{booktabs}
\usepackage{graphicx}
\usepackage{multirow}
\usepackage{epstopdf}

\usepackage[mathscr]{eucal}
\usepackage[cmex10]{amsmath}
\usepackage{epsfig,epsf,psfrag}
\usepackage{amssymb,amsmath,amsthm,amsfonts,latexsym}
\usepackage{amsmath,graphicx,bm,xcolor,url,overpic}
\usepackage{array}%array and tabular environments
\usepackage{verbatim}
\usepackage{bm}
\usepackage{algorithmic}
\usepackage{algorithm}
\usepackage{enumitem}
\usepackage{verbatim}
\usepackage{textcomp}
\usepackage{mathrsfs}
\usepackage{hyperref}
\usepackage{epstopdf}
\usepackage{subfigure}
\usepackage{acronym}
\usepackage{diagbox}
\usepackage{tikz}
\usetikzlibrary{graphs, positioning, quotes, shapes.geometric}
\usepackage{thmtools, thm-restate}
\newcommand{\openone}{\leavevmode\hbox{\small1\normalsize\kern-.33em1}} 

\catcode`~=11 \def\UrlSpecials{\do\~{\kern -.15em\lower .7ex\hbox{~}\kern .04em}} \catcode`~=13 

\allowdisplaybreaks[4]

\newcommand{\calF}{\mathcal{F}}

\newcommand{\calH}{\mathcal{H}}

\newcommand{\calN}{\mathcal{N}}

\newcommand{\calQ}{\mathcal{Q}}

\newcommand{\calS}{\mathcal{S}}

\newcommand{\calX}{\mathcal{X}}

\newcommand{\rmd}{\mathrm{d}}

\newcommand{\rmH}{\mathrm{H}}

\newcommand{\bbE}{\mathbb{E}}
\newcommand{\bbF}{\mathbb{F}}

\newcommand{\bbN}{\mathbb{N}}

\newcommand{\bbP}{\mathbb{P}}

\newcommand{\bbR}{\mathbb{R}}

\newcommand{\bbZ}{\mathbb{Z}}

\DeclareMathAlphabet{\mathbsf}{OT1}{cmss}{bx}{n}
\DeclareMathAlphabet{\mathssf}{OT1}{cmss}{m}{sl}% slanted sans serif

\DeclareSymbolFont{bsfletters}{OT1}{cmss}{bx}{n}  
\DeclareSymbolFont{ssfletters}{OT1}{cmss}{m}{n}
\DeclareMathSymbol{\bsfGamma}{0}{bsfletters}{'000}
\DeclareMathSymbol{\ssfGamma}{0}{ssfletters}{'000}
\DeclareMathSymbol{\bsfDelta}{0}{bsfletters}{'001}
\DeclareMathSymbol{\ssfDelta}{0}{ssfletters}{'001}
\DeclareMathSymbol{\bsfTheta}{0}{bsfletters}{'002}
\DeclareMathSymbol{\ssfTheta}{0}{ssfletters}{'002}
\DeclareMathSymbol{\bsfLambda}{0}{bsfletters}{'003}
\DeclareMathSymbol{\ssfLambda}{0}{ssfletters}{'003}
\DeclareMathSymbol{\bsfXi}{0}{bsfletters}{'004}
\DeclareMathSymbol{\ssfXi}{0}{ssfletters}{'004}
\DeclareMathSymbol{\bsfPi}{0}{bsfletters}{'005}
\DeclareMathSymbol{\ssfPi}{0}{ssfletters}{'005}
\DeclareMathSymbol{\bsfSigma}{0}{bsfletters}{'006}
\DeclareMathSymbol{\ssfSigma}{0}{ssfletters}{'006}
\DeclareMathSymbol{\bsfUpsilon}{0}{bsfletters}{'007}
\DeclareMathSymbol{\ssfUpsilon}{0}{ssfletters}{'007}
\DeclareMathSymbol{\bsfPhi}{0}{bsfletters}{'010}
\DeclareMathSymbol{\ssfPhi}{0}{ssfletters}{'010}
\DeclareMathSymbol{\bsfPsi}{0}{bsfletters}{'011}
\DeclareMathSymbol{\ssfPsi}{0}{ssfletters}{'011}
\DeclareMathSymbol{\bsfOmega}{0}{bsfletters}{'012}
\DeclareMathSymbol{\ssfOmega}{0}{ssfletters}{'012}

\newcommand{\tilG}{\tilde{G}}

\newcommand{\tilT}{\tilde{T}}

\newcommand{\hmu}{\hat{\mu}}

\DeclareMathOperator*{\argmax}{argmax}
\DeclareMathOperator*{\argmin}{argmin}

\def\##1\#{\begin{align}#1\end{align}}
\def\$#1\${\begin{align*}#1\end{align*}}

\def\BibTeX{{\rm B\kern-.05em{\sc i\kern-.025em b}\kern-.08em
    T\kern-.1667em\lower.7ex\hbox{E}\kern-.125emX}}
\usepackage{booktabs,multirow,colortbl}
\definecolor{sp2ecrow}{RGB}{214,227,251}
\usepackage{mathtools}
\usepackage{bbm}
\usepackage{acronym}
\acrodef{mab}[MAB]{Multi-Armed Bandit}
\acrodef{bai}[BAI]{Best Arm Identification}
\acrodef{rm}[RM]{Regret Minimization}
\acrodef{llm}[LLM]{Large Language Model}
\acrodef{vlm}[VLM]{Vision Language Model}

\theoremstyle{plain}
\newtheorem{theorem}{Theorem}[section]
\newtheorem{proposition}[theorem]{Proposition}
\newtheorem{lemma}[theorem]{Lemma}

\theoremstyle{definition}

\newtheorem{assumption}[theorem]{Assumption}
\theoremstyle{remark}

\newcommand{\banditspec}{\textsc{BanditSpec}}
\newcommand{\prompt}{\textrm{pt}}
\newcommand{\EOS}{\textrm{EOS}}
\newcommand{\return}{\textbf{return}}
\newcommand{\ST}{\tau}
\newcommand{\WT}{\textrm{WT}}
\newcommand{\alg}{\textrm{alg}}
\newcommand{\concat}{\textrm{concat}}

\newcommand{\WTReg}{\textrm{WTReg}}

\newcommand{\len}{\textrm{len}}

\newcommand{\bcalS}{\bar{\calS}}
\newcommand{\spsqec}{\textsc{SP}$^2$\textsc{ec}}
\newcommand{\hnu}{\hat{\nu}}
\newcommand{\ucb}{\textrm{UCB}}
\newcommand{\Good}{\textrm{Good}}
\newcommand{\const}{{const}}

\newcommand{\ourmethod}{\textsc{SP}$^2$\textsc{ec}}
\newcommand{\ucbspec}{\textsc{UCBSpec}}
\newcommand{\swift}{\textsc{SWIFT}}
\newcommand{\eaglethree}{\textsc{EAGLE3}}

\title{\spsqec{}: Adaptive Self-Speculative Decoding for Vision-Language Models}

\author{%
  \textbf{Yuqi~Huang}$^{1*}$ \quad
  \textbf{Xingyao~Li}$^{1*}$ \quad
  \textbf{Yunlong~Hou}$^{1*\dagger}$ \quad
  \textbf{Fengzhuo~Zhang}$^{3*}$ \\
  \textbf{Jiachun~Pan}$^{2}$ \quad
  \textbf{Vincent~Y.~F.~Tan}$^{1,2}$ \\
  $^{1}$Department of Mathematics, National University of Singapore, Singapore \\
  $^{2}$Department of Electrical and Computer Engineering, National University of Singapore, Singapore \\
  $^{3}$Yale University, Center for Algorithms, Data, and Market Design, USA
}

\begin{document}

\maketitle

\begin{abstract}
  Speculative Decoding (SD) has demonstrated its efficacy in accelerating large language model inference, but for Vision-Language Models (VLMs), the diverse input modalities make it difficult for a small trained drafter to achieve consistently high throughput. Self-speculative decoding reuses parts of the target model as a drafter which retains the multi-modal capability of the VLMs, yet its efficiency depends on choosing a suitable draft configuration for each prompt. We propose \textsc{Single-Peak SPeculative decoding} (\spsqec{}), a training-free bandit algorithm that adapts the draft configuration during generation. SP$^2$ec exploits the single-peak throughput pattern observed in existing self-speculative methods, including \swift{}, \textsc{KnapSpec}, \textsc{SpecVLM}, and our simple \textsc{cosSim}.
Our algorithmic design explicitly minimizes a novel notion of \emph{wall-time regret} for each given query, corresponding to the empirical wall-time latency. Theoretical guarantees on the wall-time regret show that \textsc{SP}$^2$\textsc{ec} can exploit the optimal draft configuration, with negligible exploration cost. Extensive experiments on Qwen2.5-VL-7B-Instruct and LLaVA-NeXT-Mistral-7B across visual and textual tasks show \emph{consistent improvements} of up to 11\% over the original self-speculative methods, demonstrating the wide applicability of the proposed \textsc{SP}$^2$\textsc{ec} algorithm.
\end{abstract}

\begin{NoHyper}
\begingroup
\let\thefootnote\relax
\footnotetext{$^{*}$Equal contribution. $^{\dagger}$Project lead.}
\endgroup
\end{NoHyper}

\section{Introduction}

Speculative decoding has been widely used to accelerate \ac{llm} inference~\citep{leviathan2023fast,chen2023accelerating}. However, its efficiency remains limited for \acp{vlm}, where diverse input modalities and long multimodal contexts make it difficult for a separately trained small draft model to consistently match the target model~\citep{kang2025vispec,yang2026longspec}. Self-speculative decoding provides a promising alternative by reusing part of the target model itself as the drafter, thereby maintaining stronger draft-target alignment for any context~\citep{elhoushi2024layerskip,xia2025swift,sadhukhan2025magicdec}. Motivated by its success in \acp{llm}, we study self-speculative decoding for \acp{vlm}. 
% In particular, 
% we focus on layer-skipping methods, which construct the draft model by skipping a subset of layers during the forward pass~\citep{xia2025swift}.
We focus on two commonly used self-speculative decoding methods as an illustration of our approach: layer-skipping methods which construct the draft model by skipping a subset of layers during the forward pass~\citep{xia2025swift}, and token-selection methods where the draft model uses a small subset of the input tokens~\citep{ji2025specvlm}.

Despite their advantages, these methods are often limited by the need to choose suitable component configurations for \emph{each prompt}, e.g., the retained layers and retained tokens.
%For instance, given a prompt, for layer-skipping methods~\citep{xia2025swift,cha2026knapspec}, the decoding throughput depends critically on which layers are retained in the draft model;
%for token-selection methods~\citep{ji2025specvlm}, more retained visual context tokens used by the draft model require more time for the drafting process.
%In these methods, a higher acceptance rate comes at the cost of increased drafting time, and there is an optimal configuration that trades off between them and achieves the minimum latency.
Specifically, throughput depends critically on the choice of retained layers in layer-skipping methods~\citep{xia2025swift,cha2026knapspec}, and on the number of retained visual tokens in token-selection methods~\citep{ji2025specvlm}. 
Retaining more components can improve acceptance but also increases drafting time, hence there is an optimal
configuration that trades off between them and achieves the minimum latency.
Crucially, this optimal configuration \emph{varies across prompts}, e.g., a configuration that achieves lower latency for one prompt may be suboptimal for another. 
This motivates the central question of this work:
\textit{Can we design a training-free and prompt-adaptive method to efficiently select the component configurations of the draft model (e.g., retained layers or tokens) and minimize wall-time latency?}

We formulate this task as a single-peak bandit problem, where the algorithm adaptively chooses the retained components with empirically accepted tokens and time costs as feedback, with the ultimate goal of minimizing the wall-time latency.
Our contributions are threefold:
\begin{itemize}[leftmargin=*]
    \item We first propose a fine-grained, cost-aware formulation of self-speculative decoding. In this framework, each retained component configuration is viewed as an arm with both reward and wall-time cost. We introduce wall-time regret as the primary metric, which measures the additional latency relative to the best fixed configuration. Based on the observed diminishing returns of the mean accepted token length and the linearity of the time cost, we establish a single-peak structure over the ordered draft model configuration sequence.
    \item We propose \spsqec{}, an algorithm that minimizes wall-time latency on a per-prompt basis. We derive theoretical guarantees on its wall-time regret and identify an instance-dependent hardness parameter. Thanks to the single-peak structure, this parameter shows that the leading term in the regret depends \emph{only} on the neighborhood of the optimal configuration, rather than on the entire configuration set.
    \item We conduct extensive experiments with Qwen2.5-VL-7B-Instruct~\citep{qwen2.5vltechnicalreport} and LLaVA-NeXT-Mistral-7B~\citep{liu2024llavanext} across various self-speculative decoding methods on both visual and textual tasks. \spsqec{} \emph{consistently} improves throughput over the original self-speculative methods, with gains of up to $11\%$.
    The results demonstrate the empirical effectiveness of \spsqec{} compared with the vanilla self-speculative decoding method and adaptive configuration selection baseline.
    % We further empirically validate the robustness of our assumptions on the time cost.
\end{itemize}

% Previous adaptive speculative decoding algorithms aim to minimize some proxies of the wall time, e.g., 
% \citet{hou2025banditspec,liu2026notabandit} minimize the number of speculative decoding rounds;
% \citet{kim2026multidrafter} constructs some reward function and attempts to maximize the number of accepted tokens;
% \citet{zhang2026learning} adopts the accepted token per second as the reward signal in policy training;
% \citet{sridhar2025tapout} utilizes the mixture of the normalized accepted length and acceptance rate as the reward.
% While these performance metrics are positively aligned with the ultimate goal--minimizing the latency for the whole decoding process--none of them directly handles the ground-truth metric, the wall time, and provides interpretable theoretical guarantees on the wall time latency.

\begin{figure}
    \centering
    \includegraphics[width=\linewidth]{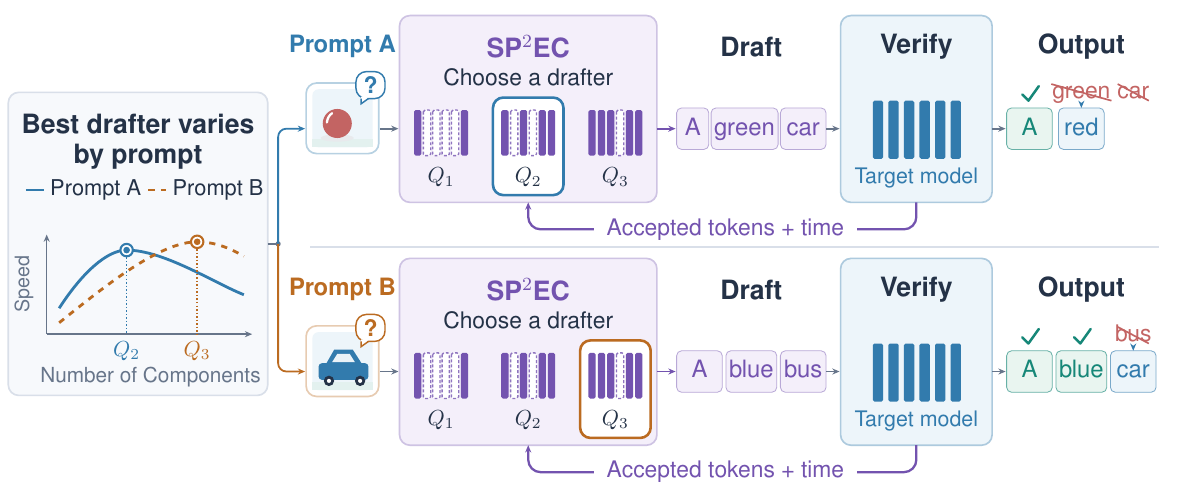}
    % \vspace{-2em}
    \caption{\textbf{Prompt-level adaptive drafter selection in \spsqec{}.} Each arm corresponds to a drafter with different layer-skipping configuration and wall-time cost. \spsqec{} uses accepted-token and time feedback from the target model to exploit the single-peak throughput structure and adaptively select a drafter during generation. Different prompts can favor different arms, motivating per-query adaptation for VLM speculative decoding.}
    \label{fig:pipeline}
\end{figure}
% \section{Related Works}\label{app:relatedworks}

\textbf{Related Works.} Speculative decoding accelerates autoregressive generation by drafting candidate continuations and verifying them in parallel with the target model. Early methods use lightweight draft models~\citep{leviathan2023fast,chen2023accelerating}, blockwise or seq2seq parallelism~\citep{stern2018blockwise,xia2023speculative}, improved drafter alignment or distillation~\citep{kim2023speculative,liu2023online,zhou2023distillspec}, and retrieval or n-gram proposals~\citep{yang2023inference,he2024rest}. Other works enrich verification with trees, cascades, multi-heads, or parallel drafters~\citep{miao2023SpecInferAG,cai2024medusa,li2024eagle,li2024eagle2,li2025eagle3,ankner2024hydra,zhao2024ouroboros,xiao2024parallelspec}. While effective, these methods often rely on auxiliary drafter quality, extra training, or prompt--drafter alignment. Self-speculative decoding instead reuses the target model through layer skipping or early exit~\citep{Zhang_2024selfspec,elhoushi2024layerskip}. Closely related works adapt skipped layers or modules under a fixed budget~\citep{xia2025swift,chen2025clasp,cha2026knapspec}. Our work is orthogonal: we allocate the skipping budget per prompt, rather than selecting layers or modules under a fixed budget. More related works are in Appendix~\ref{app:related}.

\section{Preliminaries}\label{sec:prelim}
\textbf{Speculative decoding} adopts a draft model $Q$ to accelerate inference from a target model $P$. At round $t$, given prefix $\prompt_{t-1}\in\calX^*$ and maximum draft length $L$, where $\calX^*$ denotes the set of finite sequences over alphabet $\calX$, the draft model samples $y_i\sim Q(\cdot\mid \prompt_{t-1},y_{<i})$ for $i=1,\ldots,L$. The target model verifies these tokens in parallel by computing $P_i(\cdot)=P(\cdot\mid \prompt_{t-1},y_{<i})$ and $Q_i(\cdot)=Q(\cdot\mid \prompt_{t-1},y_{<i})$. Each token is accepted with probability $\alpha_i(y_i)=\min\{1,P_i(y_i)/Q_i(y_i)\}$. Let \begin{equation}
    A_t=\max\{a\in\{0,\ldots,L\}:u_i\leq \alpha_i(y_i),\forall i\leq a\}
\end{equation} be the number of accepted draft tokens, where $u_i\sim\mathrm{Unif}(0,1)$. If $A_t=L$, one bonus token is sampled from $P(\cdot\mid \prompt_{t-1},y_{1:L})$; otherwise, the first rejected token is resampled from \begin{equation}
    r_{A_t+1}(v)=\frac{[P_{A_t+1}(v)-Q_{A_t+1}(v)]_+}{\sum_{v'\in\calX}[P_{A_t+1}(v')-Q_{A_t+1}(v')]_+}. 
\end{equation}The round outputs $x_t=(y_1,\ldots,y_{A_t},z)$ and updates $\prompt_t=\concat(\prompt_{t-1},x_t)$. Its acceptance length is $|x_t|=A_t+1$, with \begin{equation}\mathbb E\big[|x_t|\mid \prompt_{t-1}\big]=1+\sum_{i=1}^L\bbP(A_t\geq i\mid \prompt_{t-1}).\end{equation} Let $c_t$ denote the wall-time cost of this round. Then $\WT_t=\WT_{t-1}+c_t$. Decoding stops at $\ST(\prompt,\Lambda)=\min\{t:\EOS\in x_t\}$, with total wall time $\WT_{\ST(\prompt,\Lambda)}$.

% \textbf{Multi-Armed Bandit} is a standard online decision-making paradigm. There are $K$ arms indexed by $k\in[K]$. At each round $t=1,\ldots,T$, the learner selects an arm $I_t\in[K]$ and observes only the reward $R_t(I_t)$ of the selected arm. In the stochastic setting, each arm $k$ has an unknown reward distribution with mean $\mu_k=\mathbb E[R_t(k)]$. The learner aims to balance the exploration of uncertain arms and the exploitation of empirically good arms. Let $k^\star\in\arg\max_{k\in[K]}\mu_k$ be an optimal arm. The   expected regret is $\mathrm{Reg}(T)=T\mu_{k^\star}-\mathbb E[\sum_{t=1}^T R_t(I_t)]$. Equivalently, for a cost-minimization problem with observed cost $\ell_t(k)$ and mean cost $c_k=\mathbb E[\ell_t(k)]$, the regret is $\mathrm{Reg}(T)=\mathbb E[\sum_{t=1}^T \ell_t(I_t)]-T\min_{k\in[K]}c_k$. In our setting, each arm corresponds to a speculative-decoding configuration, such as a draft length, a skipped-layer pattern, or retained prefix tokens. Pulling an arm means applying that configuration to the current prompt and observing its realized wall time, acceptance length, or throughput. The bandit formulation therefore captures adaptive configuration selection under partial feedback, since only the performance of the chosen configuration is observed for each prompt.

\textbf{Multi-Armed Bandit} is a standard online decision-making paradigm. There are $K$ arms indexed by $k\in[K]$. At each round $t=1,\ldots,T$, the learner selects an arm $I_t\in[K]$ and observes only the reward $R_t(I_t)$ of the selected arm. In the stochastic setting, each arm $k$ has an unknown reward distribution with mean $\mu_k=\mathbb E[R_t(k)]$. 
Let $k^\star\in\arg\max_{k\in[K]}\mu_k$ be an optimal arm. 
The learner aims to minimize the expected regret $T\mu_{k^\star}-\mathbb E[\sum_{t=1}^T R_t(I_t)]$, which requires a balance between the exploration of uncertain arms and the exploitation of empirically good arms. 
% Equivalently, for a cost-minimization problem with observed cost $\ell_t(k)$ and mean cost $c_k=\mathbb E[\ell_t(k)]$, the regret is $\mathrm{Reg}(T)=\mathbb E[\sum_{t=1}^T \ell_t(I_t)]-T\min_{k\in[K]}c_k$. 
In our setting, each arm corresponds to a speculative-decoding configuration, such as a draft length, a skipped-layer pattern, or retained prefix tokens. Pulling an arm means applying that configuration to the current prompt and observing its realized wall time, acceptance length, or throughput. The bandit formulation therefore captures adaptive configuration selection under partial feedback, since only the performance of the chosen configuration is observed for each prompt.
\section{Fine-Grained Formulation of Adaptive Self-Speculative Decoding} \label{sec:problem_formulation}

In this section, we focus on the modelling of adaptive self-speculative decoding. 
We construct the draft model by selecting a subset of the components that constitute the target model.
We formulate this component selection problem as a single-peak stochastic bandit problem where each draft model corresponds to an arm in the bandit formulation, with the target of minimizing the wall time latency of the decoding process.
% To provide concrete examples, we focus on the layer-skipping methods and the token-selection methods.

\begin{algorithm}[t]
\textbf{Inputs:} bandit configuration $\Lambda =(P,\calQ=\{Q_S\}_{S\in\calS},L)$, arm selection algorithm $\alg$, initial prompt $\prompt_0 = \prompt\in\calX^*$.
% \textbf{Procedures:}
    \caption{\textsc{Speculative Decoding with Cost-Aware Bandits} }\label{algo:cost_aware_banditspec}    
    \begin{algorithmic}[1]
    \STATE $t=0, \calH_0=\emptyset,S_0=1, x_{S_0,0} = \emptyset,\WT_0=0$.
    \WHILE{$\EOS\notin x_{S_t,t}$} 
    \STATE $t = t + 1$.
    \STATE Select draft index $S_t = \alg(\calH_{t-1})$.
    \label{lne:arm_select}
    \STATE Conduct one SD round with $(\prompt_{t-1},P,Q_{S_t},L)$, and obtain accepted tokens $x_{S_t,t}$ with elapsed time $c_{S_t}$.\label{lne:spec}
    \STATE $\prompt_{t} = \concat(\prompt_{t-1},x_{S_t,t})$.
    \label{lne:pt_update}
    \STATE $\WT_{t} = \WT_{t-1}+c_{S_t}$.
    \STATE $\calH_{t} = \concat(\calH_{t-1},(S_t,x_{S_t,t}))$.\label{lne:his_update}
    \ENDWHILE
    \STATE \return{}  $\ST(\alg,\prompt,\Lambda) = t,\; \prompt_{\ST(\alg,\prompt,\Lambda)} = \prompt_{t}, \WT_{\ST(\alg,\prompt,\Lambda)} = \WT_t$.
    \end{algorithmic}
\end{algorithm}

%% framework
We adopt the \banditspec{} framework from \citet{hou2025banditspec} to adaptively select the retained components, and we use the recorded Wall Time (WT), \begin{equation}
    \WT_{\ST(\alg,\prompt,\Lambda)}=\sum_{t=1}^{\ST(\alg,\prompt,\Lambda)}c_{S_t},
\end{equation}as the golden metric to measure the performance (see Algorithm~\ref{algo:cost_aware_banditspec}).
Let $S$ denote the set of retained components and let $\calS$ be the set of all candidate $S$.
The bandit configuration $\Lambda =(P,\calQ=\{Q_S\}_{S\in\calS},L)$ consists of the target model $P$, the draft model $Q_S$ indexed by $S\in\calS$ and speculation length $L$.
At each speculative decoding round $t$, a candidate component selection algorithm $\alg$ will choose the retained components $S_t\in\calS$ (Line 4), which will be used as the draft model to conduct one round of speculative decoding (Line 5). Both the accepted tokens $x_{S_t,t}$ (Line 6) and the elapsed wall time $c_{S_t}$ (Line 7) are then recorded. In the end, the total speculation rounds $\ST(\alg,\prompt,\Lambda)$, the generated sequence $\prompt_{\ST(\alg,\prompt,\Lambda)}$ and the total wall time $\WT_{\ST(\alg,\prompt,\Lambda)}$ are returned.
Without loss of generality, we assume the $\EOS$ token is accepted as the last accepted token in one speculative decoding round.

%% target
\textbf{The Golden Metric: Wall-Time Regret.} Given the prompt and bandit configuration, denote $\alg^*=\argmin_{\alg}\bbE[\WT_{\ST(\alg,\prompt,\Lambda)}]$ as the \textit{optimal component selection algorithm} that minimizes the expected decoding wall time, where the expectation is taken with respect to the randomness in the algorithm and the decoding process.
Our goal is to devise an adaptive component selection algorithm $\alg$ to minimize the expected decoding wall time compared to that of the optimal algorithm $\alg^*$, i.e., the \textit{wall-time regret} (with the unit \textit{second}):
\begin{align}
    \WTReg(\Lambda,\prompt)
    :=
    \bbE[\WT_{\ST(\alg,\prompt,\Lambda)}]
    -
    \bbE[\WT_{\ST(\alg^*,\prompt,\Lambda)}].
\end{align}
In practice, this wall-time regret characterizes the expected additional time $\alg$ requires, in addition to that of the oracle optimal algorithm $\alg^*$, to complete the generation of a given prompt $\prompt$. 
Therefore, it is expected that an algorithm $\alg$ can achieve $o(\ST(\alg^*,\prompt,\Lambda))$ wall-time regret, suggesting that the additional wall time $\alg$ demands is \emph{negligible}.

% \begin{algorithm}[t]
% \textbf{Inputs:} prefix tokens $\prompt\in\calX^*$, target model $P$, the draft model $Q$, maximum speculation length $L$.\\
% \textbf{Procedures:}
% \caption{\textsc{Specuative Decoding Subroutine} (\SDR)}\label{algo:SDR}    
% \begin{algorithmic}[1]
% \STATE  Call a  standard speculative decoding algorithm with $(\prompt, P, Q, L)$.\label{lne:decode}
% \STATE \return{} the accepted and bonus tokens $x_{1:\rho}$, where $\rho\!\geq\! 1$.
% \end{algorithmic}
% \end{algorithm}

%% simplification
In order to model the problem in a fine-grained manner, we study two factors of this framework: the mean accepted tokens and the time cost.

\subsection{Diminishing Return in the Mean Accepted Tokens}
As \citet{hou2025banditspec,liu2026notabandit,zhang2026learning} indicate, the stationary mean assumption on the number of accepted tokens is aligned with the real-life scenario, we follow their assumption.
\begin{assumption}[Stationary Mean of the Number of Accepted Tokens]\label{ass:stationary_mean}
    There exists a function $\mu:\calS\to [1,L+1]$, such that \begin{align}
        \bbE[\len(x_{S_t,t})\mid S_t,\calH_{t-1},\prompt] = \mu_{S_t}.
    \end{align}
\end{assumption}
This assumption enables the algorithm to predict the quality of the draft models based on their usage history.
% Furthermore, as inspired by \citet{sadhukhan2025magicdec}, we introduce the notion of submodularity and further assume the mean number of accepted tokens $\mu:\calS\to[1,L+1]$ is submodular.
% \begin{definition}[\citet{balcan2018submodular}]
%     Let $[n]$ denote a ground set of items and let $2^{[n]}$ be the power set of $[n]$. A function $f:2^{[n]}\to\bbR$ is submodular if it satisfies
%     \begin{align}
%         f(T \cup \{i\}) - f(T) \leq f(S \cup \{i\}) - f(S), \quad \forall S \subseteq T \subseteq [n],\ i \in [n].
%     \end{align}
% \end{definition}
% \begin{assumption}[Submodularity]\label{ass:submodular}
%     $\mu:\calS\to[1,L+1]$ is a submodular function.
% \end{assumption}
Furthermore, motivated by \citet{sadhukhan2025magicdec}, we assume that the mean number of accepted tokens exhibits diminishing returns along the ordered configurations considered by our method. Let $\bar{\calS}:=\{S_k\}_{k=1}^K\subset\mathcal{S}$ denote a set of ordered configurations with $S_k\subset S_{k+1}$ with equal increments in cost~(see Section~\ref{sec:linear_cost}).

\begin{figure}[t]
    \centering
    \subfigure[Acceptance rate and time.]{%
        \includegraphics[width=0.35084\linewidth]{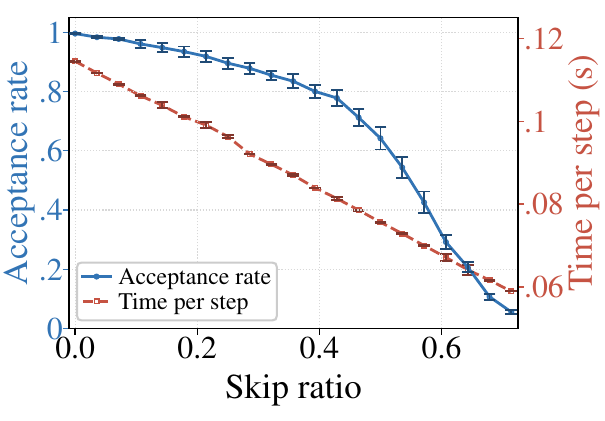}%
    }\hfill%
    \subfigure[Throughput.]{%
        \includegraphics[width=0.31262\linewidth]{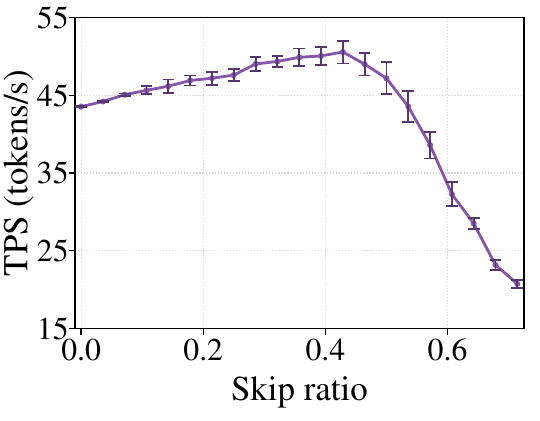}%
    }\hfill%
    \subfigure[Optimal arm distribution.]{%
        \includegraphics[width=0.31262\linewidth]{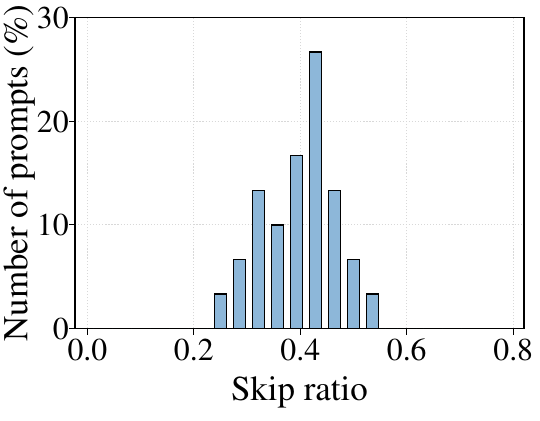}%
    }
    \caption{
    Empirical validation of the layer-skipping candidate arms with \textsc{cosSim} on Qwen2.5-VL-7B-Instruct.
    (a) Average acceptance rate shows a diminishing return pattern and time per speculative decoding step is an affine function in skip ratio, supporting Assumptions~\ref{ass:diminishing} and~\ref{ass:affine_cost}.
    (b) Tokens per second (TPS) illustrate the single-peak structure of throughput. 
    (c) Percentage of prompts for which the arm at each skip ratio is oracle best, i.e., attains the highest measured TPS. The best arm varies across prompts, motivating prompt-adaptive arm selection.
    % where the oracle optimal arm is the skip ratio with the highest TPS for each prompt.
    % The horizontal axis is the fraction of skipped layers.
    % Figure (a) empirically supports Assumptions~\ref{ass:diminishing} and~\ref{ass:affine_cost}, figure (b) illustrates the single-peak TPS property, and figure (c) shows that the optimal configuration is prompt-dependent.
    }
    \label{fig:assumption_validation}
\end{figure}

\begin{assumption}[Diminishing Returns]\label{ass:diminishing}
Along the sequence $\{S_k\}_{k=1}^K$, the mean number of accepted tokens $\mu:\mathcal{S}\to[1,L+1]$ satisfies
\begin{align}
0 \leq \mu(S_{k+1})-\mu(S_{k})
\leq \mu(S_{k})-\mu(S_{k-1}).
\end{align}
\end{assumption}
This assumption indicates that increasing the number of components improves the expected number of accepted tokens, but the marginal improvement decreases as more components are included.
% We empirically demonstrate that
% Our \textsc{cosSim} (detailed in ) instantiate this framework by constructing a set of ordered configuration $\{S_k\}_{k=1}^K$ with varying skip ratio of the model blocks.
This assumption is empirically supported by the validation on \textsc{cosSim}, a simple layer-skip speculative decoding method to be introduced in Section~\ref{subsec:layer_skip_config}, where each draft model is indexed by the layer-skip ratio.
Figure~\ref{fig:assumption_validation}(a) demonstrates that decreasing skip-ratio (i.e., increasing retained ratio) results in diminishing gains in acceptance rate.
% once a moderate fraction of layers is already kept.
In particular, keeping as few as $55\%$ layers can achieve a mean acceptance rate $80\%$.

% Our empirical observations illustrated in Figure~\ref{fig:assumption_validation}(a) shows the average acceptance rate of Qwen3-VL-4B~\citep{qwen3vltechnicalreport} over the Video-MME~\citep{videomme} benchmark as the layer skip ratio varies. 
% It demonstrates that keeping more layers improves acceptance, with the gain becoming marginal once a moderate fraction of layers is already kept.
% In particular, keeping as few as $50\%$ layers can achieve a mean acceptance rate $80\%$.

%\yunlong{for yuqi: maybe we need a figure to show, different prompts require different layer skip ratio }

\subsection{Linearity of the Time Cost}\label{sec:linear_cost}

We denote the non-decreasing time cost function as  $c:\{|S|:S\in\calS\}\to\bbR_+$, where $c(s)$ represents the time consumed by one speculative decoding round with draft model $S$.
As the time cost $c(s)$ only depends on deterministic factors, such as the number of tokens, the model, and the GPU employed, it can be considered as deterministic and can be measured or profiled on the validation dataset before the decoding process begins. Therefore, we assume the cost function $c$ is known at the beginning.
In addition, according to~\citet{sadhukhan2025magicdec}, the  time for one forward pass can be characterized as 
\begin{align}
    % c(s) = \gamma_M \cdot N_{\textrm{batched}} + \underbrace{\alpha_M\cdot N_{\textrm{context}} + \delta}_{b}
    %  =  \gamma \cdot s  + b,
    c(s) = L\cdot T_D(s;N_{\textrm{batch}},N_{\textrm{token}}) + T_V(N_{\textrm{batch}},N_{\textrm{token}},L)
\end{align}
where $T_D(s;N_{\textrm{batch}},N_{\textrm{token}})$ is the decoding time given $N_{\textrm{batch}}$ batches with sequence length $N_{\textrm{token}}$ and $s$ retained components in the draft model; $T_V(N_{\textrm{batch}},N_{\textrm{token}},L)$ is the verification time for the target model.
We start with the case where the generation length is short so that the change of $N_{\textrm{token}} $ on the time cost $c(s)$ is negligible. 
We show that the algorithm is robust to the time cost function in Appendix~\ref{app:ablation_reset} when $N_{\textrm{token}} $ increases during decoding.
\begin{assumption}\label{ass:affine_cost}
    Given the hardware and models, the time cost for one speculative decoding with $s$ retained components is
    \begin{equation}
        c(s) =  \kappa \cdot s  + b,
    \end{equation}
    where $\kappa$ and $b$ are constants depending on the hardware and models.
\end{assumption}
Figure~\ref{fig:assumption_validation}(a) reports the average time cost per speculative decoding round of \textsc{cosSim} over different layer-skip ratios.
% , where $s$ in Assumption~\ref{ass:affine_cost} corresponds to the layer retained ratio.
The time cost decreases linearly with the skip ratio, equivalently, increases linearly with the retained ratio $s$, which validates the above assumption.
Furthermore, the standard deviation of time cost is small enough that we safely treat it as deterministic.
% We provide more verifications of Assumption~\ref{ass:diminishing} and~\ref{ass:affine_cost} in Appendix~\yunlong{give the reference} for other self-speculative decoding methods\yunlong{cite}.

We define the \emph{decoding speed function} $\nu_S:= \frac{\mu_S}{c_S}$.
When the time cost function is deterministic, under Assumption~\ref{ass:stationary_mean}, the oracle optimal algorithm $\alg^*$ becomes the algorithm that always adopts the \textit{oracle optimal layer configuration} $S^*=\argmax_{S\in\calS}\nu_S$, which we denote as $\alg_{S^*}$.
\begin{lemma}\label{lem:optimal_policy} 
    When $\bbE[\ST(\alg,\prompt,\Lambda)]<\infty$ and the time cost function is deterministic, under Assumption~\ref{ass:stationary_mean}, we have \begin{equation}
        \alg_{S^*}\in\argmin_{\alg} \WTReg(\Lambda,\prompt).
    \end{equation}
\end{lemma}
% The proof is postponed to Appendix~\ref{app:proof_optimal_policy}.

\subsection{Draft Model Construction}\label{subsec:layer_skip_config}
Our algorithm will operate on an ordered sequence of draft models, each treated as an arm.
In the following, we take the layer-skip self-speculative decoding method for demonstration purposes, where we propose a simple cosine-similarity-based construction of the ordered sequence of draft models.
Our algorithm can readily be applied on top of other algorithms as an adaptive booster.

% Their draft model configurations can be constructed using our \textsc{cosSim} scheme or inherited from existing self-speculative decoding methods.

\textbf{Cosine-similarity-based construction for layer-skipping method.}
Given a target model $P$ with $N$ transformer blocks, skipping entire blocks yields up to $2^N$ layer-skip configurations, many of which may be suboptimal~\citep{elhoushi2024layerskip,xia2025swift,qiao2025ig}.
To reduce this search space, \textsc{cosSim} constructs a sequence of draft models by removing \emph{redundant} blocks one at a time in a recursive approach.
Specifically,
we define the \emph{redundancy score} of a block as the mean cosine similarity between its input and output hidden states: a higher score indicates more closely aligned representations, suggesting greater redundancy.
To obtain $K<N$ draft models,
after an initial token generation with the full model $P$, we remove the block with the highest redundancy score to obtain the first draft model $S_K$.
With this first draft model, a speculative decoding step is conducted and the redundancy scores of the remaining blocks get updated.
% We then remove the eligible block with the highest updated score and repeat, obtaining a sequence of draft models that each skip one additional block.
We then remove the block with the highest updated score among the remaining blocks to obtain the second draft model $S_{K-1}$.
This procedure can be conducted recursively which results in a sequence of draft models of decreasing size $\{S_k\}_{k=1}^K$.
% and repeat, obtaining a sequence of draft models that each skip one additional block.
%
% Each construction step generates new tokens and supplies the new arm's initial accepted-token count for bandit initialization.
% Thus, \textsc{cosSim} is efficient in both configuration-search and initial bandit warm-up. 
Algorithm~\ref{algo:cossim_construction} in Appendix~\ref{app:draft_construction} details the procedure.
% \yunlong{a bit complicated, make it more concise with more clear flows.}

\textsc{cosSim} yields an ordered sequence of layer-skip configurations by increasing retained-layer count, i.e., $\bar{\calS}=\{S_k\}_{k=1}^K$ with $S_k\subset S_{k+1}$. As $k$ decreases, the configuration sequentially removes the most redundant block, which is expected to satisfy the diminishing acceptance gains assumption and the affine cost modelling.
Figure~\ref{fig:assumption_validation}(a) and (b) support these trends and the resulting single-peak throughput; additional results, including prompt-level measurements, can be found in Appendix~\ref{app:single_peak_cossim}.

% For analysis, we index the $K$ constructed layer-skip configurations by increasing retained-layer count, yielding a nested path $\bcalS=\{S_i\}_{i=1}^K$ with $S_i\subset S_{i+1}$.
% Sequentially removing the most redundant block is intended to yield diminishing acceptance gains when this path is traversed in reverse construction order, while its one-block (two-layer) increments are consistent with the affine cost model.
% Figure~\ref{fig:assumption_validation}(a,b) supports these trends and the resulting single-peak throughput; additional results, including prompt-level measurements can be found in Appendix~\ref{app:single_peak_cossim}.

Write $\mu_i=\mu_{S_i}$, $c_i=c(|S_i|)$, and $\nu_i=\mu_i/c_i$, and denote the oracle optimal arm by $i^*=\argmax_{i\in[K]}\nu_i$, which is assumed to be \emph{unique} in the given ordered arm set.
Under the stated path-wise and cost assumptions, we obtain the following property.
% TODO: Define ass:pathwise_diminishing when revising the assumptions.
\begin{lemma}\label{lem:unimodal}
    Under Assumptions~\ref{ass:diminishing} and \ref{ass:affine_cost}, $\nu:[K]\to\bbR_+$ is a single-peak function.
\end{lemma}

We remark that the optimal retained ratio (or skip ratio) that attains the throughput peak varies across prompts.
Figure~\ref{fig:assumption_validation}(c) reports the distribution of the oracle optimal skip ratio, defined by the highest TPS for each prompt.
The mass is spread across multiple skip ratios, indicating that \emph{no single draft model is uniformly optimal} and motivating prompt-level adaptive arm selection.

\textbf{Construction from existing methods.}
In the above, we demonstrate how the draft models are constructed via the simple \textsc{cosSim} layer-skip method, which satisfies the diminishing return and affine cost assumptions (Assumptions~\ref{ass:diminishing} and~\ref{ass:affine_cost}). We remark that these assumptions are also satisfied in existing self-speculative decoding methods.
For example, 
\begin{itemize}[leftmargin=*]
    \item Given a layer-skip ratio $r\in(0,1)$, \textsc{SWIFT}~\citep{xia2025swift} combines random search and Bayesian optimization to select attention and MLP layers to skip. Decreasing $r$ results in a larger and more capable draft model.
    \item Under latency budgets $b$, \textsc{KnapSpec}~\citep{cha2026knapspec} selects retained layers through knapsack optimization. Increasing $b$ allows more forward time for the draft model.
    \item With pruning ratio $p\in(0,1)$, \textsc{SpecVLM}~\citep{ji2025specvlm} prunes visual tokens using verifier-guided importance scores and uses the reduced visual context for draft generation. Enlarging $p$ yields a more compact prefix token sequence.
\end{itemize}
The amount of retained components $|S|$ is controlled by the skip ratio $r$ in \textsc{SWIFT}, latency budget $b$ in \textsc{KnapSpec} and pruning ratio $p$ in \textsc{SpecVLM} respectively. 
Varying these parameters yields an ordered sequence of draft models $\bar{\calS}=\{S_k\}_{k=1}^K$, i.e., the set of arms in our bandit formulation.

By construction, the time cost is expected to be an affine function of $|S|$, satisfying Assumption~\ref{ass:affine_cost}.
For the diminishing return Assumption~\ref{ass:diminishing}, intuitively, these methods prioritize retaining the layers or visual tokens that help the draft model match the target model.
As $|S|$ grows, the remaining components (e.g., layers or tokens) are less essential for the draft model, so further additions are expected to yield smaller acceptance gains.
We further empirically verify that Assumptions~\ref{ass:diminishing} and \ref{ass:affine_cost} are met in Appendix \ref{app:single_peak}.
Therefore, according to Lemma~\ref{lem:unimodal}, the single-peak structure can be obtained.
Further draft model construction details of these methods are postponed to Appendix~\ref{app:existing_draft_construction}.

Given these self-speculative decoding methods, a fixed choice of the component need not remain optimal across models, datasets, or prompts (see Appendix~\ref{app:single_peak}).
In the following section, we introduce a bandit controller that adapts this choice online using accepted-token and wall-time feedback.
It strongly exploits the single-peak structure (Lemma~\ref{lem:unimodal}) to 
quickly locate the optimal components for the given prompt.
This provides a broadly applicable meta-algorithm for making existing self-speculative decoding methods more adaptive across models, datasets, and prompts.

To efficiently identify the best retained components, we aim to minimize the wall-time regret with respect to this ordered sequence of draft models $\bar{\calS}$ instead of the whole draft model set $\calS$. This enables us to make use of the single-peak structure identified by Lemma~\ref{lem:unimodal}.

\section{The \spsqec{} Algorithm}
\begin{algorithm}[t]
    \caption{\textsc{Single-Peak Speculative Decoding}~(\spsqec{})}\label{algo:spsqec}
    \textbf{Inputs:} draft model set $\bcalS=[K]$, history $\calH_{t}=\big((I_s,X_{I_s,s})\big)_{s=1}^{t}$, time cost profile $\{c_i\}_{i\in[K]}$.\\
    \textbf{Procedures:}
    \begin{algorithmic}[1]
    \STATE \textbf{if} $t\leq K-1$ \textbf{then} \return{} $I_{t+1}=t+1$.
    % \STATE Compute the lengths $Y_{I_s,s}=\len(X_{I_s,s})$ for all $s\in[t].\!\!\!$
    \STATE Set the statistics $\{\hmu_{i}(t)\}_{i\in[K]}$, where
    \begin{align}
        &
        T_{i}(t) = \sum_{s=1}^{t}\mathbbm{1}\{I_s=i\},\;\;
        \hmu_{i}(t) = \frac{\sum_{s=1}^{t} \len(X_{i,s})\mathbbm{1}\{I_s=i\}}{T_{i}(t)},\;\;
        L_t = \argmax_i \frac{\hmu_{i}(t)}{c_i},\;
    \end{align}
    \label{algline:statistics}
    \STATE \textbf{if} $\tilT_{L_{t}}(t) \equiv 1 \pmod{3}$, \textbf{then} $I_{t+1}=L_t$
    \textbf{else} $I_{t+1} = \argmax_{i\in\{L_t-1,L_t,L_t+1\}\cap[K]}\mathrm{UCB}_{i,L_t}(t)$,
    where for $i\in \{L_t-1,L_t,L_t+1\}\cap[K]$
    \begin{align}\label{equ:ucb}
        &\mathrm{UCB}_{i,L_t}(t) = \frac{\hmu_{i}(t) + \mathrm{cr}_{i,L_t}(t)}{c_i},
        \;\;
        \tilT_{i}(t) = \sum_{s=K}^{t}\mathbbm{1}\{L_s=i\},
        % L\cdot \sqrt{\frac{\alpha\,\log\!\big(l_{L_t,t}\big)}{2n_{i,t}}},
        \\
         &\mathrm{cr}_{i,L_t}(t) = \frac{L}{2}\sqrt{
            \frac{1+T_i(t)}{T_i^2(t)}\Big(
            1+2\ln\big(3\cdot\tilT_{L_t}^2(t)\cdot(1+T_i(t))^{1/2}\big)
            \Big)
        }.
    \end{align}
    \label{algline:ucb}
    \STATE \return{}  index $I_{t+1}$.
    \label{algline:return}
    \end{algorithmic}
\end{algorithm}

We formally present the \textsc{Single-Peak Speculative Decoding} (\spsqec{}) algorithm in Algorithm~\ref{algo:spsqec}, which adaptively selects the retained components given the prompt and exploits the single-peak structure in the set of draft candidates $\bcalS=[K]$.\footnote{For simplicity, we use $k\in[K]$ to indicate draft model $S_k$.}

Given the sampling history $\calH_t$, we firstly compute the statistics: the arm pulls $T_i(t)$ and the empirical mean $\hmu_i(t)$ for arm $i\in[K]$, and define the leader arm $L_t$ which has the largest empirical decoding speed (Line~\ref{algline:statistics}).
Afterwards, we decide which arm should be sampled only in the neighborhoods $\calN(L_t):=\{L_t-1,L_t,L_t+1\}\cap[K]$ of the leader arm $L_t$: 
% let $\tilT_{L_t}(t)$ be the number of times arm $L_t$ has been the leader, which can be regarded as a \emph{local time} at $L_t$. 
% The algorithm is forced to sample the leader arm $L_t$ every three times according to the local time $\tilT_{L_t}(t)$; otherwise, the algorithm samples the arm with the highest local $\ucb_{i,L_t}$ index as \eqref{equ:ucb} in the neighborhood $\calN(L_t)$ (Line~\ref{algline:ucb}).
the leader arm $L_t$ is forced to be sampled every three times according to the \emph{local time $\tilT_{L_t}(t)$}; otherwise, the arm with the highest local $\ucb_{i,L_t}$ index as~\eqref{equ:ucb} in the neighborhood $\calN(L_t)$ is sampled (Line~\ref{algline:ucb}).
In the end, the arm index $I_{t+1}$ is returned and the corresponding draft model will be used to carry out one round of speculative decoding as in Line~\ref{lne:spec} of Algorithm~\ref{algo:cost_aware_banditspec}.

We present the theoretical guarantee of the wall-time regret of \spsqec{}.
Given an instance $\Lambda$ and a prompt $\prompt$, we define the \textit{hardness parameter},
\begin{align}
    \rmH(\Lambda,\prompt):=\sum_{i\in\calN(i^*)\setminus\{i^*\}}\frac{1}{\nu_{i^*}c_i}\cdot \frac{8L^2}{\nu_{i^*}-\nu_i}  .
\end{align}
\begin{theorem}\label{thm:spsqec}
    Under Assumptions~\ref{ass:stationary_mean}, \ref{ass:diminishing} and \ref{ass:affine_cost}, and we further assume the stopping time has finite expectation $\bbE[\ST]<\infty$, then adopting $\alg$=\spsqec{} in Algorithm~\ref{algo:cost_aware_banditspec}  with  bandit configuration $\Lambda =(P,\calQ=\{Q_S\}_{S\in\bar{\calS}},L)$
    achieves wall-time regret 
    \begin{align}
    \WTReg(\Lambda,\prompt)
    &\leq
    \sum_{i\in\calN(i^*)\setminus\{i^*\}}\mu_i\Big(\frac{1}{\nu_{i}}-\frac{1}{\nu_{i^*}}\Big)\cdot \frac{8L^2\ln\bbE[\ST(\alg,\prompt,\Lambda) ]}{(c_i\nu_{i^*}-\mu_i)^2} +O(1)
    \\
    &=
    \rmH(\Lambda,\prompt) \cdot \ln \bbE[\ST(\alg,\prompt,\Lambda) ]+O(1).\label{equ:wtreg}
\end{align}
\end{theorem}
The big-O in~\eqref{equ:wtreg} hides an instance-dependent constant, but is \emph{independent} of the expected number of speculative decoding rounds $\bbE[\tau]$. 
The proof of Theorem~\ref{thm:spsqec} is postponed to Appendix~\ref{app:proof_main}.
It theoretically guarantees that the additional wall-time latency of \spsqec{} compared to the algorithm that always adopts the oracle best draft model $\alg_{i^*}$ is of order $o(\bbE[\ST(\alg,\prompt,\Lambda) ])$.
Furthermore, we have identified a critical quantity, the hardness parameter $\rmH(\Lambda,\prompt)$, which suggests the hardness of the instance under the given prompt and bandit configuration. In particular, this hardness parameter only depends on the \emph{neighborhoods} of the oracle best arm $\calN(i^*)\setminus\{i^*\}$, instead of relying on all arms in $\bcalS$ as in the previous works~\citep{liu2026notabandit,hou2025banditspec,kim2026multidrafter}.
This improvement results from the single-peak structure embedded in the arm set $\bcalS$.

The dependence on $\calN(i^*)\setminus\{i^*\}$ also suggests that the costly full-information feedback is unnecessary under our problem formulation.
In~\citet{liu2026notabandit}, all draft models are evaluated after each speculative decoding round, which proves to be effective when the draft models are small and uncorrelated. Nevertheless, Theorem~\ref{thm:spsqec} suggests that the evaluation of all draft models is unnecessary, because most arms are quickly identified as suboptimal in $O(1)$ rounds and as few as two suboptimal arms will contribute to the leading term in the final wall-time regret.

During experiments, we will report the speedup ratio compared to the vanilla autoregressive decoding, following the convention in the community. Specifically,
for any prompt $\prompt\in\calX^*$, denote the expected wall time by the autoregressive decoding and by Algorithm~\ref{algo:cost_aware_banditspec} with $\alg{}$ as $\bbE[\WT_{\textrm{auto}}(\prompt)]$ and $\bbE[\WT_{\alg}(\prompt)]$, respectively. The speedup ratio of $\alg{}$ on this prompt is defined as $\mathrm{SU}:=\frac{\bbE[\WT_{\textrm{auto}}(\prompt)]}{\bbE[\WT_{\alg}(\prompt)]}$.
\begin{proposition}[Transformation from Wall-time Regret to Speedup Ratio]
    Given a prompt $\prompt$ and a bandit configuration $\Lambda$, let $\alg^*$ denote Algorithm~\ref{algo:cost_aware_banditspec} with the optimal draft model,
    \spsqec{} achieves the speedup ratio 
    \begin{equation}
    \mathrm{SU}=\frac{\bbE[\WT_{\textrm{auto}}(\prompt)]}{\bbE[\WT_{\alg^*}(\prompt)]+\WTReg(\Lambda,\prompt)}.    
    \end{equation}
\end{proposition}
The proposition can be obtained directly by the definition of wall-time regret.
This provides a direct interpretation of the wall-time regret in Theorem~\ref{thm:spsqec}: because $\WTReg(\Lambda,\prompt)=O\big(\ln \bbE[\ST(\alg,\prompt,\Lambda) ]\big)=O\big(\ln \bbE[\WT_{\textrm{auto}}(\prompt)]\big)$, \spsqec{} approaches the optimal speedup $\frac{\bbE[\WT_{\textrm{auto}}(\prompt)]}{\bbE[\WT_{\alg^*}(\prompt)]}$ as the inference proceeds.

\section{Experiments}
\label{sec:experiments}

\textbf{Experimental Setups.} We evaluate Qwen2.5-VL-7B-Instruct~\citep{qwen2.5vltechnicalreport} and LLaVA-NeXT-Mistral-7B~\citep{liu2024llavanext} on four visual and textual benchmarks: Video-MME~\citep{videomme}, MMBench-Video~\citep{fang2024mmbenchvideo}, ARKitScenes~\citep{arkitscenes}, and AIME25~\citep{aime25}.
Unless otherwise stated, we use greedy decoding, batch size $1$, and draft length $L=4$ for all speculative decoding methods.
For stable latency measurements, we add a suffix instruction to all prompts to produce longer responses.
We report Mean Accepted Tokens (MAT), including $1$ bonus token from verification, and the normalized SpeedUp ratio (SU) over vanilla autoregressive decoding.
Reported wall-clock speedup includes optimization, warm-up, and arm-construction overhead.

We apply \spsqec{} to four self-speculative configuration schemes: our cosine-similarity-based layer skipping (\textsc{cosSim}, as described in Section~\ref{subsec:layer_skip_config}), \swift{}~\citep{xia2025swift}, \textsc{KnapSpec}~\citep{cha2026knapspec}, and \textsc{SpecVLM}~\citep{ji2025specvlm}.
For each scheme, we compare its original version (with its original drafter-configuration selection strategy) with two adaptive variants that use \ucbspec{}~\citep{hou2025banditspec} or \ourmethod{} to select the draft model online during generation.
Within each scheme, the two bandit controllers use the same candidate arm set, 
% isolating the effect of the arm-selection rule, 
with \ucbspec{} treating the same arms independently and \ourmethod{} exploiting the ordered single-peak structure.
Appendix~\ref{app:experiment_details} provides further implementation details.

The implementation is accessible via~\url{https://github.com/hyuqi/sp2ec}.

% Source: data/Table1_Main_SpeedUp.xlsx, sheet 'Main result', rows 3--14.
% Mapping: cosine = cosSim; UCB = UCBSpec; UniUCB = SP2EC.
% Reported SU cells are preserved; averages are recomputed before rounding.
% SpecVLM AIME25 formula zeros reference blank TPS cells and are not measurements.
\begin{table*}[t]
\caption{Main results on Qwen2.5-VL-7B-Instruct. MAT
denotes mean accepted tokens, and SU denotes normalized speedup. \ourmethod{} achieves consistent wall-time speedup. 
% over vanilla autoregressive decoding. For the \textsc{SpecVLM} variants, AIME25 results are unavailable (marked --). Gray rows denote variants augmented by \ourmethod{}.
}
\label{tab:main_results}
\label{tab:main_results_qwen}
\centering
\small
\setlength{\tabcolsep}{3.5pt}
\renewcommand{\arraystretch}{1.12}
\resizebox{\textwidth}{!}{
\begin{tabular}{l*{10}{c}}
\toprule
\multirow{2}{*}{Method}
& \multicolumn{2}{c}{Video-MME}
& \multicolumn{2}{c}{MMBench-Video}
& \multicolumn{2}{c}{ARKitScenes}
& \multicolumn{2}{c}{AIME25}
& \multicolumn{2}{c}{Average} \\
\cmidrule(lr){2-3}
\cmidrule(lr){4-5}
\cmidrule(lr){6-7}
\cmidrule(lr){8-9}
\cmidrule(lr){10-11}
& MAT $\uparrow$ & SU $\uparrow$
& MAT $\uparrow$ & SU $\uparrow$
& MAT $\uparrow$ & SU $\uparrow$
& MAT $\uparrow$ & SU $\uparrow$
& MAT $\uparrow$ & SU $\uparrow$ \\
\midrule
\textsc{cosSim} + \ucbspec{} & 2.53 & 1.21$\times$ & 2.84 & 1.16$\times$ & 4.33 & 1.15$\times$ & 4.08 & 1.13$\times$ & 3.45 & 1.16$\times$ \\
\rowcolor{sp2ecrow}
\textsc{cosSim} + \ourmethod{} & 2.86 & \textbf{1.28}$\times$ & 3.38 & \textbf{1.27}$\times$ & 3.87 & \textbf{1.25}$\times$ & 3.35 & \textbf{1.22}$\times$ & 3.37 & \textbf{1.26}$\times$ \\
\midrule
\swift{} & 2.43 & 1.14$\times$ & 2.60 & 1.16$\times$ & 3.22 & 1.22$\times$ & 2.92 & 1.10$\times$ & 2.79 & 1.16$\times$ \\
\swift{} + \ucbspec{} & 2.76 & 1.16$\times$ & 2.87 & 1.13$\times$ & 3.80 & 1.20$\times$ & 3.45 & 1.15$\times$ & 3.22 & 1.16$\times$ \\
\rowcolor{sp2ecrow}
\swift{} + \ourmethod{} & 2.51 & \textbf{1.23}$\times$ & 2.65 & \textbf{1.19}$\times$ & 3.55 & \textbf{1.24}$\times$ & 3.29 & \textbf{1.22}$\times$ & 3.00 & \textbf{1.22}$\times$ \\
\midrule
\textsc{KnapSpec} & 3.38 & 1.22$\times$ & 4.31 & 1.34$\times$ & 4.32 & 1.30$\times$ & 4.45 & 1.28$\times$ & 4.12 & 1.29$\times$ \\
\textsc{KnapSpec} + \ucbspec{} & 3.75 & 1.17$\times$ & 4.64 & 1.27$\times$ & 4.57 & 1.28$\times$ & 4.55 & 1.25$\times$ & 4.38 & 1.24$\times$ \\
\rowcolor{sp2ecrow}
\textsc{KnapSpec} + \ourmethod{} & 3.62 & \textbf{1.31}$\times$ & 4.39 & \textbf{1.40}$\times$ & 4.24 & \textbf{1.35}$\times$ & 4.29 & \textbf{1.32}$\times$ & 4.14 & \textbf{1.35}$\times$ \\
\midrule
\textsc{SpecVLM} & 3.54 & 1.50$\times$ & 3.60 & 1.51$\times$ & 3.60 & 1.39$\times$ & -- & -- & 3.58 & 1.47$\times$ \\
\textsc{SpecVLM} + \ucbspec{} & 3.86 & 1.55$\times$ & 3.66 & 1.50$\times$ & 3.63 & 1.41$\times$ & -- & -- & 3.72 & 1.49$\times$ \\
\rowcolor{sp2ecrow}
\textsc{SpecVLM} + \ourmethod{} & 3.99 & \textbf{1.62}$\times$ & 3.90 & \textbf{1.60}$\times$ & 3.75 & \textbf{1.47}$\times$ & -- & -- & 3.88 & \textbf{1.56}$\times$ \\
\bottomrule
\end{tabular}
}
\end{table*}

\begin{table*}[t]
\caption{Main results on LLaVA-NeXT-Mistral-7B. 
% MAT denotes mean accepted tokens, and SU denotes normalized speedup. 
\ourmethod{} demonstrates consistent wall-time speedup.
% over vanilla autoregressive decoding. For the \textsc{SpecVLM} variants, AIME25 results are unavailable (marked --), and averages use the three visual benchmarks. Gray rows denote variants augmented by \ourmethod{}.
}
\label{tab:main_results_llava}
\centering
\small
\setlength{\tabcolsep}{3.5pt}
\renewcommand{\arraystretch}{1.12}
\resizebox{\textwidth}{!}{
\begin{tabular}{l*{10}{c}}
\toprule
\multirow{2}{*}{Method}
& \multicolumn{2}{c}{Video-MME}
& \multicolumn{2}{c}{MMBench-Video}
& \multicolumn{2}{c}{ARKitScenes}
& \multicolumn{2}{c}{AIME25}
& \multicolumn{2}{c}{Average} \\
\cmidrule(lr){2-3}
\cmidrule(lr){4-5}
\cmidrule(lr){6-7}
\cmidrule(lr){8-9}
\cmidrule(lr){10-11}
& MAT $\uparrow$ & SU $\uparrow$
& MAT $\uparrow$ & SU $\uparrow$
& MAT $\uparrow$ & SU $\uparrow$
& MAT $\uparrow$ & SU $\uparrow$
& MAT $\uparrow$ & SU $\uparrow$ \\
\midrule
\textsc{cosSim} + \ucbspec{} & 3.30 & 1.19$\times$ & 3.19 & 1.20$\times$ & 3.09 & 1.18$\times$ & 4.56 & 1.25$\times$ & 3.54 & 1.21$\times$ \\
\rowcolor{sp2ecrow}
\textsc{cosSim} + \ourmethod{} & 3.00 & \textbf{1.26}$\times$ & 2.87 & \textbf{1.26}$\times$ & 2.80 & \textbf{1.23}$\times$ & 4.28 & \textbf{1.40}$\times$ & 3.24 & \textbf{1.29}$\times$ \\
\midrule
\swift{} & 3.00 & 1.12$\times$ & 3.05 & 1.22$\times$ & 3.11 & 1.12$\times$ & 4.11 & 1.21$\times$ & 3.32 & 1.17$\times$ \\
\swift{} + \ucbspec{} & 3.01 & 1.14$\times$ & 3.33 & 1.22$\times$ & 3.15 & 1.15$\times$ & 4.35 & 1.26$\times$ & 3.46 & 1.19$\times$ \\
\rowcolor{sp2ecrow}
\swift{} + \ourmethod{} & 2.87 & \textbf{1.19}$\times$ & 3.24 & \textbf{1.28}$\times$ & 3.04 & \textbf{1.22}$\times$ & 4.23 & \textbf{1.32}$\times$ & 3.35 & \textbf{1.25}$\times$ \\
\midrule
\textsc{KnapSpec} & 3.41 & 1.19$\times$ & 3.13 & 1.20$\times$ & 3.06 & 1.18$\times$ & 4.57 & 1.32$\times$ & 3.54 & 1.22$\times$ \\
\textsc{KnapSpec} + \ucbspec{} & 3.44 & 1.19$\times$ & 3.27 & 1.17$\times$ & 3.15 & 1.16$\times$ & 4.66 & 1.26$\times$ & 3.63 & 1.20$\times$ \\
\rowcolor{sp2ecrow}
\textsc{KnapSpec} + \ourmethod{} & 3.24 & \textbf{1.30}$\times$ & 3.02 & \textbf{1.30}$\times$ & 3.05 & \textbf{1.24}$\times$ & 4.56 & \textbf{1.40}$\times$ & 3.47 & \textbf{1.31}$\times$ \\
\midrule
\textsc{SpecVLM} & 3.95 & 1.36$\times$ & 4.00 & 1.37$\times$ & 3.89 & 1.42$\times$ & -- & -- & 3.95 & 1.38$\times$ \\
\textsc{SpecVLM} + \ucbspec{} & 4.20 & 1.42$\times$ & 4.09 & 1.38$\times$ & 4.01 & 1.44$\times$ & -- & -- & 4.10 & 1.41$\times$ \\
\rowcolor{sp2ecrow}
\textsc{SpecVLM} + \ourmethod{} & 4.17 & \textbf{1.48}$\times$ & 4.08 & \textbf{1.42}$\times$ & 4.00 & \textbf{1.46}$\times$ & -- & -- & 4.08 & \textbf{1.45}$\times$ \\
\bottomrule
\end{tabular}
}
\end{table*}

\begin{figure*}[t]
    \centering
    \includegraphics[width=\textwidth]{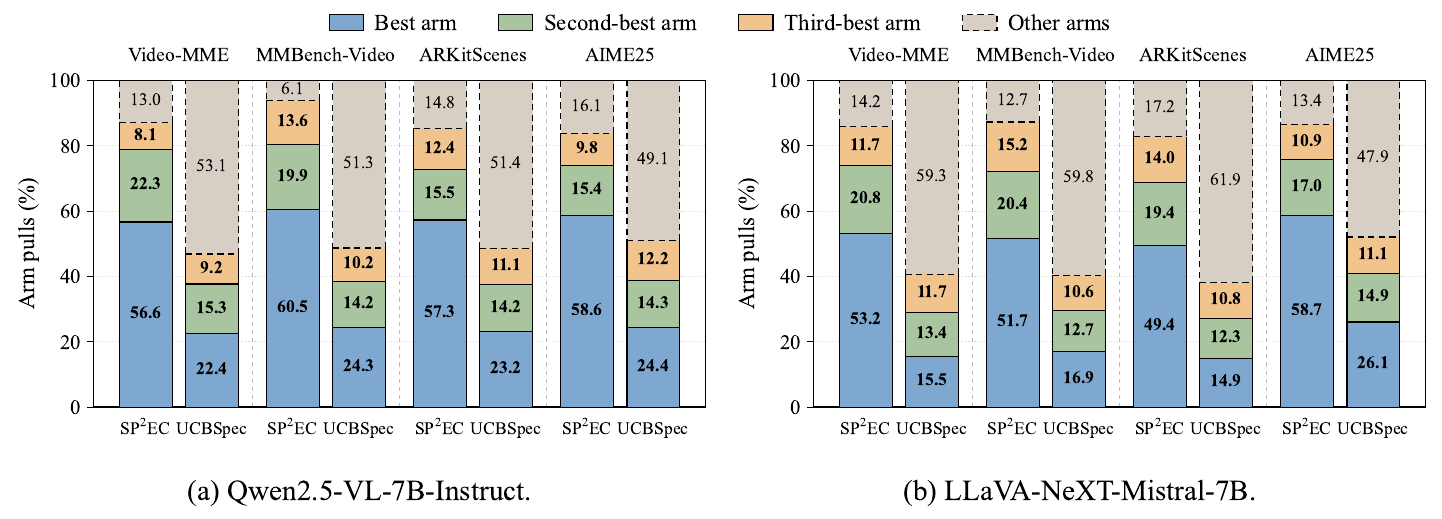}
    \caption{Arm-pull allocation on (a) Qwen2.5-VL-7B-Instruct and (b) LLaVA-NeXT-Mistral-7B across Video-MME, MMBench-Video, ARKitScenes, and AIME25.
    Each dataset group contains two stacked bars: \ourmethod{} (left) and \ucbspec{} (right).
    Segments show the percentage of pulls allocated to the best, second-best, third-best, and other arms.
    Bold labels highlight the percentages allocated to the top three arms; dashed outlines identify the other arms.}
    \label{fig:arm_pull_distribution}
\end{figure*}

\textbf{Experimental Results.}
Tables~\ref{tab:main_results_qwen} and~\ref{tab:main_results_llava} report results on Qwen2.5-VL-7B-Instruct and LLaVA-NeXT-Mistral-7B, respectively.
Across all $30$ available model-scheme-dataset combinations, \ourmethod{} consistently improves the self-speculative decoding baselines: \swift{}, \textsc{KnapSpec}, and \textsc{SpecVLM}, and achieves higher end-to-end speedup than the adaptive speculative decoding baseline~\ucbspec{}.
% For \textsc{cosSim}, \swift{}, \textsc{KnapSpec}, and \textsc{SpecVLM}, its average speedups are $1.26\times$, $1.22\times$, $1.35\times$, and $1.56\times$ on Qwen2.5-VL, and $1.29\times$, $1.25\times$, $1.31\times$, and $1.45\times$ on LLaVA-NeXT, respectively.
In particular, our algorithm achieves an additional average speed-up ranging from $0.06\times$ to $0.09\times$ compared to the self-speculative decoding baselines, with a maximum increment of $0.12\times$ on Video-MME benchmark with \textsc{SpecVLM} and on AIME25 with \swift{}.

This comparison also exposes the limitation of \ucbspec{}: online adaptation can harm efficiency when \emph{exploration is too costly}.
For example, for \textsc{KnapSpec}, adding \ucbspec{} \emph{reduces} average speedup from $1.29\times$ to $1.24\times$ for Qwen2.5-VL and from $1.22\times$ to $1.20\times$ on LLaVA-NeXT, whereas \ourmethod{} increases them to $1.35\times$ and $1.31\times$.
This observation is further corroborated by Figure~\ref{fig:arm_pull_distribution}: across the four Qwen2.5-VL benchmarks, \ucbspec{} allocates $49.1$-$53.1\%$ of its pulls \emph{outside} the top three arms.
% , compared with only $6.1$-$16.1\%$ for \ourmethod{}.
As each exploratory pull incurs drafting and target-model verification costs, allocating many pulls on lower-throughput draft configurations as \ucbspec{} does can offset the gains from online adaptation.
In contrast, \ourmethod{} exploits the single-peak structure and quickly concentrates pulls on the best draft configurations, e.g., only $6.1$-$16.1\%$ of arm pulls outside the top three arms.
This turns online adaptation into consistent acceleration and also highlights the importance of exploiting the single-peak structure in self-speculative decoding.
% highlighting the crucial role of controlling exploration in self-speculative decoding.
% This highlighting the essential need to make use of the single-peak structure in self-speculative decoding.

We also remark that higher MAT does not necessarily translate into higher speedup.
In particular, a more accurate drafter can incur greater computation cost, offsetting the benefit of accepting more tokens. Therefore, there is a trade-off between the drafting accuracy and computation latency, in which SP$^2$EC achieves the optimal balance theoretically and empirically. We further examine this trade-off through a breakdown of drafting, verification, and optimization overhead in Appendix~\ref{app:timing_breakdown}. 

\textbf{Ablations and Additional Results.}
Appendix~\ref{app:ablation} reports ablations on draft length $L$, input size (visual-token budget), output length, and effect of resetting the optimizer every $512$ speculative steps.
Across the reported length and input-size ranges, \ourmethod{} maintains the highest speedup among the compared variants within each scheme.
Its performance also varies least with hyperparameters.
These results support its robustness through online adaptation to more efficient and suitable draft models.

We compare with \eaglethree{}~\citep{li2025eagle3} in Appendix~\ref{app:eagle3_comparison}.
Our methods require no auxiliary training and avoid dependence on alignment between a separate draft model's training distribution and  test-time prompt distribution.
The comparison suggests that small auxiliary draft models can struggle with complex mixed visual and textual context, limiting token acceptance and speedup.

Additional results with \textsc{cosSim} on Qwen3-VL-4B-Instruct and LLaVA-1.5-7B are provided in Appendix~\ref{legacy:cossim_qwen3}.

\section{Conclusion}
\label{app:conclusion}

We presented \ourmethod{}, a training-free adaptive self-speculative decoding algorithm for vision-language models. 
\ourmethod{} helps to boost the performance of the existing self-speculative decoding methods by adapting the choice of components to the input prompt.
We formulate this component selection problem with wall-time regret and exploit the single-peak structure of the ordered draft model sequence, allowing \ourmethod{} to focus exploration near the best component configuration rather than treating all configurations independently. 

Our experiments on the two vision-language models, Qwen2.5-VL-7B-Instruct and LLaVA-NeXT-Mistral-7B, with four self-speculative decoding methods, namely \textsc{cosSim}, \swift{}~\citep{xia2025swift}, \textsc{KnapSpec}~\citep{cha2026knapspec}, and \textsc{SpecVLM}~\citep{ji2025specvlm}, show that \ourmethod{} consistently improves wall-clock speedup over training-free baselines, while the comparison with \eaglethree{} highlights the benefit of avoiding trained auxiliary drafters in multimodal settings. 

\paragraph{Limitations and Further Applications.} Our theory relies on structural assumptions on the candidate path, and our experiments focus on two VLM families.
This paper instantiates the framework with layer skipping and token selection; similar single-peak trade-offs may arise in other ordered acceleration knobs, such as KV-cache sparsification; we discuss such extensions in Appendix~\ref{app:more_application}.

\newpage
\section*{Acknowledgements and Disclosure}
Y.~Hou and V.~Y.~F.~Tan  are supported by a Singapore Ministry of Education AcRF Tier~2 grant (A-8004062-00-00). V.~Y.~F.~Tan is also supported by   a Singapore National Research Foundation Investigatorship (A-8004657-00-00). 

\bibliographystyle{plainnat}
\bibliography{reference}

@inproceedings{
liu2026notabandit,
title={Not-a-Bandit: Provably No-Regret Drafter Selection in Speculative Decoding for {LLM}s},
author={Hongyi Liu and Jiaji Huang and Zhen Jia and Youngsuk Park and Yu-Xiang Wang},
booktitle={The Fourteenth International Conference on Learning Representations},
year={2026},
url={https://openreview.net/forum?id=JMmljf895g}
}

@article{wang2024emu3,
  title={Emu3: Next-token prediction is all you need},
  author={Wang, Xinlong and Zhang, Xiaosong and Luo, Zhengxiong and Sun, Quan and Cui, Yufeng and Wang, Jinsheng and Zhang, Fan and Wang, Yueze and Li, Zhen and Yu, Qiying and others},
  journal={arXiv preprint arXiv:2409.18869},
  year={2024}
}

@article{
li2024enhancing,
title={Vi{DE}: Tuning-Free Video Coherence via Temporal Attention Reweighting and Prompt Blending},
author={Xingyao Li and Hao Zhang and Fengzhuo Zhang and Jiachun Pan and Yunlong Hou and Vincent Y. F. Tan and Zhuoran Yang},
journal={Transactions on Machine Learning Research},
issn={2835-8856},
year={2026},
url={https://openreview.net/forum?id=A5WZZxglxT},
note={}
}

@article{wang2025error,
  title={Error analyses of auto-regressive video diffusion models: A unified framework},
  author={Wang, Jing and Zhang, Fengzhuo and Li, Xiaoli and Tan, Vincent YF and Pang, Tianyu and Du, Chao and Sun, Aixin and Yang, Zhuoran},
  journal={arXiv preprint arXiv:2503.10704},
  year={2025}
}

@inproceedings{yang2026longspec,
  title={LongSpec: Long-Context Lossless Speculative Decoding with Efficient Drafting and Verification},
  author={Yang, Penghui and Du, Cunxiao and Zhang, Fengzhuo and Wang, Haonan and Pang, Tianyu and Du, Chao and An, Bo},
  booktitle={Proceedings of the 64th Annual Meeting of the Association for Computational Linguistics (Volume 1: Long Papers)},
  pages={1826--1844},
  year={2026}
}

@inproceedings{li2026annealed,
  title={Annealed relaxation of speculative decoding for faster autoregressive image generation},
  author={Li, Xingyao and Zhang, Fengzhuo and Du, Cunxiao and Ji, Hui},
  booktitle={Proceedings of the AAAI Conference on Artificial Intelligence},
  volume={40},
  number={28},
  pages={23212--23220},
  year={2026}
}

@article{li2025diffuspec,
  title={Diffuspec: Unlocking diffusion language models for speculative decoding},
  author={Li, Guanghao and Fu, Zhihui and Fang, Min and Zhao, Qibin and Tang, Ming and Yuan, Chun and Wang, Jun},
  journal={arXiv preprint arXiv:2510.02358},
  year={2025}
}

@inproceedings{wang2025alignment,
  title={Alignment-augmented speculative decoding with alignment sampling and conditional verification},
  author={Wang, Jikai and Tian, Zhenxu and Li, Juntao and Xia, Qingrong and Duan, Xinyu and Wang, Zhefeng and Huai, Baoxing and Zhang, Min},
  booktitle={Proceedings of the 2025 Conference on Empirical Methods in Natural Language Processing},
  pages={6762--6774},
  year={2025}
}

@inproceedings{yu2011Unimodal,
author = {Yu, Jia Yuan and Mannor, Shie},
title = {Unimodal bandits},
year = {2011},
isbn = {9781450306195},
publisher = {Omnipress},
address = {Madison, WI, USA},
booktitle = {Proceedings of the 28th International Conference on International Conference on Machine Learning},
pages = {41–48},
numpages = {8},
series = {ICML'11}
}

@article{tran2025goodspeed,
  title={GoodSpeed: Optimizing Fair Goodput with Adaptive Speculative Decoding in Distributed Edge Inference},
  author={Tran, Phuong and Liu, Tzu-Hao and Le, Long Tan and Nguyen, Tung-Anh and La, Van Quan and Yu, Eason and Shu, Han and Hong, Choong Seon and Tran, Nguyen H},
  journal={arXiv preprint arXiv:2512.09963},
  year={2025}
}

@article{mamou2024dynamic,
  title={Dynamic speculation lookahead accelerates speculative decoding of large language models},
  author={Mamou, Jonathan and Pereg, Oren and Korat, Daniel and Berchansky, Moshe and Timor, Nadav and Wasserblat, Moshe and Schwartz, Roy},
  journal={arXiv preprint arXiv:2405.04304},
  year={2024}
}

@article{li2025semi,
  title={Semi-clairvoyant scheduling of speculative decoding requests to minimize LLM inference latency},
  author={Li, Ruixiao and Chen, Fahao and Li, Peng},
  journal={arXiv preprint arXiv:2505.17074},
  year={2025}
}

@inproceedings{wu2025tetris,
  title={TETRIS: Optimal draft token selection for batch speculative decoding},
  author={Wu, Zhaoxuan and Zhou, Zijian and Verma, Arun and Prakash, Alok and Rus, Daniela and Low, Bryan Kian Hsiang},
  booktitle={Proceedings of the 63rd Annual Meeting of the Association for Computational Linguistics (Volume 1: Long Papers)},
  pages={33329--33345},
  year={2025}
}

@article{gautam2025token,
  title={Token-Driven GammaTune: Adaptive Calibration for Enhanced Speculative Decoding},
  author={Gautam, Aayush and Shrestha, Susav and Reddy, Narasimha},
  journal={arXiv preprint arXiv:2504.00030},
  year={2025}
}

@article{yang2025dsde,
  title={DSDE: Dynamic Speculative Decoding with KLD Stability for Real-World Serving},
  author={Yang, Mingyu and Choi, Jae-Young and Moon, Kihyo and Jang, Minsung and Jeon, Eunjoo},
  journal={arXiv preprint arXiv:2509.01083},
  year={2025}
}

@article{liao2026hsd,
  title={HSD: Training-Free Acceleration for Document Parsing Vision-Language Model with Hierarchical Speculative Decoding},
  author={Liao, Wenhui and Li, Hongliang and Xie, Pengyu and Cai, Xinyu and Shen, Yufan and Xin, Yi and Qin, Qi and Ye, Shenglong and Li, Tianbin and Hu, Ming and others},
  journal={arXiv preprint arXiv:2602.12957},
  year={2026}
}

@article{zhang2026sparrow,
  title={Sparrow: Text-Anchored Window Attention with Visual-Semantic Glimpsing for Speculative Decoding in Video LLMs},
  author={Zhang, Libo and Zhang, Zhaoning and Hong, Wangyang and Qiao, Peng and Li, Dongsheng},
  journal={arXiv preprint arXiv:2602.15318},
  year={2026}
}

@article{wang2025specprune,
  title={Specprune-vla: Accelerating vision-language-action models via action-aware self-speculative pruning},
  author={Wang, Hanzhen and Xu, Jiaming and Xiang, Yushun and Pan, Jiayi and Zhou, Yongkang and Li, Yong-Lu and Dai, Guohao},
  journal={arXiv preprint arXiv:2509.05614},
  year={2025}
}

@inproceedings{wang2025spec,
  title={Spec-vla: speculative decoding for vision-language-action models with relaxed acceptance},
  author={Wang, Songsheng and Yu, Rucheng and Yuan, Zhihang and Yu, Chao and Gao, Feng and Wang, Yu and Wong, Derek F},
  booktitle={Proceedings of the 2025 Conference on Empirical Methods in Natural Language Processing},
  pages={26916--26928},
  year={2025}
}

@article{huang2026edgesd,
  title={EdgeSD: Efficient Speculative Decoding with Vision-Decoding Disaggregation for MLLM Inference in Edge-Cloud Networks},
  author={Huang, Hualong and Zhan, Wenhan and Duan, Hancong and Peng, Kai and Min, Geyong and Zhao, Zijia and Zhao, Zitian and Ye, Yalan},
  journal={IEEE Transactions on Mobile Computing},
  year={2026},
  publisher={IEEE}
}

@article{lv2026hippo,
  title={HIPPO: Accelerating Video Large Language Models Inference via Holistic-aware Parallel Speculative Decoding},
  author={Lv, Qitan and Liu, Tianyu and Wu, Wen and Xu, Xuenan and Zhou, Bowen and Wu, Feng and Zhang, Chao},
  journal={arXiv preprint arXiv:2601.08273},
  year={2026}
}

@article{tong2026sage,
  title={SAGE: Accelerating Vision-Language Models via Entropy-Guided Adaptive Speculative Decoding},
  author={Tong, Yujia and Zhang, Tian and Wan, Yunyang and Lin, Kaiwei and Yuan, Jingling and Hu, Chuang},
  journal={arXiv preprint arXiv:2602.00523},
  year={2026}
}

@article{xie2025hivis,
  title={HiViS: Hiding Visual Tokens from the Drafter for Speculative Decoding in Vision-Language Models},
  author={Xie, Zhinan and Wang, Peisong and Qiu, Shuang and Cheng, Jian},
  journal={arXiv preprint arXiv:2509.23928},
  year={2025}
}

@article{lee2026tabed,
  title={TABED: Test-Time Adaptive Ensemble Drafting for Robust Speculative Decoding in LVLMs},
  author={Lee, Minjae and Kang, Wonjun and Ahn, Byeongkeun and Classen, Christian and Galim, Kevin and Oh, Seunghyuk and Yan, Minghao and Koo, Hyung Il and Lee, Kangwook},
  journal={arXiv preprint arXiv:2601.20357},
  year={2026}
}

@article{liu2025small,
  title={Small drafts, big verdict: Information-intensive visual reasoning via speculation},
  author={Liu, Yuhan and Qin, Lianhui and Wang, Shengjie},
  journal={arXiv preprint arXiv:2510.20812},
  year={2025}
}

@article{wang2025flash,
  title={FLASH: Latent-Aware Semi-Autoregressive Speculative Decoding for Multimodal Tasks},
  author={Wang, Zihua and Li, Ruibo and Du, Haozhe and Zhou, Joey Tianyi and Zhang, Yu and Yang, Xu},
  journal={arXiv preprint arXiv:2505.12728},
  year={2025}
}

@inproceedings{liu2025adaptive,
  title={Adaptive draft-verification for efficient large language model decoding},
  author={Liu, Xukun and Lei, Bowen and Zhang, Ruqi and Xu, Dongkuan DK},
  booktitle={Proceedings of the AAAI Conference on Artificial Intelligence},
  volume={39},
  number={23},
  pages={24668--24676},
  year={2025}
}

@inproceedings{ou2024lossless,
  title={Lossless acceleration of large language model via adaptive n-gram parallel decoding},
  author={Ou, Jie and Chen, Yueming and others},
  booktitle={Proceedings of the 2024 Conference of the North American Chapter of the Association for Computational Linguistics: Human Language Technologies (Volume 6: Industry Track)},
  pages={10--22},
  year={2024}
}

@article{agrawal2024adaedl,
  title={Adaedl: Early draft stopping for speculative decoding of large language models via an entropy-based lower bound on token acceptance probability},
  author={Agrawal, Sudhanshu and Jeon, Wonseok and Lee, Mingu},
  journal={arXiv preprint arXiv:2410.18351},
  year={2024}
}

@article{ankner2024hydra,
  title={Hydra: Sequentially-dependent draft heads for medusa decoding},
  author={Ankner, Zachary and Parthasarathy, Rishab and Nrusimha, Aniruddha and Rinard, Christopher and Ragan-Kelley, Jonathan and Brandon, William},
  journal={arXiv preprint arXiv:2402.05109},
  year={2024}
}

@article{huo2025spec,
  title={Spec-LLaVA: Accelerating Vision-Language Models with Dynamic Tree-Based Speculative Decoding},
  author={Huo, Mingxiao and Zhang, Jiayi and Wang, Hewei and Xu, Jinfeng and Chen, Zheyu and Tai, Huilin and Chen, Yijun},
  journal={arXiv preprint arXiv:2509.11961},
  year={2025}
}

@inproceedings{yang2025aasd,
  title={AASD: Accelerate Inference by Aligning Speculative Decoding in Multimodal Large Language Models},
  author={Yang, Chaoqun and Chen, Ran and Zhang, Muyang and Pang, Weiguang and Chen, Yuzhi and Xu, Rongtao and Fu, Kexue and Wang, Changwei and Gao, Longxiang},
  booktitle={2025 62nd ACM/IEEE Design Automation Conference (DAC)},
  pages={1--7},
  year={2025},
  organization={IEEE}
}

@inproceedings{
lee2024batch,
title={In-batch Ensemble Drafting: Robust Speculative Decoding for {LVLM}s},
author={Minjae Lee and Wonjun Kang and Byeongkeun Ahn and Christian Classen and Minghao Yan and Hyung Il Koo and Kangwook Lee},
booktitle={First Workshop on Scalable Optimization for Efficient and Adaptive Foundation Models},
year={2025},
url={https://openreview.net/forum?id=ffDhpmwqdu}
}

@inproceedings{gagrani2024speculative,
  title={On speculative decoding for multimodal large language models},
  author={Gagrani, Mukul and Goel, Raghavv and Jeon, Wonseok and Park, Junyoung and Lee, Mingu and Lott, Christopher},
  booktitle={Proceedings of the IEEE/CVF Conference on Computer Vision and Pattern Recognition},
  pages={8285--8289},
  year={2024}
}

@article{xiao2024parallelspec,
  title={Parallelspec: Parallel drafter for efficient speculative decoding},
  author={Xiao, Zilin and Zhang, Hongming and Ge, Tao and Ouyang, Siru and Ordonez, Vicente and Yu, Dong},
  journal={arXiv preprint arXiv:2410.05589},
  year={2024}
}

@inproceedings{zhao2024ouroboros,
  title={Ouroboros: Generating longer drafts phrase by phrase for faster speculative decoding},
  author={Zhao, Weilin and Huang, Yuxiang and Han, Xu and Xu, Wang and Xiao, Chaojun and Zhang, Xinrong and Fang, Yewei and Zhang, Kaihuo and Liu, Zhiyuan and Sun, Maosong},
  booktitle={Proceedings of the 2024 Conference on Empirical Methods in Natural Language Processing},
  pages={13378--13393},
  year={2024}
}

@article{yang2023inference,
  title={Inference with reference: Lossless acceleration of large language models},
  author={Yang, Nan and Ge, Tao and Wang, Liang and Jiao, Binxing and Jiang, Daxin and Yang, Linjun and Majumder, Rangan and Wei, Furu},
  journal={arXiv preprint arXiv:2304.04487},
  year={2023}
}

@article{zhou2023distillspec,
  title={Distillspec: Improving speculative decoding via knowledge distillation},
  author={Zhou, Yongchao and Lyu, Kaifeng and Rawat, Ankit Singh and Menon, Aditya Krishna and Rostamizadeh, Afshin and Kumar, Sanjiv and Kagy, Jean-Fran{\c{c}}ois and Agarwal, Rishabh},
  journal={arXiv preprint arXiv:2310.08461},
  year={2023}
}

@article{liu2023online,
  title={Online speculative decoding},
  author={Liu, Xiaoxuan and Hu, Lanxiang and Bailis, Peter and Cheung, Alvin and Deng, Zhijie and Stoica, Ion and Zhang, Hao},
  journal={arXiv preprint arXiv:2310.07177},
  year={2023}
}

@article{kim2023speculative,
  title={Speculative decoding with big little decoder},
  author={Kim, Sehoon and Mangalam, Karttikeya and Moon, Suhong and Malik, Jitendra and Mahoney, Michael W and Gholami, Amir and Keutzer, Kurt},
  journal={Advances in Neural Information Processing Systems},
  volume={36},
  pages={39236--39256},
  year={2023}
}

@article{stern2018blockwise,
  title={Blockwise parallel decoding for deep autoregressive models},
  author={Stern, Mitchell and Shazeer, Noam and Uszkoreit, Jakob},
  journal={Advances in Neural Information Processing Systems},
  volume={31},
  year={2018}
}

@inproceedings{xia2023speculative,
  title={Speculative decoding: Exploiting speculative execution for accelerating seq2seq generation},
  author={Xia, Heming and Ge, Tao and Wang, Peiyi and Chen, Si-Qing and Wei, Furu and Sui, Zhifang},
  booktitle={Findings of the Association for Computational Linguistics: EMNLP 2023},
  pages={3909--3925},
  year={2023}
}

@inproceedings{
sadhukhan2025magicdec,
title={MagicDec: Breaking the Latency-Throughput Tradeoff for Long Context Generation with Speculative Decoding},
author={Ranajoy Sadhukhan and Jian Chen and Zhuoming Chen and Vashisth Tiwari and Ruihang Lai and Jinyuan Shi and Ian En-Hsu Yen and Avner May and Tianqi Chen and Beidi Chen},
booktitle={The Thirteenth International Conference on Learning Representations},
year={2025},
url={https://openreview.net/forum?id=CS2JWaziYr}
}

@inproceedings{elhoushi2024layerskip,
  title={Layerskip: Enabling early exit inference and self-speculative decoding},
  author={Elhoushi, Mostafa and Shrivastava, Akshat and Liskovich, Diana and Hosmer, Basil and Wasti, Bram and Lai, Liangzhen and Mahmoud, Anas and Acun, Bilge and Agarwal, Saurabh and Roman, Ahmed and others},
  booktitle={Proceedings of the 62nd Annual Meeting of the Association for Computational Linguistics (Volume 1: Long Papers)},
  pages={12622--12642},
  year={2024}
}

@inproceedings{qiao2025ig,
    title = "{IG}-Pruning: Input-Guided Block Pruning for Large Language Models",
    author = "Qiao, Kangyu  and
      Zhang, Shaolei  and
      Feng, Yang",
    booktitle = "Proceedings of the 2025 Conference on Empirical Methods in Natural Language Processing",
    month = nov,
    year = "2025",
    publisher = "Association for Computational Linguistics",
    url = "https://aclanthology.org/2025.emnlp-main.537/",
    doi = "10.18653/v1/2025.emnlp-main.537",
    pages = "10618--10629",
    ISBN = "979-8-89176-332-6",
}

@inproceedings{xia2025swift,
  title={{SWIFT}: On-the-Fly Self-Speculative Decoding for {LLM} Inference Acceleration},
  author={Heming Xia and Yongqi Li and Jun Zhang and Cunxiao Du and Wenjie Li},
  booktitle={The Thirteenth International Conference on Learning Representations},
  year={2025},
  url={https://openreview.net/forum?id=EKJhH5D5wA}
}

@misc{kim2026multidrafter,
      title={Multi-Drafter Speculative Decoding with Alignment Feedback}, 
      author={Taehyeon Kim and Hojung Jung and Se-Young Yun},
      year={2026},
      eprint={2604.05417},
      archivePrefix={arXiv},
      primaryClass={cs.CL},
      url={https://arxiv.org/abs/2604.05417}, 
}

@inproceedings{
zhang2026learning,
title={Learning To Draft: Adaptive Speculative Decoding with Reinforcement Learning},
author={Jiebin Zhang and Zhenghan Yu and Liang Wang and Nan Yang and Eugene J. Yu and Zheng Li and Yifan Song and Dawei Zhu and Xingxing Zhang and Furu Wei and Sujian Li},
booktitle={The Fourteenth International Conference on Learning Representations},
year={2026},
url={https://openreview.net/forum?id=IK9cbzzXLt}
}

@misc{sridhar2025tapout,
      title={TapOut: A Bandit-Based Approach to Dynamic Speculative Decoding}, 
      author={Aditya Sridhar and Nish Sinnadurai and Sean Lie and Vithursan Thangarasa},
      year={2025},
      eprint={2511.02017},
      archivePrefix={arXiv},
      primaryClass={cs.LG},
      url={https://arxiv.org/abs/2511.02017}, 
}

@misc{zhang2025sparsetodense,
      title={Sparse-to-Dense: A Free Lunch for Lossless Acceleration of Video Understanding in LLMs}, 
      author={Xuan Zhang and Cunxiao Du and Sicheng Yu and Jiawei Wu and Fengzhuo Zhang and Wei Gao and Qian Liu},
      year={2025},
      eprint={2505.19155},
      archivePrefix={arXiv},
      primaryClass={cs.CV},
      url={https://arxiv.org/abs/2505.19155}, 
}

@InProceedings{Combes2014unimodal,
  title = 	 {Unimodal Bandits: Regret Lower Bounds and Optimal Algorithms},
  author = 	 {Combes, Richard and Proutiere, Alexandre},
  booktitle = 	 {Proceedings of the 31st International Conference on Machine Learning},
  pages = 	 {521--529},
  year = 	 {2014},
  volume = 	 {32},
  number =       {1},
  series = 	 {Proceedings of Machine Learning Research},
  month = 	 {22--24 Jun},
  publisher =    {PMLR}
}

@misc{hou2025banditspec,
      title={BanditSpec: Adaptive Speculative Decoding via Bandit Algorithms}, 
      author={Yunlong Hou and Fengzhuo Zhang and Cunxiao Du and Xuan Zhang and Jiachun Pan and Tianyu Pang and Chao Du and Vincent Y. F. Tan and Zhuoran Yang},
      year={2025},
      eprint={2505.15141},
      archivePrefix={arXiv},
      primaryClass={cs.LG},
      url={https://arxiv.org/abs/2505.15141}, 
}

@inproceedings{magureanu2014lipschitz,
  title={Lipschitz bandits: Regret lower bound and optimal algorithms},
  author={Magureanu, Stefan and Combes, Richard and Proutiere, Alexandre},
  booktitle={Conference on Learning Theory},
  pages={975--999},
  year={2014},
  organization={PMLR}
}

@inproceedings{cao2019nearly,
  title={Nearly Optimal Adaptive Procedure with Change Detection for Piecewise-stationary Bandit},
  author={Cao, Yang and Wen, Zheng and Kveton, Branislav and Xie, Yao},
  booktitle={Proceedings of the 22nd International Conference on Artificial Intelligence and Statistics},
  pages={418--427},
  year={2019},
  organization={PMLR}
}

@article{hou2024almost,
  title={Almost Minimax Optimal Best Arm Identification in Piecewise Stationary Linear Bandits},
  author={Hou, Y. and Tan, V. and Zhong, Z.},
  journal={Advances in Neural Information Processing Systems},
  volume={37},
  pages={128967--129041},
  year={2024}
}

@inproceedings{he2024rest,
    title = "{REST}: Retrieval-Based Speculative Decoding",
    author = "He, Zhenyu  and
      Zhong, Zexuan  and
      Cai, Tianle  and
      Lee, Jason  and
      He, Di",
    booktitle = "Proceedings of the 2024 Conference of the North American Chapter of the Association for Computational Linguistics: Human Language Technologies (Volume 1: Long Papers)",
    month = jun,
    year = "2024",
    publisher = "Association for Computational Linguistics",
    pages = "1582--1595",
}

@inproceedings{li2024eagle2, 
	author = {Yuhui Li and Fangyun Wei and Chao Zhang and Hongyang Zhang}, 
	title = {{EAGLE-2}: Faster Inference of Language Models with Dynamic Draft Trees}, 
	booktitle = {Empirical Methods in Natural Language Processing},
	year = {2024}
}

@inproceedings{leviathan2023fast,
  title={Fast inference from transformers via speculative decoding},
  author={Leviathan, Yaniv and Kalman, Matan and Matias, Yossi},
  booktitle={International Conference on Machine Learning},
  pages={19274--19286},
  year={2023},
  organization={PMLR}
}

@article{yin2024theoretical,
  title={A theoretical perspective for speculative decoding algorithm},
  author={Yin, Ming and Chen, Minshuo and Huang, Kaixuan and Wang, Mengdi},
  journal={arXiv preprint arXiv:2411.00841},
  year={2024}
}

@article{huang2024specdec++,
  title={SpecDec++: Boosting Speculative Decoding via Adaptive Candidate Lengths},
  author={Huang, Kaixuan and Guo, Xudong and Wang, Mengdi},
  journal={arXiv preprint arXiv:2405.19715},
  year={2024}
}

@article{chen2023accelerating,
  title={Accelerating large language model decoding with speculative sampling},
  author={Chen, Charlie and Borgeaud, Sebastian and Irving, Geoffrey and Lespiau, Jean-Baptiste and Sifre, Laurent and Jumper, John},
  journal={arXiv preprint arXiv:2302.01318},
  year={2023}
}

@article{antos2010active,
  title={Active learning in heteroscedastic noise},
  author={Antos, Andr{\'a}s and Grover, Varun and Szepesv{\'a}ri, Csaba},
  journal={Theoretical Computer Science},
  volume={411},
  number={29-30},
  pages={2712--2728},
  year={2010},
  publisher={Elsevier}
}

@inproceedings{Abbasi2011improved,
 author = {Abbasi-yadkori, Yasin and P\'{a}l, D\'{a}vid and Szepesv\'{a}ri, Csaba},
 booktitle = {Advances in Neural Information Processing Systems},
 pages = {2312-2320},
 title = {Improved Algorithms for Linear Stochastic Bandits},
 volume = {24},
 year = {2011}
}

@article{auer2002finite,
  title={Finite-time Analysis of the Multi-armed Bandit Problem},
  author={Auer, P and Cesa-Bianchi, N and Fischer, Paul},
  journal={Machine Learning},
  volume={47},
  pages={235--256},
  year={2002}
}

@book{lattimore2020bandit,
  title={Bandit algorithms},
  author={Lattimore, Tor and Szepesv{\'a}ri, Csaba},
  year={2020},
  publisher={Cambridge University Press}
}

@article{miao2023SpecInferAG,
  title={SpecInfer: Accelerating Generative LLM Serving with Speculative Inference and Token Tree Verification},
  author={Xupeng Miao and Gabriele Oliaro and Zhihao Zhang and Xinhao Cheng and Zeyu Wang and Rae Ying Yee Wong and Zhuoming Chen and Daiyaan Arfeen and Reyna Abhyankar and Zhihao Jia},
  journal={ArXiv},
  year={2023},
  volume={abs/2305.09781},
  url={https://api.semanticscholar.org/CorpusID:258740799}
}

@article{agrawal2017near,
  title={Near-optimal regret bounds for thompson sampling},
  author={Agrawal, Shipra and Goyal, Navin},
  journal={Journal of the ACM (JACM)},
  volume={64},
  number={5},
  pages={1--24},
  year={2017},
  publisher={ACM New York, NY, USA}
}

@InProceedings{agrawal2012analysis,
  title = 	 {Analysis of Thompson Sampling for the Multi-armed Bandit Problem},
  author = 	 {Agrawal, Shipra and Goyal, Navin},
  booktitle = 	 {Proceedings of the 25th Annual Conference on Learning Theory},
  pages = 	 {39.1--39.26},
  year = 	 {2012},
  editor = 	 {Mannor, Shie and Srebro, Nathan and Williamson, Robert C.},
  volume = 	 {23},
  series = 	 {Proceedings of Machine Learning Research},
  address = 	 {Edinburgh, Scotland},
  month = 	 {25--27 Jun},
  publisher =    {PMLR},
  url = 	 {https://proceedings.mlr.press/v23/agrawal12.html},
}

@article{robbins1952some,
  title={Some aspects of the sequential design of experiments},
  author={Robbins, Herbert},
  year={1952},
  journal = {Bulletin of the American Mathematical Society},
  volume={58},
  pages={527--535},
}

@article{lai1985asymptotically,
  title={Asymptotically efficient adaptive allocation rules},
  author={Lai, Tze Leung and Robbins, Herbert},
  journal={Advances in applied mathematics},
  volume={6},
  number={1},
  pages={4--22},
  year={1985},
  publisher={Academic Press}
}

@article{cai2024medusa,
  title={Medusa: Simple llm inference acceleration framework with multiple decoding heads},
  author={Cai, T. and Li, Y. and Geng, Z. and Peng, H. and Lee, J. D. and Chen, D. and Dao, T.},
  journal={arXiv preprint arXiv:2401.10774},
  year={2024}
}

@article{li2024eagle,
  title={Eagle: Speculative sampling requires rethinking feature uncertainty},
  author={Li, Yuhui and Wei, Fangyun and Zhang, Chao and Zhang, Hongyang},
  journal={arXiv preprint arXiv:2401.15077},
  year={2024}
}

@misc{qwen3vltechnicalreport,
      title={Qwen3-VL Technical Report},
      author={Shuai Bai and Yuxuan Cai and Ruizhe Chen and Keqin Chen and Xionghui Chen and Zesen Cheng and Lianghao Deng and Wei Ding and Chang Gao and Chunjiang Ge and Wenbin Ge and Zhifang Guo and Qidong Huang and Jie Huang and Fei Huang and Binyuan Hui and Shutong Jiang and Zhaohai Li and Mingsheng Li and Mei Li and Kaixin Li and Zicheng Lin and Junyang Lin and Xuejing Liu and Jiawei Liu and Chenglong Liu and Yang Liu and Dayiheng Liu and Shixuan Liu and Dunjie Lu and Ruilin Luo and Chenxu Lv and Rui Men and Lingchen Meng and Xuancheng Ren and Xingzhang Ren and Sibo Song and Yuchong Sun and Jun Tang and Jianhong Tu and Jianqiang Wan and Peng Wang and Pengfei Wang and Qiuyue Wang and Yuxuan Wang and Tianbao Xie and Yiheng Xu and Haiyang Xu and Jin Xu and Zhibo Yang and Mingkun Yang and Jianxin Yang and An Yang and Bowen Yu and Fei Zhang and Hang Zhang and Xi Zhang and Bo Zheng and Humen Zhong and Jingren Zhou and Fan Zhou and Jing Zhou and Yuanzhi Zhu and Ke Zhu},
      year={2025},
      eprint={2511.21631},
      archivePrefix={arXiv},
      primaryClass={cs.AI},
      url={https://arxiv.org/abs/2511.21631},
}

@misc{videomme,
      title={Video-MME: The First-Ever Comprehensive Evaluation Benchmark of Multi-modal LLMs in Video Analysis}, 
      author={Chaoyou Fu and Yuhan Dai and Yongdong Luo and Lei Li and Shuhuai Ren and Renrui Zhang and Zihan Wang and Chenyu Zhou and Yunhang Shen and Mengdan Zhang and Peixian Chen and Yanwei Li and Shaohui Lin and Sirui Zhao and Ke Li and Tong Xu and Xiawu Zheng and Enhong Chen and Caifeng Shan and Ran He and Xing Sun},
      year={2025},
      eprint={2405.21075},
      archivePrefix={arXiv},
      primaryClass={cs.CV},
      url={https://arxiv.org/abs/2405.21075}, 
}

@misc{arkitscenes,
      title={ARKitScenes: A Diverse Real-World Dataset For 3D Indoor Scene Understanding Using Mobile RGB-D Data}, 
      author={Gilad Baruch and Zhuoyuan Chen and Afshin Dehghan and Tal Dimry and Yuri Feigin and Peter Fu and Thomas Gebauer and Brandon Joffe and Daniel Kurz and Arik Schwartz and Elad Shulman},
      year={2021},
      eprint={2111.08897},
      archivePrefix={arXiv},
      primaryClass={cs.AI},
      url={https://arxiv.org/abs/2111.08897}, 
}

@misc{writingbench,
      title={WritingBench: A Comprehensive Benchmark for Generative Writing}, 
      author={Yuning Wu and Jiahao Mei and Ming Yan and Chenliang Li and Shaopeng Lai and Yuran Ren and Zijia Wang and Ji Zhang and Mengyue Wu and Qin Jin and Fei Huang},
      year={2025},
      eprint={2503.05244},
      archivePrefix={arXiv},
      primaryClass={cs.CL},
      url={https://arxiv.org/abs/2503.05244}, 
}

@misc{li2025eagle3,
      title={EAGLE-3: Scaling up Inference Acceleration of Large Language Models via Training-Time Test}, 
      author={Yuhui Li and Fangyun Wei and Chao Zhang and Hongyang Zhang},
      year={2025},
      eprint={2503.01840},
      archivePrefix={arXiv},
      primaryClass={cs.CL},
      url={https://arxiv.org/abs/2503.01840}, 
}

@inproceedings{Zhang_2024selfspec,
   title={Draft\& Verify: Lossless Large Language Model Acceleration via Self-Speculative Decoding},
   url={http://dx.doi.org/10.18653/v1/2024.acl-long.607},
   DOI={10.18653/v1/2024.acl-long.607},
   booktitle={Proceedings of the 62nd Annual Meeting of the Association for Computational Linguistics (Volume 1: Long Papers)},
   publisher={Association for Computational Linguistics},
   author={Zhang, Jun and Wang, Jue and Li, Huan and Shou, Lidan and Chen, Ke and Chen, Gang and Mehrotra, Sharad},
   year={2024},
   pages={11263–11282} }

@misc{bajpai2025fastvlmselfspec,
      title={FastVLM: Self-Speculative Decoding for Fast Vision-Language Model Inference}, 
      author={Divya Jyoti Bajpai and Manjesh Kumar Hanawal},
      year={2025},
      eprint={2510.22641},
      archivePrefix={arXiv},
      primaryClass={cs.LG},
      url={https://arxiv.org/abs/2510.22641}, 
}

@misc{huang2025specvlmfastspeculativedecoding,
      title={SpecVLM: Fast Speculative Decoding in Vision-Language Models}, 
      author={Haiduo Huang and Fuwei Yang and Zhenhua Liu and Xuanwu Yin and Dong Li and Pengju Ren and Emad Barsoum},
      year={2025},
      eprint={2509.11815},
      archivePrefix={arXiv},
      primaryClass={cs.CV},
      url={https://arxiv.org/abs/2509.11815}, 
}

@misc{kang2025vispec,
      title={ViSpec: Accelerating Vision-Language Models with Vision-Aware Speculative Decoding}, 
      author={Jialiang Kang and Han Shu and Wenshuo Li and Yingjie Zhai and Xinghao Chen},
      year={2025},
      eprint={2509.15235},
      archivePrefix={arXiv},
      primaryClass={cs.CV},
      url={https://arxiv.org/abs/2509.15235}, 
}

@inproceedings{liu2025dropinadapspec,
    title = "A Drop-In Solution for On-the-Fly Adaptation of Speculative Decoding in Large Language Models",
    author = "Liu, Jiesong  and
      Park, Brian  and
      Shen, Xipeng",
    editor = "Che, Wanxiang  and
      Nabende, Joyce  and
      Shutova, Ekaterina  and
      Pilehvar, Mohammad Taher",
    booktitle = "Proceedings of the 63rd Annual Meeting of the Association for Computational Linguistics (Volume 1: Long Papers)",
    month = jul,
    year = "2025",
    address = "Vienna, Austria",
    publisher = "Association for Computational Linguistics",
    url = "https://aclanthology.org/2025.acl-long.482/",
    doi = "10.18653/v1/2025.acl-long.482",
    pages = "9778--9794",
    ISBN = "979-8-89176-251-0"
}

@misc{fang2024mmbenchvideo,
      title={MMBench-Video: A Long-Form Multi-Shot Benchmark for Holistic Video Understanding}, 
      author={Xinyu Fang and Kangrui Mao and Haodong Duan and Xiangyu Zhao and Yining Li and Dahua Lin and Kai Chen},
      year={2024},
      eprint={2406.14515},
      archivePrefix={arXiv},
      primaryClass={cs.CV},
      url={https://arxiv.org/abs/2406.14515}, 
}

@inproceedings{chen2025clasp,
    title = "{CL}a{S}p: In-Context Layer Skip for Self-Speculative Decoding",
    author = "Chen, Longze  and
      Shan, Renke  and
      Wang, Huiming  and
      Wang, Lu  and
      Liu, Ziqiang  and
      Luo, Run  and
      Wang, Jiawei  and
      Alinejad-Rokny, Hamid  and
      Yang, Min",
    booktitle = "Proceedings of the 63rd Annual Meeting of the Association for Computational Linguistics (Volume 1: Long Papers)",
    month = jul,
    year = "2025",
    address = "Vienna, Austria",
    publisher = "Association for Computational Linguistics",
    url = "https://aclanthology.org/2025.acl-long.1525/",
    doi = "10.18653/v1/2025.acl-long.1525",
    pages = "31608--31618",
    ISBN = "979-8-89176-251-0",
}

@misc{cha2026knapspec,
      title={KnapSpec: Self-Speculative Decoding via Adaptive Layer Selection as a Knapsack Problem}, 
      author={Seongjin Cha and Gyuwan Kim and Dongsu Han and Tao Yang and Insu Han},
      year={2026},
      eprint={2602.20217},
      archivePrefix={arXiv},
      primaryClass={cs.LG},
      url={https://arxiv.org/abs/2602.20217}, 
}

@inproceedings{
    ganesan2025massv,
    title={{MASSV}: Multimodal Adaptation and Self-Data Distillation for Speculative Decoding of Vision-Language Models},
    author={Mugilan Ganesan and Shane Segal and Ankur Aggarwal and Nish Sinnadurai and Sean Lie and Vithursan Thangarasa},
    booktitle={ES-FoMo III: 3rd Workshop on Efficient Systems for Foundation Models},
    year={2025},
    url={https://openreview.net/forum?id=ukDi9JyaL3}
}

@misc{lin2025speculative,
      title={Speculative Decoding Reimagined for Multimodal Large Language Models}, 
      author={Luxi Lin and Zhihang Lin and Zhanpeng Zeng and Rongrong Ji},
      year={2025},
      eprint={2505.14260},
      archivePrefix={arXiv},
      primaryClass={cs.CV},
      url={https://arxiv.org/abs/2505.14260}, 
}

@inproceedings{
    hu2026dream,
    title={{DREAM}: Drafting with Refined Target Features and Entropy-Adaptive Cross-Attention Fusion for Multimodal Speculative Decoding},
    author={Yunhai Hu and Tianhua Xia and Zining Liu and Rahul Raman and Xingyu Liu and BO BAO and Eric Sather and Vithursan Thangarasa and Sai Qian Zhang},
    booktitle={The Thirty-ninth Annual Conference on Neural Information Processing Systems},
    year={2025},
    url={https://openreview.net/forum?id=M5jz47umjR}
}

@inproceedings{ji2025specvlm,
    title = "{S}pec{VLM}: Enhancing Speculative Decoding of Video {LLM}s via Verifier-Guided Token Pruning",
    author = "Ji, Yicheng  and
      Zhang, Jun  and
      Xia, Heming  and
      Chen, Jinpeng  and
      Shou, Lidan  and
      Chen, Gang  and
      Li, Huan",
    editor = "Christodoulopoulos, Christos  and
      Chakraborty, Tanmoy  and
      Rose, Carolyn  and
      Peng, Violet",
    booktitle = "Proceedings of the 2025 Conference on Empirical Methods in Natural Language Processing",
    month = nov,
    year = "2025",
    address = "Suzhou, China",
    publisher = "Association for Computational Linguistics",
    url = "https://aclanthology.org/2025.emnlp-main.366/",
    doi = "10.18653/v1/2025.emnlp-main.366",
    pages = "7205--7219",
    ISBN = "979-8-89176-332-6",
}

@inproceedings{
    liu2025pearl,
    title={{PEARL}: Parallel Speculative Decoding with Adaptive Draft Length},
    author={Tianyu Liu and Yun Li and Qitan Lv and Kai Liu and Jianchen Zhu and Winston Hu and Xiao Sun},
    booktitle={The Thirteenth International Conference on Learning Representations},
    year={2025},
    url={https://openreview.net/forum?id=QOXrVMiHGK}
}

@inproceedings{zhang2025draft,
    title = "Draft Model Knows When to Stop: Self-Verification Speculative Decoding for Long-Form Generation",
    author = "Zhang, Ziyin  and
      Xu, Jiahao  and
      Liang, Tian  and
      Chen, Xingyu  and
      He, Zhiwei  and
      Wang, Rui  and
      Tu, Zhaopeng",
    editor = "Christodoulopoulos, Christos  and
      Chakraborty, Tanmoy  and
      Rose, Carolyn  and
      Peng, Violet",
    booktitle = "Proceedings of the 2025 Conference on Empirical Methods in Natural Language Processing",
    month = nov,
    year = "2025",
    address = "Suzhou, China",
    publisher = "Association for Computational Linguistics",
    url = "https://aclanthology.org/2025.emnlp-main.844/",
    doi = "10.18653/v1/2025.emnlp-main.844",
    pages = "16685--16697",
    ISBN = "979-8-89176-332-6",
}

@article{wang2025opt,
    title = "{OPT}-Tree: Speculative Decoding with Adaptive Draft Tree Structure",
    author = "Wang, Jikai  and
      Su, Yi  and
      Li, Juntao  and
      Xia, Qingrong  and
      Ye, Zi  and
      Duan, Xinyu  and
      Wang, Zhefeng  and
      Zhang, Min",
    journal = "Transactions of the Association for Computational Linguistics",
    volume = "13",
    year = "2025",
    address = "Cambridge, MA",
    publisher = "MIT Press",
    url = "https://aclanthology.org/2025.tacl-1.8/",
    doi = "10.1162/tacl_a_00735",
    pages = "188--199",
}

@article{xiong2026dyspec,
	author = {Xiong, Yunfan and Zhang, Ruoyu and Li, Yanzeng and Zou, Lei},
	date = {2025/05/08},
	doi = {10.1007/s11280-025-01344-0},
	id = {Xiong2025},
	isbn = {1573-1413},
	journal = {World Wide Web},
	number = {3},
	pages = {36},
	title = {DySpec: Faster speculative decoding with dynamic token tree structure},
	url = {https://doi.org/10.1007/s11280-025-01344-0},
	volume = {28},
	year = {2025}}

@misc{qwen2.5vltechnicalreport,
      title={Qwen2.5-VL Technical Report}, 
      author={Shuai Bai and Keqin Chen and Xuejing Liu and Jialin Wang and Wenbin Ge and Sibo Song and Kai Dang and Peng Wang and Shijie Wang and Jun Tang and Humen Zhong and Yuanzhi Zhu and Mingkun Yang and Zhaohai Li and Jianqiang Wan and Pengfei Wang and Wei Ding and Zheren Fu and Yiheng Xu and Jiabo Ye and Xi Zhang and Tianbao Xie and Zesen Cheng and Hang Zhang and Zhibo Yang and Haiyang Xu and Junyang Lin},
      year={2025},
      eprint={2502.13923},
      archivePrefix={arXiv},
      primaryClass={cs.CV},
      url={https://arxiv.org/abs/2502.13923}, 
}

@misc{liu2024llavanext,
    title={LLaVA-NeXT: Improved reasoning, OCR, and world knowledge},
    url={https://llava-vl.github.io/blog/2024-01-30-llava-next/},
    author={Liu, Haotian and Li, Chunyuan and Li, Yuheng and Li, Bo and Zhang, Yuanhan and Shen, Sheng and Lee, Yong Jae},
    month={January},
    year={2024}
}

@misc{aime25,
    title = {{AIME 2025} Dataset},
    author = {{OpenCompass}},
    year = {2025},
    url = {https://huggingface.co/datasets/opencompass/AIME2025},
    note = {Problems from the American Invitational Mathematics Examination 2025-I and 2025-II},
}

@INPROCEEDINGS{mvbench2024,
  author={Li, Kunchang and Wang, Yali and He, Yinan and Li, Yizhuo and Wang, Yi and Liu, Yi and Wang, Zun and Xu, Jilan and Chen, Guo and Lou, Ping and Wang, Limin and Qiao, Yu},
  booktitle={2024 IEEE/CVF Conference on Computer Vision and Pattern Recognition (CVPR)}, 
  title={MVBench: A Comprehensive Multi-modal Video Understanding Benchmark}, 
  year={2024},
  volume={},
  number={},
  pages={22195-22206},
  doi={10.1109/CVPR52733.2024.02095}}

@software{AngelSlim2025,
    title={{AngelSlim}},
    author={Tencent AngelSlim Project Contributors},
    year={2025},
    month={6},
    url={https://github.com/Tencent/AngelSlim},
}

@inproceedings{kveton2014matroid,
author = {Kveton, Branislav and Wen, Zheng and Ashkan, Azin and Eydgahi, Hoda and Eriksson, Brian},
title = {Matroid bandits: fast combinatorial optimization with learning},
year = {2014},
booktitle = {Proceedings of the Thirtieth Conference on Uncertainty in Artificial Intelligence},
pages = {420–429},
numpages = {10}
}

@InProceedings{hou2023probably,
  title = 	 {Probably Anytime-Safe Stochastic Combinatorial Semi-Bandits},
  author =       {Hou, Yunlong and Tan, Vincent Y. F. and Zhong, Zixin},
  booktitle = 	 {Proceedings of the 40th International Conference on Machine Learning},
  pages = 	 {13353--13409},
  year = 	 {2023},
  volume = 	 {202},
  series = 	 {Proceedings of Machine Learning Research},
  month = 	 {23--29 Jul},
  publisher =    {PMLR},
}

%%%%%%%%%%%%%%%%%%%%%%%%%%%%%%%%%%%%%%%%%%%%%%%%%%%%%%%%%%%%
\newpage
\appendix
\section{Related Work}\label{app:related}

\textbf{Speculative decoding for vision-language models.}
Speculative decoding was introduced to accelerate autoregressive generation, with most early methods focusing on text-only \acp{llm}~\citep{leviathan2023fast,chen2023accelerating}. As autoregressive modeling becomes increasingly prevalent in multimodal generation, extending speculative decoding to multimodal models has attracted growing attention~\citep{qwen3vltechnicalreport,wang2024emu3,wang2025error,li2024enhancing}. However, applying speculative decoding to \acp{vlm} presents distinct challenges, including long visual-token prefixes, expensive prefill and verification, and modality-dependent discrepancies between draft and target models.
% Most speculative decoding methods target text-only \acp{llm}, whereas \acp{vlm} face long visual prefixes, costly prefill and verification, and modality-sensitive draft--target alignment.
Multimodal studies examine direct speculation and LVLM drafting~\citep{gagrani2024speculative,lee2024batch,li2026annealed}, multimodal drafters, visual adapters, and alignment objectives~\citep{bajpai2025fastvlmselfspec,huang2025specvlmfastspeculativedecoding,ganesan2025massv}, dynamic trees and test-time drafting~\citep{lin2025speculative,yang2025aasd,huo2025spec}, and visual-token hiding, latent-aware speculation, or ensemble drafting~\citep{xie2025hivis,wang2025flash,liu2025small,lee2026tabed,tong2026sage}. Related extensions accelerate document \acp{vlm}, video \acp{llm}, edge-cloud MLLM, and VLA models via hierarchical drafting, token pruning, sparse-to-dense decoding, adaptive visual features, disaggregation, relaxed acceptance, or action-aware pruning~\citep{kang2025vispec,hu2026dream,zhang2025sparsetodense,ji2025specvlm,liao2026hsd,lv2026hippo,huang2026edgesd,wang2025spec,wang2025specprune,zhang2026sparrow}. Many require trained multimodal drafters, specialized visual modules, or modality-specific token processing. In contrast, our method builds the drafter directly from the target \ac{vlm} via layer skipping, or token pruning, requires no auxiliary training, and adapts the skip budget prompt by prompt. 

\textbf{Adaptive speculative decoding.}
Recent adaptive methods adjust candidate length, stopping rules, or draft--verify schedules~\citep{huang2024specdec++,liu2025pearl,zhang2025draft}, use bandit- or entropy-based stopping~\citep{sridhar2025tapout,agrawal2024adaedl}, adaptive n-gram or draft-verification schemes~\citep{ou2024lossless,liu2025adaptive}, and dynamic lookahead or calibration~\citep{mamou2024dynamic,yang2025dsde,gautam2025token}. Other works adapt draft trees or branching structures~\citep{wang2025opt,xiong2026dyspec}, tune serving-level speculation windows, batching, scheduling, or distributed execution~\citep{liu2025dropinadapspec,wu2025tetris,li2025semi,tran2025goodspeed}, or modify verification and drafting objectives via alignment-aware or diffusion-based designs~\citep{wang2025alignment,li2025diffuspec}. Closest to our setting, BanditSpec~\citep{hou2025banditspec} casts hyperparameter selection as a bandit problem, but treats configurations as independent arms and optimizes stopping-time regret. Not-a-Bandit~\citep{liu2026notabandit} uses full-information feedback when unchosen drafters are cheap to evaluate, which is costly for layer-skipped \acp{vlm}. In contrast, \ourmethod{} uses partial feedback, optimizes wall-time regret, and exploits the ordered single-peak structure of the skipping path by exploring near the current leader and its neighbors.

\textbf{Multi-Armed Bandit.} The multi-armed bandit problem models the exploration-exploitation trade-off in decision making~\citep{robbins1952some,lai1985asymptotically,lattimore2020bandit,auer2002finite}.
For the stochastic K-armed setup, the most commonly used algorithms include UCB algorithm~\citep{auer2002finite} and Thompson Sampling~\citep{agrawal2012analysis,agrawal2017near}. The literature also introduces additional structures to the arm set, for instance, linear bandits~\citep{Abbasi2011improved}, single-peak or unimodal bandits~\citep{Combes2014unimodal,yu2011Unimodal}, Lipschitz bandits~\citep{magureanu2014lipschitz}.

\section{More Applications and Future Directions}\label{app:more_application}
\textbf{Other self-speculative schemes.} While we focus on the layer-skip-style (and input-pruning-style) speculative decoding~\citep{xia2025swift,elhoushi2024layerskip,chen2025clasp}, our method can also be extended to other self-speculative decoding methods, as long as the single-peak structure holds in the arm sets.

For instance, Video Large Language Models suffer from high inference latency in \emph{long} video processing due to their auto-regressive decoding mechanism. In previous work, \citet{zhang2025sparsetodense} propose to use a sparse module that rapidly generates speculative tokens using efficient top-K attention, and a dense module that verifies these tokens in parallel via full self-attention.
Specifically, they propose to use a draft model, which is obtained by simply replacing the dense attention modules with sparse ones in the target model. Using a sparse attention module reduces the time required to read the selected KV cache during attention, which can be particularly advantageous when KV cache loading time dominates the forward pass for the target model. This method has achieved over $1.9\times$ wall time speed-up compared to the vanilla decoding method, and is better than the other speculative decoding methods~\citep{zhang2025sparsetodense}.

However, \citet{zhang2025sparsetodense} adopt a \emph{fixed} number of $1024$ KV cache pairs, which lacks adaptivity in the face of heterogeneous tasks. 
We expect that when the input task is easier, a few KV cache pairs are enough for the acceleration; when the task is hard, more KV cache pairs can better enhance the inference.
So the acceptance rate shows a diminishing return structure. In addition, the KV cache loading time is linear in the KV cache pairs when dealing with a long context. Thus, the time cost is also an affine function in the KV cache pairs. 
All in all, this problem is likely to have the same single-peak structure as we introduced in Section~\ref{sec:problem_formulation}.

\textbf{Joint adaptation of generation hyperparameters.}
Broadly speaking, \spsqec{} suggests a way to adapt hyperparameters online during LLM generation, by exploiting single-peak throughput structure.
The draft-length ablation in Appendix~\ref{app:ablation_draft_length} also shows that performance tends to peak at an intermediate draft length, motivating joint adaptation of the draft model configuration and draft length.
A natural extension is to alternate between adapting the layer-skip ratio and the draft length, holding the other parameter fixed during each stage and using accepted-token and wall-time feedback.
When throughput is single-peaked in each parameter with the other fixed, this approach could exploit the structure of both one-dimensional problems.
The same idea could extend to additional generation hyperparameters.
We leave the analysis of these alternating updates and their global optimality guarantees to future work.

\textbf{Beyond the single-peak structure.}
Identifying the optimal configuration of retained components (e.g., layers) among all candidates is challenging due to the combinatorial size of the configuration space. Existing methods provide principled approaches for constructing an ordered sequence of draft models, allowing us to exploit the resulting single-peak structure for efficient inference. Beyond this setting, richer structural properties of the draft-model space may also be considered, such as those studied in matroid bandits~\citep{kveton2014matroid} and combinatorial semi-bandits~\citep{hou2023probably}. Validating such structural assumptions and designing algorithms that explicitly exploit them could further improve inference efficiency. We leave this as future work.

% We formulate the KV cache selection problem as a bandit problem. We regard an arm as $s\in[0,1]:=\calI$, which indicates $s$ proportion of the KV cache pairs will be adopted in the drafting step. 
% As one may expect, using more KV cache in the draft method can yield draft tokens that are more likely to be accepted by the target model. 
% However, this gain comes at the cost of more KV cache loading time for the draft model. For instance, when $s=1$ meaning using full KV cache, then all draft tokens will be accepted but the drafting time is exactly the time for $L$ forward passes of the target model, which is non-negligible.
% This is in sharp contrast with the literature, \citet{hou2025banditspec}, where they assume the drafting time can be omitted.
\section{Proofs of Main Results}\label{app:proof_main}

\subsection{Proof of Theorem~\ref{thm:spsqec}}

For simplicity, we use the shorthand notation $\ST$ for $\ST(\alg,\prompt,\Lambda)$, and omit the condition prompt $\prompt$ in this proof.

\textbf{Step 1}
We first decompose the wall time:
\begin{align}
    \bbE[\WT_{\ST}]
    =
    \bbE\bigg[\sum_{t=1}^{\ST}c_{I_t}\bigg]
    =
    \sum_{i\in[K]}c_i\cdot \bbE\big[T_i(\ST)\big],
\end{align}
where we adopt Doob's optional stopping Theorem (Lemma~\ref{lem:DoobOptStopping}) and $T_i(t):=\sum_{s=1}^t \mathbbm{1}\{I_s=i\}.$
The wall-time regret can be simplified as 
\begin{align}
    \WTReg(\Lambda,\prompt)
    = 
    \sum_{i\in[K]}c_i\cdot \bbE\big[T_i(\ST(\alg))\big] - c_{i^*}\cdot \bbE\big[\ST(\alg_{i^*})\big]
\end{align}
By the unbiasedness of speculative decoding~\citep{yin2024theoretical,hou2025banditspec,leviathan2023fast}, i.e., $\prompt_{\ST(\alg)}=\prompt_{\textrm{target}}$ no matter how the draft model is chosen by $\alg$. Furthermore, 
\begin{align}
    \bbE\big[\len(\prompt_{\ST(\alg)})-\len(\prompt_0)\big]
    =
    \bbE\bigg[\sum_{t=1}^{\ST(\alg)}\len(x_{I_t,t})\bigg]
    =
    \sum_{i\in[K]} \mu_i \cdot \bbE\big[T_i(\ST(\alg))\big]
\end{align}

Therefore,
\begin{align}\label{equ:regret_decomposition}
    &\sum_{i\in[K]}c_i\cdot \bbE\big[T_i(\ST(\alg))\big] - c_{i^*}\cdot \bbE\big[\ST(\alg_{i^*})\big]
    \\
    &\quad=
    \sum_{i\in[K]}c_i\cdot \bbE\big[T_i(\ST(\alg))\big] - \frac{\mu_{i^*}}{\nu_{i^*}}\cdot \bbE\big[\ST(\alg_{i^*})\big]
    \\
    &\quad=
    \sum_{i\in[K]}c_i\cdot \bbE\big[T_i(\ST(\alg))\big] - \sum_{i\in[K]}\frac{\mu_i}{\nu_{i^*}}\cdot \bbE\big[T_i(\ST(\alg))\big]
    \\
    &\quad=
    \sum_{i\in[K]}\mu_i(\frac{1}{\nu_{i}}-\frac{1}{\nu_{i^*}})\cdot \bbE\big[T_i(\ST(\alg))\big].
\end{align}
Hence, we only need to bound $\bbE\big[T_i(\ST(\alg))\big] $ in the following steps.

\textbf{Step 2}
For any $i, j\in[K]$, we define $\tilT_{j,i}(t):= \sum_{s=K}^{t}\mathbbm{1}\{L_s=i,I_{s+1}=j\}$, which is the number of times candidate $i$ is the leader and candidate $j$ is adopted up to time step $t$.
We can decompose 
\begin{align}
    \bbE\big[T_i(\ST)\big] &\leq 1+ \bbE\bigg[\sum_{t=K}^{\ST-1} \mathbbm{1}\{I_{t+1}=i\}\bigg]
    \\
    &=1+
    \underbrace{\bbE\bigg[\sum_{t=K}^{\ST-1} \mathbbm{1}\{I_{t+1}=i,L_t\neq i^*\}\bigg]}_{(\clubsuit)}
    +
    \underbrace{\bbE\bigg[\sum_{t=K}^{\ST-1} \mathbbm{1}\{I_{t+1}=i,L_t=i^*\}\bigg]}_{(\spadesuit)}
\end{align}
According to our algorithm design, when $L_t=i^*$, $I_{t+1}$ must be in the neighborhood of $i^*$. Thus, the analysis of $(\spadesuit)$ is similar to the standard Multi-Armed Bandits, and would yield regret of order $O(\ln \bbE[\ST])$. In contrast, we will prove that $(\clubsuit)$ is of order $O(1)$ thanks to the unimodal structure of the candidate arms, which is of smaller order compared to $(\spadesuit)$.
We prove the above two claims in the following paragraph.

\noindent
\underline{Bound $(\clubsuit)$}

We note that 
\begin{align}
    (\clubsuit)
    = 
    \sum_{j\neq i^* }\bbE\bigg[\sum_{t=K}^{\ST-1} \mathbbm{1}\{I_{t+1}=i,L_t=j\}\bigg]
    \leq
    \sum_{i\neq i^* }\bbE\bigg[\tilT_i(\ST-1)\bigg].
\end{align}
Therefore, we only need to bound $\bbE\bigg[\tilT_i(\ST)\bigg]$.
Let $i$ be any suboptimal candidate and $i^\dagger:=\argmax_{j=i\pm 1}\nu_j$. We further define two complementary events:
\begin{align}
    A_{\varepsilon}:=\{t<\tau:L_t=i, T_{i^\dagger}(t)\geq \varepsilon\cdot \tilT_i(t)\}
    \quad\mbox{and}\quad
    B_{\varepsilon}:=\{t<\tau:L_t=i, T_{i^\dagger}(t)< \varepsilon\cdot \tilT_i(t)\},
\end{align}
so that 
\begin{align}
    \bbE[\tilT_i(\tau-1)]\leq \bbE[|A_\varepsilon|+|B_\varepsilon|].
\end{align}
Under event $A_\varepsilon$, both $i$ and $i^\dagger$ are pulled at least $\varepsilon\tilT_i(t)$ times. If $i$ is the leader most of the time, then we can distinguish these two arms with high probability.
Under event $B_\varepsilon$, we show that it can happen with small probability, as there is a better candidate in the neighborhood of $i$, due to the unimodal structure.

$\bullet$ Under event $A_\varepsilon$

% If $t\in A_\varepsilon$, then $L_t=i$ and $\frac{\hmu_{i}(t)}{c_i}>\frac{\hmu_{i^*}(t)}{c_{i^*}}$. Furthermore, we have (1) $T_i(t)\geq \frac{\tilT_i(t)}{3}$; (2) $T_{i^\dagger}(t)\geq \varepsilon\tilT_i(t)$.
% Denote $s:=\tilT_i(t)$, by using the following Lemma~\ref{lem:A_error_bound}, we obtain
% \begin{align}
%     \bbP\bigg(\frac{\hmu_{i}(t)}{c_i}>\frac{\hmu_{i^\dagger}(t)}{c_{i^\dagger}}\bigg)
%     &\leq
%     \bbP\bigg(\frac{\hmu_{i}(t)}{c_i}-\nu_i\geq \frac{\Delta^\nu_{i^\dagger,i}}{2}\bigg)
%     +
%     \bbP\bigg(\frac{\hmu_{i^\dagger}(t)}{c_{i^\dagger}}-\nu_{i^\dagger}\leq -\frac{\Delta^\nu_{i^\dagger,i}}{2}\bigg)
%     \\
%     &\leq 
%     2 \exp\big(-\const\cdot s),
% \end{align}
% where $\const$ is some instance-dependent constant. Therefore, 
% \begin{align}
%     \sum_{s=1}^\infty \bbP\bigg(\frac{\hmu_{i}(t)}{c_i}>\frac{\hmu_{i^\dagger}(t)}{c_{i^\dagger}}\bigg)
%     \leq
%     \sum_{s=1}^\infty 2 \exp\big(-\const\cdot s)
%     \leq 
%     2\int_{0}^\infty \exp\big(-\const\cdot s) \rmd s = \frac{2}{\const}<\infty.
% \end{align}
% Because $\bbE[|A_\varepsilon|]\leq \sum_{t=1}^\infty \bbP\bigg(\frac{\hmu_{i}(t)}{c_i}>\frac{\hmu_{i^\dagger}(t)}{c_{i^\dagger}}\bigg) $, we have 
% \begin{align}\label{equ:bound_A_epsilon}
%     \bbE[|A_\varepsilon|]\leq \frac{2}{\const}<\infty.
% \end{align}

% $\bullet$ Under event $A_\varepsilon$

For each $s\in\mathbb{N}_+$, define the stopping time
\begin{align}
    \phi_s
    :=
    \inf\Big\{
        t\geq 1:
        K\leq t< \tau,\,
        L_t=i,\,
        \tilT_i(t)=s,\,
        T_{i^\dagger}(t)\geq \varepsilon s
    \Big\},
\end{align}
with the convention that $\phi_s=\tau+1$ if the above set is empty.
Since the conditions in the definition of $\phi_s$ are adapted,
$\phi_s$ is an $\calF$-stopping time and satisfies
$\phi_s\leq\tau+1$ almost surely.

For every $s$, there is at most one time step $t$ such that
$L_t=i$ and $\tilT_i(t)=s$. Moreover,
$\phi_s\leq\tau$ if and only if this time step belongs to
$A_\varepsilon$. Therefore,
\begin{align}
    |A_\varepsilon|
    =
    \sum_{s=1}^{\infty}
    \mathbf{1}\{\phi_s\leq\tau\},
\end{align}
and, by Tonelli's theorem,
\begin{align}
    \bbE[|A_\varepsilon|]
    =
    \sum_{s=1}^{\infty}
    \bbP(\phi_s\leq\tau).
    \label{eq:A_epsilon_stopping_decomposition}
\end{align}

Let
$$
    \Delta^\nu_{i^\dagger,i}
    :=
    \nu_{i^\dagger}-\nu_i
    >0.$$
On the event $\{\phi_s\leq\tau\}$, candidate $i$ is the leader, so
\begin{align}
    \frac{\hmu_i(\phi_s)}{c_i}
    \geq
    \frac{\hmu_{i^\dagger}(\phi_s)}{c_{i^\dagger}}.
\end{align}
Furthermore, by the forced-sampling rule and the definition of
$\phi_s$,
\begin{align}
    T_i(\phi_s)\geq \frac{s}{3},
    \qquad
    T_{i^\dagger}(\phi_s)\geq \varepsilon s.
\end{align}
Consequently,
\begin{align}
    \bbP(\phi_s\leq\tau)
    &\leq
    \bbP\left(
        \frac{\hmu_i(\phi_s)}{c_i}-\nu_i
        \geq
        \frac{\Delta^\nu_{i^\dagger,i}}{2},
        \ \phi_s\leq\tau
    \right)
    % \nonumber\\
    % &\quad+
    +
    \bbP\left(
        \frac{\hmu_{i^\dagger}(\phi_s)}{c_{i^\dagger}}
        -\nu_{i^\dagger}
        \leq
        -\frac{\Delta^\nu_{i^\dagger,i}}{2},
        \ \phi_s\leq\tau
    \right).
    \label{eq:A_epsilon_deviation_decomposition}
\end{align}

The accepted-token length lies in an interval of length $L$.
Applying Lemma~\ref{lem:A_error_bound} to candidate $i$, with
sample-count lower bound $s/3$ and deviation level
$c_i\Delta^\nu_{i^\dagger,i}/2$, gives
\begin{align}
    &\bbP\left(
        \frac{\hmu_i(\phi_s)}{c_i}-\nu_i
        \geq
        \frac{\Delta^\nu_{i^\dagger,i}}{2},
        \ \phi_s\leq\tau
    \right)
    % \nonumber\\
    % &\qquad\leq
    \leq
    \exp\left(
        -\frac{
            s c_i^2
            \big(\Delta^\nu_{i^\dagger,i}\big)^2
        }{
            6L^2
        }
    \right).
\end{align}
Similarly, applying Lemma~\ref{lem:A_error_bound} to the negated
reward of candidate $i^\dagger$, with sample-count lower bound
$\varepsilon s$, gives
\begin{align}
    &\bbP\left(
        \frac{\hmu_{i^\dagger}(\phi_s)}{c_{i^\dagger}}
        -\nu_{i^\dagger}
        \leq
        -\frac{\Delta^\nu_{i^\dagger,i}}{2},
        \ \phi_s\leq\tau
    \right)
    % \nonumber\\
    % &\qquad\leq
    \leq
    \exp\left(
        -\frac{
            \varepsilon s c_{i^\dagger}^2
            \big(\Delta^\nu_{i^\dagger,i}\big)^2
        }{
            2L^2
        }
    \right).
\end{align}

Define the instance-dependent constant
\begin{align}
    \const
    :=
    \min\left\{
        \frac{
            c_i^2
            \big(\Delta^\nu_{i^\dagger,i}\big)^2
        }{
            6L^2
        },
        \frac{
            \varepsilon c_{i^\dagger}^2
            \big(\Delta^\nu_{i^\dagger,i}\big)^2
        }{
            2L^2
        }
    \right\}
    >0.
\end{align}
Combining the preceding inequalities yields
\begin{align}
    \bbP(\phi_s\leq\tau)
    \leq
    2\exp(-\const\cdot s).
\end{align}
Therefore,
\begin{align}
    \bbE[|A_\varepsilon|]
    &=
    \sum_{s=1}^{\infty}
    \bbP(\phi_s\leq\tau)
    \nonumber\\
    &\leq
    \sum_{s=1}^{\infty}
    2\exp(-\const\cdot s)
    \nonumber\\
    &\leq
    2\int_{0}^{\infty}
    \exp(-\const\cdot x)\rmd x
    =
    \frac{2}{\const}
    <\infty.
    \label{equ:bound_A_epsilon}
\end{align}

\begin{lemma}\label{lem:A_error_bound}
    Let $\calF=(\calF_t)_{t\in\bbN}$ be a filtration, and $X_t$ is $\calF_t$-adapted, s.t., $\bbE[X_t\mid\calF_{t-1}]=0$ and $X_t\in [a,b]$ with $b-a\leq c$. $B_t\in\{0,1\}$ is an $\calF_{t-1}$-measurable random variable.
    $\tau$ and $\phi$ are $\calF$-measurable stopping times such that 
    % either (1) $T_\phi\geq s,\phi<\tau$ or (2) $\phi=\tau$,
    $\phi\leq\tau+1$ almost surely and
    $T_\phi\geq s$ on the event $\{\phi\leq\tau\}$,
    where $\bbE[\tau]<\infty$.
    % and $\phi\in[\tau]$ is a stopping time.
    Define $\Gamma_t=\sum_{s=1}^t B_s\cdot X_s$ and $T_t =\sum_{s=1}^t B_s$. We have
    \begin{align}
    \bbP( \Gamma_\phi \geq T_\phi \delta, \phi\leq \tau)
    \leq 
    \exp\bigg( - \frac{2s\delta^2}{c^2}\bigg).
\end{align}
\end{lemma}

$\bullet$ Under event $B_\varepsilon$

The better candidate in the neighbor $i^\dagger$ is sampled fewer times. Let $\delta\in(0,\frac{\nu_{i^\dagger}-\nu_i}{2})$, we will show that this event can happen only when at least one of the following events occurs
\newline
(1) the leader candidate $i$ is overestimated,
\begin{align}
    C(\delta) := \{t:L_t = i, |\hnu_i(t)-\nu_i|>\delta\};
\end{align}
\newline
(2) some other bad candidate $j$ is overestimated,
\begin{align}
    D_j(\delta):=\{t:L_t=i,I_{t+1}=j,|\hnu_j(t)-\nu_j|>\delta\}
    \quad\mbox{and}\quad
    D(\delta) := \bigcup_{j\in\calN(i)\setminus\{i^\dagger\}} D_j(\delta);
\end{align}
\newline
(3) $i^\dagger$ is underestimated,
\begin{align}
    E := \{t<\ST: L_t=i, \ucb_{i^\dagger,i}(t)\leq \nu_{i^\dagger}\}.
\end{align}

Let $t\in B_\varepsilon$, when $i$ is the leader,
\begin{align}
    \tilT_i(t) = \tilT_{i^\dagger,i}(t) + \sum_{j\in\calN(i)\setminus\{i^\dagger\} } \tilT_{j,i}(t).
\end{align}
Because $t\in B_\varepsilon$, $\tilT_{i^\dagger,i}(t)\leq \varepsilon \tilT_i(t)$ which indicates 
\begin{align}
    (1-\varepsilon)\tilT_{i}(t) \leq \sum_{j\in\calN(i)\setminus\{i^\dagger\} } \tilT_{j,i}(t).
\end{align}
Let $\varepsilon = \frac{1}{6}$, we show that we either have 
\begin{itemize}
    \item[(a)]  there exists $j\in\calN(i)\setminus\{i,i^\dagger\}$, such that $\tilT_{j,i}(t)\geq \frac{\tilT_i(t)}{3}$.
    \item[(b)] $\tilT_{i,i}(t)\geq \frac{\tilT_i(t)}{2}+1$.
\end{itemize}
If  (a) $\tilT_{j,i}(t)\geq \frac{\tilT_i(t)}{3}$ holds:
there exists a unique time step $\phi(t)<t$ such that $L_{\phi(t)}=i$, $I_{\phi(t)+1}=j$ and $\tilT_{j,i}({\phi(t)}) = \lfloor\frac{\tilT_i(t)}{6}\rfloor$.
We will show that there exists a $\tilT_0$ such that for $\tilT_i(t)\geq \tilT_0$ and $\delta$ small enough, we have $\phi\in C(\delta)\cup D(\delta)\cup E$.
We prove by contradiction. If not, then $\ucb_{i^\dagger,i}(\phi)\geq \nu_{i^\dagger}$ and $\hnu_j(\phi)\leq \nu_j+\delta$. Therefore,
\begin{align}
    \ucb_{j,i}(\phi)&\leq \hnu_j(\phi)+ \frac{1}{c_j}\textrm{cr}_{j,i}(\phi)
    \\
    &
    \leq \nu_j+\delta + \frac{L}{2c_j}\sqrt{
            \frac{1+T_j(\phi(t))}{T_j^2(\phi(t))}\Big(
            1+2\ln\big(3\cdot\tilT_i^2(\phi(t))\cdot(1+T_j(\phi(t)))^{1/2}\big)
            \Big)
        }
    \\
    &
    \leq \nu_j+\delta + \frac{L}{2c_j}\sqrt{
            \frac{1+\tilT_{j,i}(\phi(t))}{\tilT_{j,i}^2(\phi(t))}\Big(
            1+2\ln\big(3\cdot\tilT_i^2(t)\cdot(1+\tilT_{j,i}(\phi(t)))^{1/2}\big)
            \Big)
        }
    \\
    &\leq
    \nu_j+\delta + \underbrace{\frac{L}{2c_j}\sqrt{
            \frac{36(1+\tilT_i(t))}{\tilT_i^2(t)}\Big(
            1+2\ln\big(3\cdot\tilT_i^2(t)\cdot(1+\tilT_i(t)/6)^{1/2}\big)
            \Big)
        }}_{\tilde{\textrm{cr}}(\tilT_i(t))/c_j}.
\end{align}
Note that $\tilde{\textrm{cr}}(x)/c_j=\Theta(\sqrt{\frac{\ln x}{x}})$ is decreasing in $x$ for $x\geq 1$ and converges to $0$ as $x$ grows.
Hence, if we choose $\delta<\frac{\nu_{i^\dagger}-\nu_i}{2}$ and choose $\tilT_0$ such that $\tilde{\textrm{cr}}(\tilT_0)/c_j<\delta$, we have
\begin{align}
    \ucb_{j,i}(\phi)\leq \nu_j+2\delta\leq \ucb_{i^\dagger,i}(\phi).
\end{align}
This leads to a contradiction.
Hence, $\phi\in C(\delta)\cup D(\delta)\cup E$.

For (b) $\tilT_{i,i}(t)\geq \frac{\tilT_i(t)}{2}+1$:
there are at least $\frac{\tilT_i(t)}{6}$ time steps $s$, such that $\tilT_i(s)-1$ is not a multiple of $3$ but $L_s=I_{s+1}=i$.
Denote the non-forced pulls of arm $i$ given arm $i$ is the leader arm as $\tilT^{\rm{nf}}_{i,i}(t):=\sum_{s=K}^t\mathbbm{1}\{L_s=I_{s+1}=i,\tilT_i(s)\not\equiv 1 \pmod{3}\}$.
There exists a unique time step $\phi(t)<t$, such that $L_{\phi(t)}=i,I_{\phi(t)+1}=i$ and 
% $\tilT_{i,i}(\phi(t))=\lfloor\frac{\tilT_i(t)}{6}\rfloor$ 
$\tilT^{\rm{nf}}_{i,i}(\phi(t))=\lfloor\frac{\tilT_i(t)}{6}\rfloor$ 
and $\tilT_i(\phi(t))-1$ cannot be divided by $3$. Then $
\ucb_{i,i}(\phi(t))\geq \ucb_{i^\dagger,i}(\phi(t))$. Applying the same argument as in case (a), we can conclude that $\phi(t)\in C(\delta)\cup D(\delta)\cup E$ when $\tilT_i(t)$ is large.

Define $B_{\varepsilon,\tilT_0}:=\{t:t\in B_\varepsilon,\tilT_i(t)>\tilT_0\}$. By the uniqueness of the $\phi$ in the above discussions in cases (a) and (b), it defines a mapping from $ B_{\varepsilon,\tilT_0}$ to $C(\delta)\cup D(\delta)\cup E$.
Therefore,
\begin{align}
    \{t:t\in B_{\varepsilon,\tilT_0},\tilT_i(t)\geq \tilT_0\}
    \subset
    \bigcup_{s\in C(\delta)\cup D(\delta)\cup E} \{t:t\in B_{\varepsilon,\tilT_0},\phi(t)=s\}.
\end{align}
Furthermore, 
for any time step $s$, if $t\in B_{\varepsilon,\tilT_0}$ and $\phi(t)=s$, then we have 
\begin{align}
    \bigg\lfloor \frac{\tilT_i(t)}{6}\bigg\rfloor \in \bigg(\bigcup_{j\in\calN(i)\setminus\{i^\dagger,i\}} \Big\{\tilT_{j,i}(s)\Big\}\bigg)\bigcup \Big\{\tilT^{\rm{nf}}_{i,i}(s) \Big\}.
\end{align}
This implies
\begin{align}
    |\{t:t\in B_{\varepsilon,\tilT_0},\phi(t)=s\}|\leq 2\times 6=12
\end{align}
which further yields
\begin{align}\label{equ:bound_B_epsilon}
    |B_\varepsilon|\leq \tilT_0 + 12\cdot \Big(|C(\delta)|+ |D(\delta)|+|E|\Big).
\end{align}

Lastly, we upper bound the cardinality of the three sets.

By applying Lemma~\ref{lem:A_error_bound}, we have
\begin{align}\label{equ:bound_CD}
    &\bbE[|C(\delta)|] \leq 2\sum_{s=1}^\infty \exp\bigg( -\frac{2\lfloor s/3 \rfloor (c_i\delta)^2}{L^2}\bigg)
    <\infty
    \\
    &
    \bbE[|D(\delta)|] \leq 2\sum_{j\in\calN(i)\setminus\{i^\dagger\}} \sum_{s=1}^\infty \exp\bigg( -\frac{2s (c_j\delta)^2}{L^2}\bigg)
    <\infty.
\end{align}
To bound $|E|$, we adopt Lemma~\ref{lem:ConfInt} for each leader $i$ with its neighborhood. Specifically, when candidate $i$ is the leader, we let $s=\tilT_{i}(t)$. Then with probability at least $1-1/s^2$, we have
\begin{align}\label{equ:bound_E_cfintv1}
    |\hmu_{j}(t)-\mu_j|\leq \frac{L}{2}\sqrt{
            \frac{1+T_j(t)}{T_j^2(t)}\Big(
            1+2\ln\big(3\cdot s^2\cdot(1+T_j(t))^{1/2}\big)
            \Big)
        }
        \quad
        \forall t>0, j\in\calN(i).
\end{align}
This indicates $\bbP(L_t=i,\ucb_{i^\dagger,i}(t)\leq \nu_{i^\dagger})\leq 1/s^2$. Thus
\begin{align}\label{equ:bound_E}
    \bbE[|E|]\leq \sum_{s=1}^\infty \frac{1}{s^2} = \frac{\pi^2}{6}<\infty.
\end{align}

In conclusion, \eqref{equ:bound_B_epsilon}, \eqref{equ:bound_CD} and \eqref{equ:bound_E} yield
\begin{align}\label{equ:bound_B_epsilon2}
        \bbE[|B_\varepsilon|]\leq \tilT_0 + \const^\prime <\infty,
\end{align}
where $\const^\prime$ is some constant that only depends on the instance.
This bound does not scale with $\bbE[\ST]$.

According to \eqref{equ:bound_A_epsilon} and \eqref{equ:bound_B_epsilon2}, we have that $(\clubsuit)$ does not scale with $\bbE[\ST]$
\begin{align}\label{equ:bound_clubsuit}
    (\clubsuit) < \infty.
\end{align}

\noindent
\underline{Bound $(\spadesuit)$}
Define
\begin{align}
    \Good(s) = \Big\{|\hmu_{j}(t)-\mu_j|\leq b(T_j(t),s)
        \quad
        \forall t>0, j\in\calN(i^*)\Big\}.
\end{align}
According to Lemma~\ref{lem:ConfInt},
\begin{align}\label{equ:bound_E_cfintv}
    \bbP(\Good(s))\geq 1-\frac{1}{s^2}.
\end{align}
By our sampling policy,
\begin{align}
    &\sum_{t=K}^{\ST-1} \mathbbm{1}\{I_{t+1}=i,L_t=i^*\}
    \\
    &\quad\leq
    \sum_{t=K}^{\ST-1} \mathbbm{1}\{I_{t+1}=i,L_t=i^*,\ucb_{i,i^*}(t)>\ucb_{i^*,i^*}(t)\}
    \\
    &\quad\leq
    \sum_{t=K}^{\ST-1} \mathbbm{1}\{I_{t+1}=i,L_t=i^*,\ucb_{i,i^*}(t)>\nu_{i^*},\Good(\tilT_{i^*}(t))\}
    \\
    &\qquad+
    \sum_{t=K}^{\ST-1}\mathbbm{1}\{L_t=i^*,\Good(\tilT_{i^*}(t))^c\}
\end{align}

% \begin{lemma}
%     \begin{align}
%         \sum_{t=1}^{\ST} \mathbbm{1}\{I_t=i,L_t=i^*,\ucb_{i,i^*}(t)>\nu_{i^*}\}
%         \leq
%     \end{align}
% \end{lemma}
For $x,y\in\bbZ_+$, define 
\begin{align}
    b(x,y):=\frac{L}{2}\sqrt{
            \frac{1+x}{x^2}\Big(
            1+2\ln\big(3\cdot y^2\cdot(1+x)^{1/2}\big)
            \Big)
        },
\end{align}
so that $\textrm{cr}_{i,i^*}(t)=b(T_i(t),\tilT_{i^*}(t))$.
By the definition of $\ucb_{i,i^*}$, $T_i(t)\geq \tilT_{i,i^*}(t)$ and $\tilT_{i,i^*}(t)\leq \tilT_{i^*}(t)\leq \tau$, we have 
\begin{align}\label{equ:bound_spade}
    &\sum_{t=K}^{\ST-1} \mathbbm{1}\{I_{t+1}=i,L_t=i^*,\ucb_{i,i^*}(t)>\nu_{i^*},\Good(\tilT_{i^*}(t))\}
    \\
    &\quad\leq
    \sum_{t=K}^{\ST-1} \mathbbm{1}\bigg\{I_{t+1}=i,L_t=i^*,\hmu_i+b(T_i(t),\tilT_{i^*}(t)) >c_i\nu_{i^*},\Good(\tilT_{i^*}(t))\bigg\}
    \\
    &\quad\leq
    \sum_{t=K}^{\ST-1} \mathbbm{1}\bigg\{I_{t+1}=i,L_t=i^*,T_i(t)\leq t_i(\ST),\hmu_i+b(T_i(t),\tilT_{i^*}(t)) >c_i\nu_{i^*},\Good(\tilT_{i^*}(t))\bigg\}
    \\
    &\qquad+
    \sum_{t=K}^{\ST-1} \mathbbm{1}\bigg\{I_{t+1}=i,L_t=i^*,T_i(t)> t_i(\ST),\hmu_i+b(T_i(t),\tilT_{i^*}(t)) >c_i\nu_{i^*},\Good(\tilT_{i^*}(t))\bigg\}
    \\
    &\quad \leq
    t_i(\ST) + \sum_{t=K}^{\ST-1} \mathbbm{1}\bigg\{I_{t+1}=i,L_t=i^*,T_i(t)> t_i(\ST),\hmu_i+b(T_i(t),\tilT_{i^*}(t)) >c_i\nu_{i^*},\Good(\tilT_{i^*}(t))\bigg\},
    % \sum_{t=1}^{\ST} \mathbbm{1}\bigg\{I_t=i,L_t=i^*,\hmu_i+b(\tilT_{i,i^*}(t),\tilT_{i^*}(t)) >c_i\nu_{i^*}\bigg\},
\end{align}
where 
\begin{align}
    t_i(\ST) &= \frac{8L^2\ln \ST}{(c_i\nu_{i^*}-\mu_i)^2}+\frac{2L^2}{(c_i\nu_{i^*}-\mu_i)^2}\bigg(
        2\ln\frac{L}{c_i\nu_{i^*}-\mu_i}+ \frac{(c_i\nu_{i^*}-\mu_i)^2}{L^2} +1+2\ln 3
    \bigg)
    \\
    &
    =O\bigg( \frac{L^2\ln \ST}{(c_i\nu_{i^*}-\mu_i)^2}\bigg).
\end{align}
% where the last inequality comes from the monotonicity of $b(x,y)$ in $x$. Furthermore
% \begin{align}
%     &\sum_{t=1}^{\ST} \mathbbm{1}\bigg\{I_t=i,L_t=i^*,\hmu_i+b(\tilT_{i,i^*}(t),\tilT_{i^*}(t)) >c_i\nu_{i^*}\bigg\}
%     \\
%     &\quad=
%     \sum_{t=1}^{\ST} \mathbbm{1}\bigg\{I_t=i,L_t=i^*,\tilT_{i,i^*}(t)\leq t_i(\tau),\hmu_i+b(\tilT_{i,i^*}(t),\tilT_{i^*}(t)) >c_i\nu_{i^*}\bigg\}
%     \\
%     &\qquad+
%     \sum_{t=1}^{\ST} \mathbbm{1}\bigg\{I_t=i,L_t=i^*,\tilT_{i,i^*}(t)>t_i(\tau),\hmu_i+b(\tilT_{i,i^*}(t),\tilT_{i^*}(t)) >c_i\nu_{i^*}\bigg\}
%     \\
%     &\quad\leq
%     t_i(\tau) + 
% \end{align}

% Define
% \begin{align}
%     \Good_i(s) = \Big\{|\hmu_{i}(t)-\mu_i|\leq b(T_i(t),s)
%         \quad
%         \forall t>0, j\in\calN(i^*)\Big\}.
% \end{align}
% According to Lemma~\ref{lem:ConfInt},
% \begin{align}\label{equ:bound_E_cfintv}
%     \bbP(\Good_i(s))\geq 1-\frac{1}{s^2}.
% \end{align}
When $T_i(t)\geq t_i(\tau)$, conditional on $\Good(\tilT_{i^*}(t))$,
\begin{align}
    \hmu_i+b(T_i(t),\tilT_{i^*}(t)) 
    \leq
    \mu_i+2b(t_i(\tau),\tau)
    \leq
    c_i\nu_{i^*},
\end{align}
where we adopt the monotonicity of $b(x,y)$ in the first inequality and Lemma~\ref{lem:antos} in the second inequality.

By \eqref{equ:bound_spade} and \eqref{equ:bound_E_cfintv},
\begin{align}
    &\bbE\bigg[\sum_{t=K}^{\ST-1} \mathbbm{1}\{I_{t+1}=i,L_t=i^*,\ucb_{i,i^*}(t)>\nu_{i^*},\Good(\tilT_{i^*}(t))\}\bigg]
    \\
    &\leq
    \bbE[t_i(\ST)] + \bbE\bigg[\sum_{s=1}^\tau \mathbbm{1}\{\Good^c(s)\}\bigg]
    \\
    &\leq
    \frac{8L^2\ln\bbE[\ST]}{(c_i\nu_{i^*}-\mu_i)^2}+\frac{2L^2}{(c_i\nu_{i^*}-\mu_i)^2}\bigg(
        2\ln\frac{L}{c_i\nu_{i^*}-\mu_i}+ \frac{(c_i\nu_{i^*}-\mu_i)^2}{L^2} +1+2\ln 3
    \bigg)
    +
    \frac{\pi^2}{6}
    \\
    &=O\bigg(\frac{L^2\ln\bbE[\ST]}{(c_i\nu_{i^*}-\mu_i)^2}\bigg).
\end{align}
This gives the bound for $(\spadesuit)$:
\begin{align}\label{equ:bound_spadesuit}
    \bbE\bigg[\sum_{t=K}^{\ST-1} \mathbbm{1}\{I_{t+1}=i,L_t=i^*\}\bigg]
    \leq
    \frac{8L^2\ln\bbE[\ST]}{(c_i\nu_{i^*}-\mu_i)^2} + O(1).
\end{align}

\textbf{Step 3}
We summarize the results in the previous steps: \eqref{equ:regret_decomposition}, \eqref{equ:bound_clubsuit} and \eqref{equ:bound_spadesuit} yield
\begin{align}
    \WTReg(\Lambda,\prompt)
    &\leq
    \sum_{i\in\calN(i^*)\setminus\{i^*\}}\mu_i(\frac{1}{\nu_{i}}-\frac{1}{\nu_{i^*}})\cdot \frac{8L^2\ln\bbE[\ST]}{(c_i\nu_{i^*}-\mu_i)^2} +O(1)
    \\
    &=
    \rmH(\Lambda,\prompt) \cdot \ln \bbE[\ST]+O(1),
\end{align}
where 
\begin{align}
    \rmH(\Lambda,\prompt)=\sum_{i\in\calN(i^*)\setminus\{i^*\}}\frac{1}{\nu_{i^*}c_i}\cdot \frac{8L^2}{\nu_{i^*}-\nu_i}  .
\end{align}

\section{Technical Lemmas}\label{app:tech_lemma}
\subsection{Proof of Lemma~\ref{lem:optimal_policy}}\label{app:proof_optimal_policy}
Denote $\nu_S= \frac{\mu_S}{c_S}$ for $S\in\calS$.
By Doob's optional stopping Theorem (Lemma~\ref{lem:DoobOptStopping}), we have 
\begin{align}
    \bbE[\WT_{\ST(\alg,\prompt,\Lambda)}\mid\prompt]
    &=
    \bbE\bigg[\sum_{t=1}^{\ST(\alg,\prompt,\Lambda)}c_{I_t}\mid\prompt\bigg]
    \\
    &=
    \sum_{S\in\calS}c_S\cdot \bbE\big[T_S(\ST(\alg,\prompt,\Lambda))\mid\prompt\big]
    \\
    &=
    \sum_{S\in\calS}\frac{\mu_S}{\nu_S}\cdot \bbE\big[T_S(\ST(\alg,\prompt,\Lambda))\mid\prompt\big].
\end{align}
Similarly, the expected length (which only depends on the target model) can be decomposed into
\begin{align}
    \bbE\big[\len(\prompt_{\ST(\alg,\prompt,\Lambda)})-\len(\prompt)\mid\prompt\big]
    =
    \bbE\bigg[\sum_{t=1}^{\ST(\alg,\prompt,\Lambda)}\len(x_{S_t,t})\mid\prompt\bigg]
    =
    \sum_{S\in\calS} \mu_S \cdot \bbE\big[T_S(\ST(\alg,\prompt,\Lambda))\mid\prompt\big].
\end{align}
Therefore, 
\begin{align}
    \bbE[\WT_{\ST(\alg,\prompt,\Lambda)}\mid\prompt]
    \geq \min_{S\in\calS}\frac{1}{\nu_S}\bbE\big[\len(\prompt_{\ST(\alg,\prompt,\Lambda)})-\len(\prompt)\mid\prompt\big],
\end{align}
where the minimum expected wall time is obtained at $S_t=S^*=\argmax_{S}\nu_S $ for all $t$. This indicates $\alg^*=\alg_{S^*}.$

\subsection{Proof of Lemma~\ref{lem:unimodal}}
Let $\Delta_{\mu_i} = \mu_{i+1}-\mu_i$, which is non-increasing in $i$ according to the diminishing assumption, and let $s_i:=|S_i|$ and $s_{i+1}-s_i=h$.
We have
\begin{align}
    \Delta_{\nu_i} := \frac{\mu_{i+1}}{c_{i+1}} - \frac{\mu_{i}}{c_{i}}
    =\frac{
        \Delta_{\mu_i} \cdot (\kappa s_i+b) - \kappa h\mu_i
    }{
    (\kappa s_{i+1}+b)(\kappa s_i+b)
    }.
\end{align}
Let $g_i:=\Delta_{\mu_i} \cdot (\kappa s_i+b) - \kappa h\mu_i$, and 
\begin{align}
    \Delta_{g_i}&:=g_{i+1}-g_i
    \\
    &=\Delta_{\mu_{i+1}} \cdot (\kappa s_{i+1}+b) - \kappa h\mu_{i+1}
        -
        \Delta_{\mu_i} \cdot (\kappa s_i+b) + \kappa h\mu_i
    \\
    &=(\kappa s_{i+1}+b)(\Delta_{\mu_{i+1}}-\Delta_{\mu_i})
    \leq 0
\end{align}
Therefore, $g_i$ is non-increasing in $i$ and $\Delta_{\nu_i}$ can change sign from positive to negative at most once. This concludes that $\nu$ is a unimodal function.

\subsection{Proof of Lemma~\ref{lem:A_error_bound}}
For any $\lambda>0$, we define
\begin{align}
    &Y_s = B_s(\lambda X_s-\frac{\lambda^2c^2}{8}) ,\quad
    \tilG_t = \exp\bigg(\sum_{s=1}^t Y_s\bigg) \mathbbm{1}\{t\leq \tau \}\\
    &\quad\mbox{and}\quad
    G_t = \exp\big( \lambda (\Gamma_t-\delta T_t)\big)\mathbbm{1}\{t\leq \tau \}
\end{align}

$\bullet$ $\{\tilG_t\}$ is a supermartingale.

As $X_s\in[a,b]$ with $b-a<c$, and $\bbE[X_s\mid \calF_{s-1}]=0$, we have $X_s$ is a $\frac{c^2}{4}$-subGaussian random variable conditional on $\calF_{s-1}$. Therefore, $\bbE[\exp(\lambda X_s)\mid\calF_{s-1}]\leq \exp\big(\frac{\lambda^2c^2}{8}\big)$. Furthermore,
\begin{align}
    \bbE[\exp( Y_s)\mid\calF_{s-1}]
    &=
    \bbE\bigg[\exp\Big(B_s\Big(\lambda X_s-\frac{\lambda^2c^2}{8}\Big)\Big)\mid\calF_{s-1}\bigg]
    \\
    &=
    (1-B_s) + B_s \bbE\bigg[\exp\Big(\lambda X_s-\frac{\lambda^2c^2}{8}\Big)\mid\calF_{s-1}\bigg]
    \leq 
    1.
\end{align}
This indicates $\bbE[\tilG_t|\calF_{t-1}]\leq \tilG_{t-1}$, thus $\{\tilG_t\}$ is a supermartingale.

$\bullet$ Upper bound $G_t$.

By the definitions,
\begin{align}
    G_t &= \exp\big( \lambda (\Gamma_t-\delta T_t)\big)\mathbbm{1}\{t\leq \tau \}
    \\
    &=
    \exp\bigg( \sum_{s=1}^t Y_s - \lambda\delta T_t + \frac{\lambda^2c^2}{8} T_t\bigg)\mathbbm{1}\{t\leq \tau \}
    \\
    &=
    \tilG_t\cdot \exp\bigg( - T_t \Big(\lambda \delta- \frac{\lambda^2c^2}{8} \Big)\bigg)
    \\
    &\overset{\lambda=\frac{4\delta}{c^2}}{=}
    \tilG_t\cdot \exp\bigg( - T_t \frac{2\delta^2}{c^2}\bigg).
\end{align}
When $\phi\leq \tau$ and $T_\phi\geq s$,
\begin{align}
    G_\phi = \tilG_\phi \exp\bigg( - T_\phi \frac{2\delta^2}{c^2}\bigg) 
    \leq
    \tilG_\phi \exp\bigg( - \frac{2s\delta^2}{c^2}\bigg).
\end{align}
When $\phi=\tau+1$,$G_\phi =\tilG_\phi =0$, the above inequality still holds.
Note that
\begin{align}
    \bbP( \Gamma_\phi \geq T_\phi \delta, \phi\leq \tau)
    \leq
    \bbP\Big( \exp\big( \lambda (\Gamma_\phi-\delta T_\phi)\big) \mathbbm{1}\{\phi\leq \tau\}\geq 1\Big)
    \leq
    \bbE[G_\phi],
\end{align}
we have
\begin{align}\label{equ:boundGphi}
    \bbP( \Gamma_\phi \geq T_\phi \delta, \phi\leq \tau)
    \leq 
    \bbE[\tilG_\phi]\exp\bigg( - \frac{2s\delta^2}{c^2}\bigg).
\end{align}

$\bullet$ Bound $\bbE[\tilG_\phi]$.

As $\phi$ can be infinity, we cannot directly apply the supermartingale property to upper bound $\bbE[\tilG_\phi]$.
Define $\phi_m = \phi\wedge m $ for a given $m\in\bbN$. $\phi_m$ is upper bounded by $m$.
Therefore, we can apply the optional stopping theorem for the nonnegative supermartingale and obtain $\bbE[\tilG_{\phi_m}]\leq \bbE[\tilG_{0}]=1$.
Because $\bbE[\tau]<\infty$, then $\phi_m\to\phi$ almost surely.
This indicates $\tilG_{\phi_m}\to \tilG_\phi $ almost surely.
By Fatou's Lemma,
\begin{align}
    \bbE[\tilG_\phi]\leq \liminf_{m\to\infty} \bbE[\tilG_{\phi_m}]\leq 1.
\end{align}
Combining with \eqref{equ:boundGphi}, we have 
\begin{align}
    \bbP( \Gamma_\phi \geq T_\phi \delta, \phi\leq \tau)
    \leq 
    \exp\bigg( - \frac{2s\delta^2}{c^2}\bigg).
\end{align}

\section{Supporting Lemmas}
\begin{lemma}[Doob's optional stopping, Theorem 3.8 in~\citet{lattimore2020bandit}]\label{lem:DoobOptStopping}
Let $\bbF=(\calF_t)_{t\in\bbN}$ be a filtration and $(X_t)_{t\in\bbN}$ be an $\bbF$-adapted martingale and $\tau$ an $\bbF$-stopping time such that at least one of the following holds:
\begin{itemize}
    \item[(a)] There exists an $n\in\bbN$ such that $\bbP[\tau>n]=0$.
    \item[(b)] $\bbE[\tau]<\infty$, and there exists a constant $c\in\bbR$ such that for all $t\in\bbN$, $\bbE[|X_{t+1}-X_t|\mid\calF_t]\leq c$ almost surely on the event that $\tau>t$.
    \item[(c)] There exists a constant $c$ such that $|X_{t\wedge \tau}|\leq c$ almost surely for all $t\in\bbN$.
\end{itemize}
Then $X_{\tau}$ is almost surely well-defined, and $\bbE[X_{\tau}]=\bbE[X_0]$. 
Furthermore, when $(X_t)$ is a super/sub-martingale rather than a martingale, then equality is replaced with less/greater-than, respectively.
    
\end{lemma}

% \begin{lemma}[Chernoff-Hoeffding bound, Fact 1 in~\citet{auer2002finite}]\label{lem:ChernoffHoeffding}
%     Let $X_1,\ldots,X_n$ be random variables with common range $[0,1]$ and $\bbE[X_n\mid X_1,\ldots,X_{n-1}] = \mu$. Let $S_n = X_1+\ldots + X_n.$ Then for all $a\geq 0$,
%     \begin{align}
%         \bbP[S_n\geq n\mu+a]\leq \exp\left(
%             -\frac{2a^2}{n}
%         \right)
%         \quad
%         \mbox{ and }
%         \quad
%         \bbP[S_n\geq n\mu-a]\leq \exp\left(
%             -\frac{2a^2}{n}
%         \right)
%     \end{align}
% \end{lemma}

\begin{lemma}[Confidence Intervals, Lemma 6 in~\citet{Abbasi2011improved}]\label{lem:ConfInt}
    Assume that the noise $\eta_t$ is conditionally $1$-sub-Gaussian. With probability at least $1 - \delta$,
    \begin{align}
    \forall i \in \{1, 2, \dots, K\}, \; \forall t \geq 0, \quad |\hmu_{i,t} - \mu_i| \leq 
    \sqrt{\frac{(1 + n_{i,t})}{n_{i,t}^2} \left(1 + 2 \log \left(\frac{K (1 + n_{i,t})^{1/2}}{\delta}\right)\right)}.
    % c_{i,t},
\end{align}
% \begin{align}
%     \forall i \in \{1, 2, \dots, d\}, \; \forall t \geq 0, \quad |\overline{X}_{i,t} - \mu_i| \leq 
%     \sqrt{\frac{(1 + N_{i,t})}{N_{i,t}^2} \left(1 + 2 \log \left(\frac{d (1 + N_{i,t})^{1/2}}{\delta}\right)\right)}.
%     % c_{i,t},
% \end{align}
% where
% \begin{align}
%     c_{i,t} = \sqrt{\frac{(1 + N_{i,t})}{N_{i,t}^2} \left(1 + 2 \log \left(\frac{d (1 + N_{i,t})^{1/2}}{\delta}\right)\right)}.
% \end{align}
\end{lemma}

\begin{lemma}[Lemma 8 in~\citet{antos2010active}]\label{lem:antos}
    Let $a>0$. For any $t\geq (2/a)[\log(1/a)-b]^+,at+b>\log t$.
\end{lemma}

\section{Experimental Details}
\label{app:experiment_details}

\textbf{Inputs and prompts.}
In the main experiments, we use greedy decoding, batch size $1$, and draft length $L=4$ for all speculative decoding methods.
We use $128$ frames per prompt on Video-MME, MMBench-Video, and ARKitScenes, and limit the input visual token size to be less than $16{,}384$;
whereas AIME25 evaluates text-only mathematical reasoning.
To obtain stable latency measurements, we add a suffix instruction to all the prompts to produce longer responses.
The resulting outputs are typically around $1000$--$2000$ tokens on the visual benchmarks, and we limit the output to be below $10{,}000$ tokens on AIME25.

\textbf{UCBSpec adaptation.}
The original \ucbspec{} uses the number of accepted tokens as its reward.
We extend the reward to \emph{tokens per second} to make a fair comparison with our \ourmethod{}.
Within each construction scheme, \ucbspec{} and \ourmethod{} use the same candidate draft models, whose construction is detailed in Appendix~\ref{app:draft_construction}.

\textbf{Optimizer state and resets.}
\ourmethod{} clears its arm-pull and reward history at the start of each new prompt and adapts online without auxiliary training.
By default, it also clears this history every $512$ speculative decoding steps within a prompt.
Appendix~\ref{app:ablation_reset} compares this default with a variant that resets only between prompts.

\section{Draft Model Construction}\label{app:draft_construction}

\subsection{Cosine-Similarity-Based Construction}\label{app:cossim_construction}

Each transformer block contains one attention layer and one MLP layer.
The target model $P$ has $N$ blocks, with block $b\in[N]=\{1,\ldots,N\}$ containing layers $2b-1$ and $2b$.
For a retained-layer set $S\subseteq[2N]$, $Q_S$ denotes the draft model obtained by skipping all layers outside $S$, keeping the retained layers in their original order and using the same parameters as $P$.
We always retain the first and last two blocks; the other retained blocks are eligible for removal.
Hence, at most $K_{\max}=\lfloor N-4\rfloor$ arms can be constructed, skipping one additional block at each step.

A block's redundancy score $r_b$ is the mean cosine similarity between its input and output hidden states over the token positions used for the current update. Let $\mathcal{U}$ denote the tokens generated in a speculative step, used for the current score update.
For each eligible retained block $b$, its redundancy score is
\begin{align*}
    r_b=\frac{1}{|\mathcal{U}|}\sum_{u\in\mathcal{U}}
    \frac{\langle h_{b,u}^{\mathrm{in}},h_{b,u}^{\mathrm{out}}\rangle}
    {\|h_{b,u}^{\mathrm{in}}\|_2\|h_{b,u}^{\mathrm{out}}\|_2},
\end{align*}
where $h_{b,u}^{\mathrm{in}},h_{b,u}^{\mathrm{out}}$ are the input and output hidden states of block $b$ for token $u$.
A higher score indicates greater redundancy; equivalently, the implementation removes the block with minimum importance $1-r_b$.
We initialize these scores during a four-token target-model warm-up.
After each new draft model's speculative step, we refresh the scores using its hidden states for accepted draft tokens and, if present, the first rejected draft token.

\begin{algorithm}[ht]
    \caption{\textsc{cosSim} Draft Model Construction}\label{algo:cossim_construction}
    \textbf{Inputs:} target model $P$, prefix $\prompt$, draft length $L$, output limit, arm limit $K_{\max}$.\par
    \begin{algorithmic}[1]
        \STATE Initialize retained layers $S\leftarrow[2N]$, eligible blocks $\mathcal B\leftarrow\{3,\ldots,N-2\}$, and an empty draft model list $\calQ\leftarrow()$.
        \STATE Generate four warm-up tokens $x$ with $P$, append $\prompt\leftarrow\concat(\prompt,x)$, and initialize redundancy scores $\{r_b\}_{b\in\mathcal B}$.
        \FOR{$j=1,\ldots,K_{\max}$, while generation is unfinished}
            \STATE Choose the eligible block with the highest redundancy score: $b\leftarrow\argmax_{b\in\mathcal B}r_b$.
            \STATE Skip both layers of this block: $S\leftarrow S\setminus\{2b-1,2b\}$, $\mathcal B\leftarrow\mathcal B\setminus\{b\}$.
            \STATE Run one speculative decoding step: $(x,c_S)\leftarrow\operatorname{SD}(P,Q_S,\prompt,L)$, where $x$ is the output token sequence and $c_S$ is the drafting and verification time.
            \STATE Add draft model $Q_S$ as an arm, $\calQ\leftarrow\concat(\calQ,(Q_S))$, and record its accepted-token count $\len(x)$ and time cost $c_S$.
            \STATE Append the output, $\prompt\leftarrow\concat(\prompt,x)$, and refresh the scores $\{r_b\}_{b\in\mathcal B}$ as described above.
        \ENDFOR
        \STATE Order the $K=|\calQ|$ draft models by increasing retained-layer count: $\calQ=(Q_{S_i})_{i=1}^K$, with $|S_1|<\cdots<|S_K|$.
        \STATE Initialize each arm's statistics from its recorded observation, setting its pull count $T_i\leftarrow1$ for $i\in[K]$.
        \STATE \return{} the draft models $\calQ$, their initialized statistics, and the updated prefix $\prompt$.
    \end{algorithmic}
\end{algorithm}

Generation ends when an end-of-sequence token is produced or the output limit is reached.
Each construction step generates output and provides its arm's first observation, so no separate initialization pulls are needed.
The arm construction overhead is mainly the calculation of cosine similarity, which is negligible. Hence \textsc{cosSim} is efficient in both draft model configuration-building and bandit exploration.

Appendix~\ref{app:single_peak_cossim} reports the empirical acceptance, time-cost, and throughput trends for this method.

\subsection{Construction from Existing Methods}\label{app:existing_draft_construction}

In each method specified below, we vary the computation budget to obtain $K$ distinct draft model configurations $S_1,\ldots,S_K$.
Here, $S_k$ specifies the retained attention and MLP layers for \textsc{SWIFT} and \textsc{KnapSpec}, or the retained visual tokens for \textsc{SpecVLM}.
Each configuration defines a draft model $Q_{S_k}$, which serves as arm $k$; all draft models reuse the target model's parameters.
We will order the arms by increasing computation budget, as specified below.
At each speculative round, the controller selects one draft model and records its accepted-token count and wall-time cost.
Section~\ref{sec:experiments} compares the original configuration-selection strategy, \textsc{UCBSpec}, and \spsqec{}; the two bandit controllers use the same candidate draft models within each construction scheme.

\textbf{SWIFT.}
\textsc{SWIFT}~\citep{xia2025swift} combines random search and Bayesian optimization to find a layer-skip configuration at a prescribed skip ratio $r$.
For a target model with $N$ transformer blocks, the first and last blocks are retained, leaving $2(N-2)$ \emph{eligible} attention and MLP layers that can be skipped.
Each Bayesian-optimization proposal assigns a continuous score in $[0,1]$ to each eligible layer; higher scores give higher priority for skipping.
In \textsc{SWIFT}, the proposed configuration is evaluated by its greedy next-token agreement with the target model on the accumulated context.

To construct arms, we fix the score vector of the best-performing Bayesian proposal and rank all eligible layers by decreasing score.
For each candidate skip ratio $r_k$, we skip the first $\lfloor 2(N-2)r_k\rfloor$ layers in this ranking and retain all other layers to obtain $S_k$.
Attention and MLP layers share one ranking and can be skipped independently.
The resulting draft model $Q_{S_k}$ is arm $k$.
We use the same ranking for every ratio and order the arms by $r_1>\cdots>r_K$, so successive draft models retain more layers and form nested retained-layer sets.

\textbf{KnapSpec.}
\textsc{KnapSpec}~\citep{cha2026knapspec} profiles attention and MLP layers separately and uses their estimated latencies as costs in a dynamic program.
The search considers retaining or skipping each attention and MLP layer independently, giving different \emph{candidates}, corresponding to different layer skip configurations.
For each intermediate depth and discretized budget, it compares candidate hidden states with the target model's hidden states at the same depth and token positions, retaining the candidate with the highest mean cosine similarity.

Eventually, at the final depth, the dynamic program provides layer-skip configurations for each feasible budget.
We express budgets in terms of retained computation: increasing $b$ allows more layer execution time, corresponding to a smaller saved-latency budget in \textsc{KnapSpec}'s discretized cost model.
To construct arms, we collect the configurations returned for the different budgets, and retain the highest-ranked configurations according to \textsc{KnapSpec}'s throughput estimates.
Each retained-layer set $S_k$ defines a draft model $Q_{S_k}$, and we index these arms by increasing retained-latency budget, $b_1<\cdots<b_K$.
We remark that the selected layers can change across budgets, so successive configurations need not be nested.

\textbf{SpecVLM.}
\textsc{SpecVLM}~\citep{ji2025specvlm} uses the target model's attention from language tokens to visual tokens during prefill to identify important visual tokens.
Its first stage retains highly attended visual tokens; its second stage samples from the remaining visual tokens uniformly in space to preserve spatial coverage.

To construct arms, we apply this two-stage selection for each candidate pruning ratio $p_k$, with a target retained fraction $1-p_k$ of the visual tokens.
The selected visual-token set $S_k$, together with all language tokens, determines a pruned draft input context.
In our self-speculative setting, $Q_{S_k}$ uses the target model's parameters with this reduced visual context, and serves as arm $k$.
Thus, these arms differ in their visual-token selections rather than their model parameters or retained layers.
Target-model verification always uses the full visual and textual context.
We order the arms by $p_1>\cdots>p_K$, equivalently by increasing retained visual-token count.

We remark that we modified the KV-cache handling to support online switching between these draft models.
The original \textsc{SpecVLM}'s fixed-ratio implementation prefills the shortened input to construct the draft KV cache.
We instead retain a complete draft prompt KV cache initialized from the target model's full-context prefill.
For arm $k$, draft attention uses the entries for $S_k$, all language-prompt tokens, and the cached generated history.
When the arm changes, we gather the selected entries into a reusable compact cache while preserving the original token positions and reusing the generated history.
This allows previously excluded visual tokens to be selected again without repeating prompt prefill or maintaining a separate cache for each arm.
The cached prompt states are therefore computed from the full context, rather than from a separately pruned input.

In all three constructions, the original method determines which layers or visual tokens to retain at each budget, while \spsqec{} adapts the choice among the resulting draft models during generation.
Intuitively, these methods prioritize retaining the layers or visual tokens that help the drafter match the target model.
As the budget grows, more of the essential computation is already retained, so additional components are expected to yield smaller acceptance gains.
For a fixed draft length, this also means diminishing gains in the mean accepted-token count, while the time per speculative round grows approximately linearly with retained computation.
Improved acceptance can initially compensate for the extra drafting time; once its gains become too small, the added cost reduces throughput.
This tradeoff hence suggests an approximately single-peak throughput profile.
Appendix~\ref{app:single_peak} provides average and prompt-level empirical evidence for their acceptance, time-cost, and single-peak throughput trends.

\section{Empirical Validation of the Single-Peak Structure}
\label{app:single_peak}

In this section, we provide empirical validation of the single-peak structure for \textsc{cosSim}, \textsc{SWIFT}, \textsc{KnapSpec} and \textsc{SpecVLM}. We report both average (across prompts) and prompt-level statistics. For Qwen2.5-VL-7B-Instruct, we use AIME25 prompts for \textsc{cosSim}, \textsc{SWIFT}, and \textsc{KnapSpec}, and Video-MME prompts for \textsc{SpecVLM}. For LLaVA-NeXT-Mistral-7B, all four methods use Video-MME prompts. For these experiments, we modify the suffix instruction appended to each prompt to encourage longer responses and obtain more stable estimates of acceptance rates, time costs, and throughput.

For each model--method pair, the first figure reports (a) mean acceptance rate and time per speculative decoding step, (b) Tokens per Second (TPS), and (c) the fraction of prompts for which each draft model attains the highest measured TPS.
The second figure plots acceptance, time, and TPS for three randomly selected prompts, with $\pm 1$ SE bars.

\subsection{Cosine-Similarity-Based Layer Skipping}
\label{app:single_peak_cossim}

\subsubsection{Qwen2.5-VL-7B-Instruct}

\begin{figure}[!ht]
    \centering
    \subfigure[Acceptance rate and time.]{%
        \includegraphics[width=0.35084\linewidth]{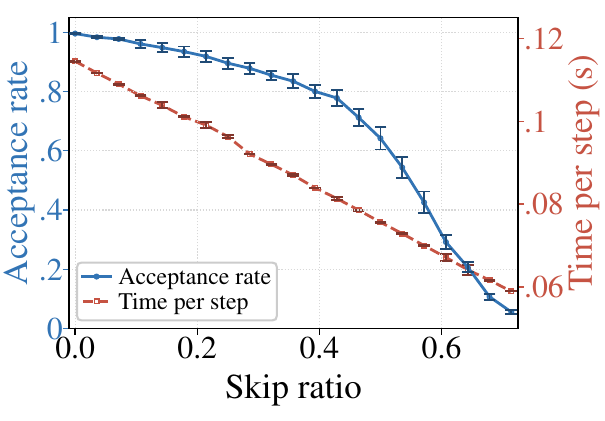}%
        \label{fig:cossim_avg_alpha_time}%
    }\hfill%
    \subfigure[TPS.]{%
        \includegraphics[width=0.31262\linewidth]{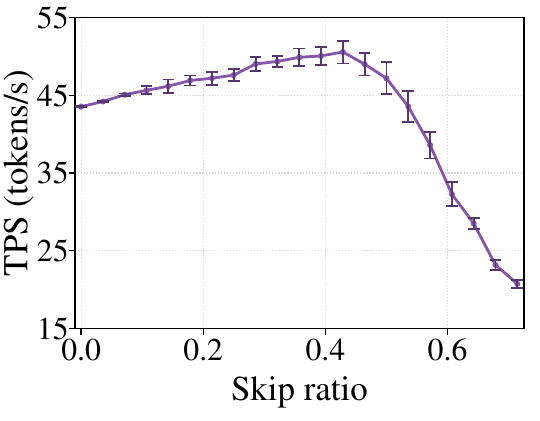}%
        \label{fig:cossim_avg_tps}%
    }\hfill%
    \subfigure[Optimal arm distribution.]{%
        \includegraphics[width=0.31262\linewidth]{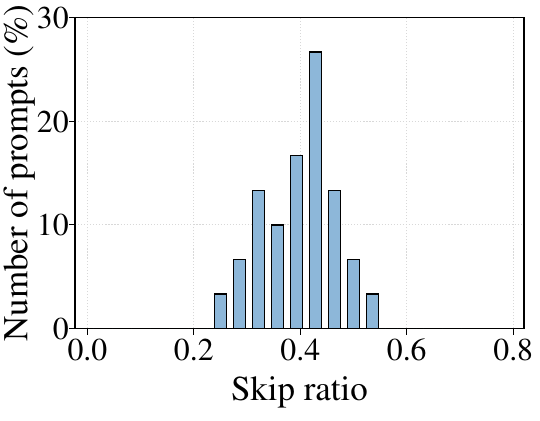}%
        \label{fig:cossim_best_arm_distribution}%
    }
    \caption{Average statistics across AIME25 prompts with \textsc{cosSim} on Qwen2.5-VL-7B-Instruct.
    The horizontal axis is the fraction of skipped layers.
    (a) average acceptance rate and time per speculative decoding step.
    (b) Tokens per Second (TPS).
    (c) Percentage of prompts for which the arm at each skip ratio is oracle best, i.e., attains the highest measured TPS.
    Error bars in (a) and (b) show $\pm 1$ standard error (SE) for the corresponding metric.}
    \label{fig:cossim_average_structure}
\end{figure}

\begin{figure}[!ht]
    \centering
    \subfigure[Acceptance rate.]{%
        \includegraphics[width=0.32536\linewidth]{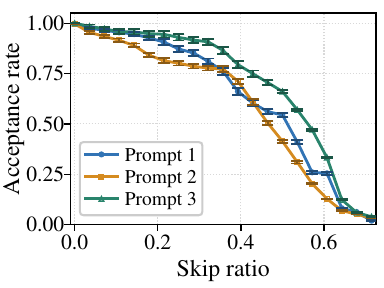}%
        \label{fig:cossim_prompt_alpha}%
    }\hfill%
    \subfigure[Time per step.]{%
        \includegraphics[width=0.32536\linewidth]{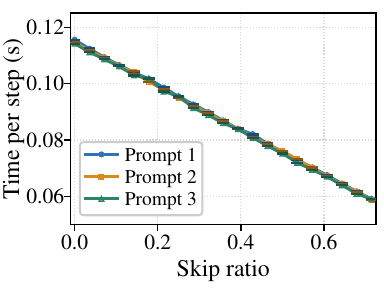}%
        \label{fig:cossim_prompt_time}%
    }\hfill%
    \subfigure[TPS.]{%
        \includegraphics[width=0.32536\linewidth]{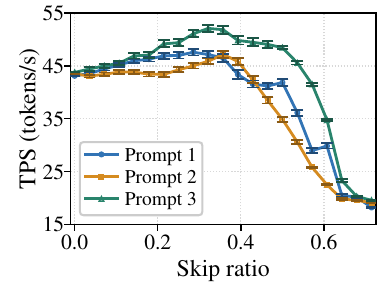}%
        \label{fig:cossim_prompt_tps}%
    }
    \caption{Prompt-level statistics for three randomly selected AIME25 prompts with \textsc{cosSim} on Qwen2.5-VL-7B-Instruct.
    The figures are for (a) acceptance rate, (b) time per speculative decoding step, and (c) TPS, respectively.
    Colors identify the same prompts across all three figures.
    Error bars show $\pm 1$ SE specific to each prompt and metric at each skip ratio.}
    \label{fig:cossim_prompt_structure}
\end{figure}

% Sources: data/qwen25_cosSim_avg_alpha_t_TPS.xlsx and
% data/qwen25_cosSim_perPrompt_alpha_t_TPS.xlsx, sheet AIME25, rows 2--22.
% Each metric is followed by its own supplied standard-error column.
% All means and SEs, including TPS SEs, are used directly without smoothing.
Figure~\ref{fig:cossim_average_structure}(a) shows acceptance declining slowly at modest skip ratios and more sharply under aggressive skipping.
Equivalently, restoring retained computation yields diminishing acceptance gains.
Mean step time is almost perfectly linear, with negligible SEs at the plotted scale, consistent with skipping one attention layer and one MLP layer together in each block.
Average TPS peaks near $0.43$ before falling sharply.
Figure~\ref{fig:cossim_prompt_structure} shows similar timing curves, and different prompt-specific TPS peaks.

\clearpage
\subsubsection{LLaVA-NeXT-Mistral-7B}

\begin{figure}[!ht]
    \centering
    \subfigure[Acceptance rate and time.]{%
        \includegraphics[width=0.35084\linewidth]{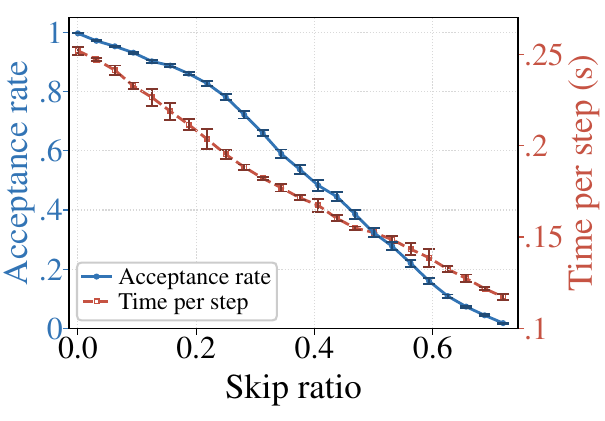}%
        \label{fig:llava16_cossim_avg_alpha_time}%
    }\hfill%
    \subfigure[TPS.]{%
        \includegraphics[width=0.31262\linewidth]{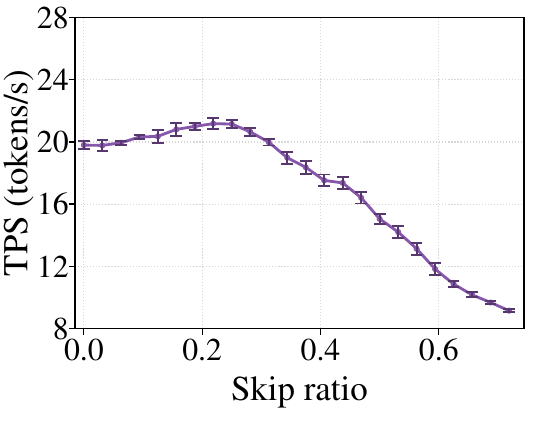}%
        \label{fig:llava16_cossim_avg_tps}%
    }\hfill%
    \subfigure[Optimal arm distribution.]{%
        \includegraphics[width=0.31262\linewidth]{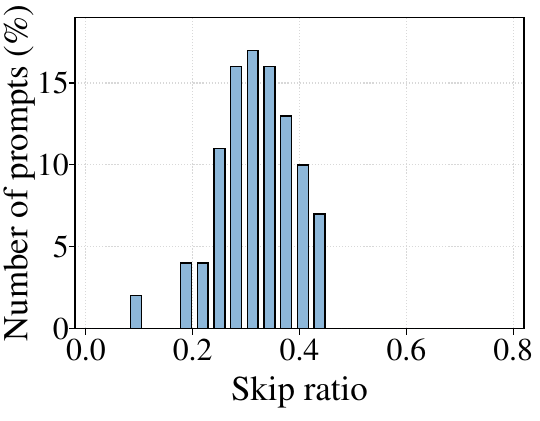}%
        \label{fig:llava16_cossim_best_arm_distribution}%
    }
    \caption{Average statistics across Video-MME prompts with \textsc{cosSim} on LLaVA-NeXT-Mistral-7B.
    The horizontal axis is the fraction of skipped layers.
    (a) average acceptance rate and time per speculative decoding step.
    (b) Tokens per Second (TPS).
    (c) Percentage of prompts for which the arm at each skip ratio is oracle best, i.e., attains the highest measured TPS.
    Error bars in (a) and (b) show $\pm 1$ standard error (SE) for the corresponding metric.}
    \label{fig:llava16_cossim_average_structure}
\end{figure}

\begin{figure}[!ht]
    \centering
    \subfigure[Acceptance rate.]{%
        \includegraphics[width=0.32536\linewidth]{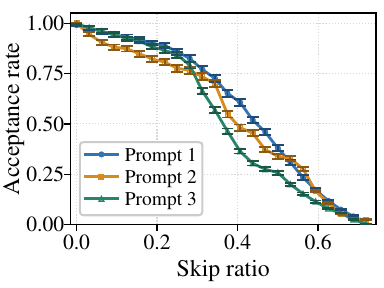}%
        \label{fig:llava16_cossim_prompt_alpha}%
    }\hfill%
    \subfigure[Time per step.]{%
        \includegraphics[width=0.32536\linewidth]{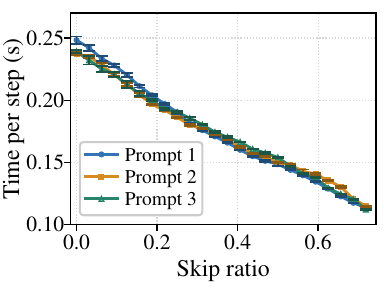}%
        \label{fig:llava16_cossim_prompt_time}%
    }\hfill%
    \subfigure[TPS.]{%
        \includegraphics[width=0.32536\linewidth]{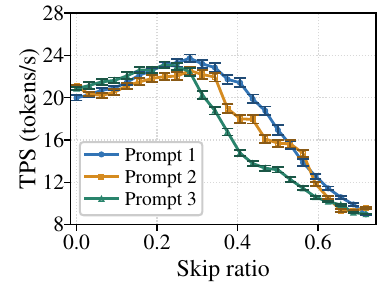}%
        \label{fig:llava16_cossim_prompt_tps}%
    }
    \caption{Prompt-level statistics for three randomly selected Video-MME prompts with \textsc{cosSim} on LLaVA-NeXT-Mistral-7B.
    The figures are for (a) acceptance rate, (b) time per speculative decoding step, and (c) TPS, respectively.
    Colors identify the same prompts across all three figures.
    Error bars show $\pm 1$ SE specific to each prompt and metric at each skip ratio.}
    \label{fig:llava16_cossim_prompt_structure}
\end{figure}

% Sources: data/llava16/llava16_cosSim_avg_alpha_t_TPS.xlsx and
% data/llava16/llava16_cosSim_perPrompt_alpha_t_TPS.xlsx, sheet Video-MME.
% Each metric is followed by its own supplied standard-error column.
% All means and SEs, including TPS SEs, are used directly without smoothing.
Figures~\ref{fig:llava16_cossim_average_structure} and~\ref{fig:llava16_cossim_prompt_structure} exhibit the same acceptance--time tradeoff on LLaVA-NeXT-Mistral-7B.
Step time is linear, with small SEs relative to its magnitude; here acceptance drops slightly more rapidly.

\clearpage
\subsection{SWIFT}
\label{app:single_peak_swift}

\subsubsection{Qwen2.5-VL-7B-Instruct}

\begin{figure}[!ht]
    \centering
    \subfigure[Acceptance rate and time.]{%
        \includegraphics[width=0.35084\linewidth]{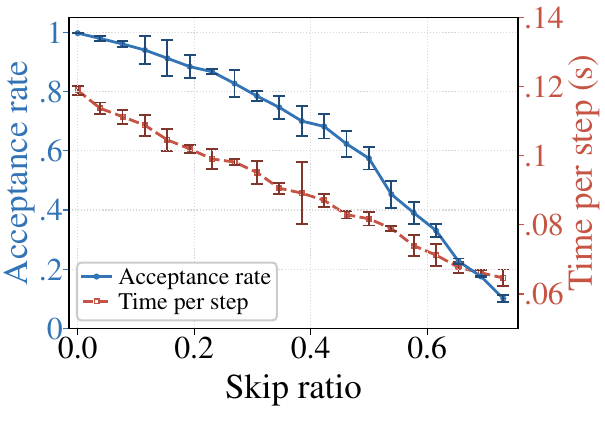}%
        \label{fig:swift_avg_alpha_time}%
    }\hfill%
    \subfigure[TPS.]{%
        \includegraphics[width=0.31262\linewidth]{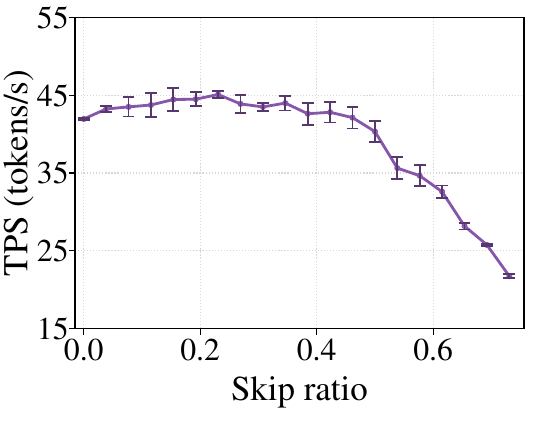}%
        \label{fig:swift_avg_tps}%
    }\hfill%
    \subfigure[Optimal arm distribution.]{%
        \includegraphics[width=0.31262\linewidth]{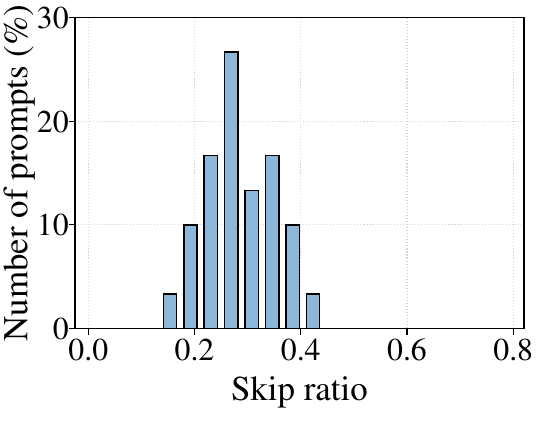}%
        \label{fig:swift_best_arm_distribution}%
    }
    \caption{Average statistics across AIME25 prompts with \textsc{SWIFT} on Qwen2.5-VL-7B-Instruct.
    The horizontal axis is the fraction of skipped layers.
    (a) average acceptance rate and time per speculative decoding step.
    (b) Tokens per Second (TPS).
    (c) Percentage of prompts for which the arm at each skip ratio is oracle best, i.e., attains the highest measured TPS.
    Error bars in (a) and (b) show $\pm 1$ standard error (SE) for the corresponding metric.}
    \label{fig:swift_average_structure}
\end{figure}

% Sources: data/qwen25_swift_avg_alpha_t_TPS.xlsx and
% data/qwen25_swift_perPrompt_alpha_t_TPS.xlsx, sheet AIME25.
% All means and metric-specific SEs are used directly without smoothing.
The acceptance curve in Figure~\ref{fig:swift_average_structure}(a) has the same diminishing-return trend, but step time exhibits more local variation than for \textsc{cosSim}. 
\textsc{SWIFT} obtains its layer ranking through random search and Bayesian optimization and skips attention and MLP layers independently.
Because these layers have different execution costs, equal increments in skip ratio need not yield equal time savings; the changing mix of skipped layers can therefore explain some deviations from a straight line.
Average TPS has a maximum near $0.23$, with small local fluctuations around the overall single-peak trend.

\begin{figure}[!ht]
    \centering
    \subfigure[Acceptance rate.]{%
        \includegraphics[width=0.32536\linewidth]{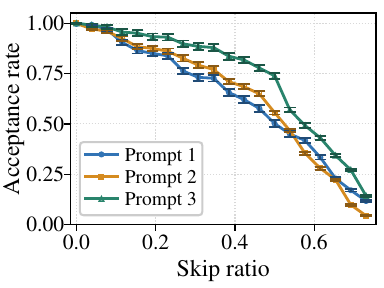}%
        \label{fig:swift_prompt_alpha}%
    }\hfill%
    \subfigure[Time per step.]{%
        \includegraphics[width=0.32536\linewidth]{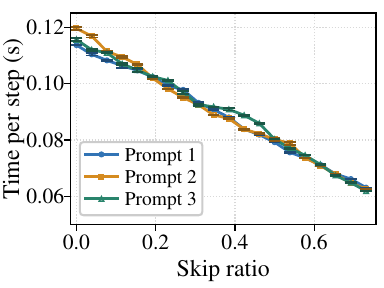}%
        \label{fig:swift_prompt_time}%
    }\hfill%
    \subfigure[TPS.]{%
        \includegraphics[width=0.32536\linewidth]{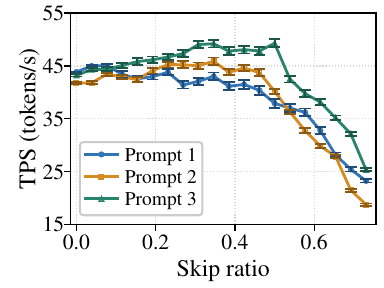}%
        \label{fig:swift_prompt_tps}%
    }
    \caption{Prompt-level statistics for three randomly selected AIME25 prompts with \textsc{SWIFT} on Qwen2.5-VL-7B-Instruct.
    The figures are for (a) acceptance rate, (b) time per speculative decoding step, and (c) TPS, respectively.
    Colors identify the same prompts across all three figures.
    Error bars show $\pm 1$ SE specific to each prompt and metric at each skip ratio.}
    \label{fig:swift_prompt_structure}
\end{figure}

\clearpage
\subsubsection{LLaVA-NeXT-Mistral-7B}

% Sources: data/llava16/llava16_swift_avg_alpha_t_TPS.xlsx and
% data/llava16/llava16_swift_perPrompt_alpha_t_TPS.xlsx, sheet Video-MME.
% All means and metric-specific SEs are used directly without smoothing.

\begin{figure}[!ht]
    \centering
    \subfigure[Acceptance rate and time.]{%
        \includegraphics[width=0.35084\linewidth]{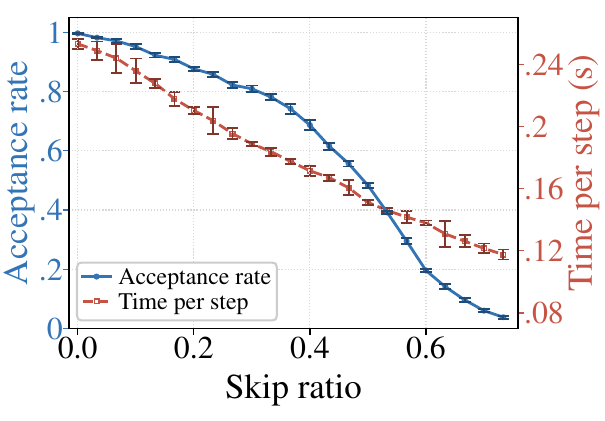}%
        \label{fig:llava16_swift_avg_alpha_time}%
    }\hfill%
    \subfigure[TPS.]{%
        \includegraphics[width=0.31262\linewidth]{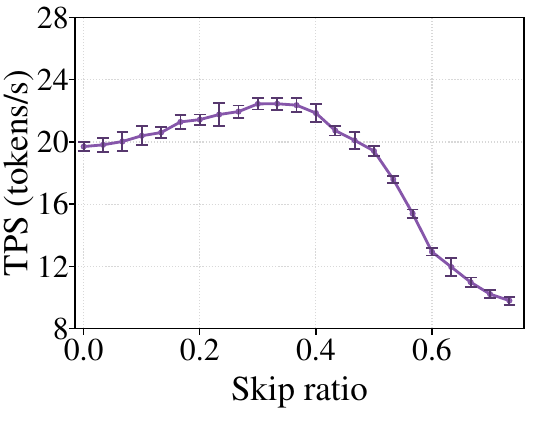}%
        \label{fig:llava16_swift_avg_tps}%
    }\hfill%
    \subfigure[Optimal arm distribution.]{%
        \includegraphics[width=0.31262\linewidth]{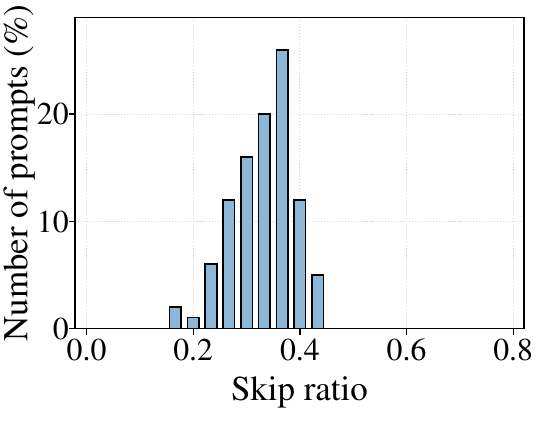}%
        \label{fig:llava16_swift_best_arm_distribution}%
    }
    \caption{Average statistics across Video-MME prompts with \textsc{SWIFT} on LLaVA-NeXT-Mistral-7B.
    The horizontal axis is the fraction of skipped layers.
    (a) average acceptance rate and time per speculative decoding step.
    (b) Tokens per Second (TPS).
    (c) Percentage of prompts for which the arm at each skip ratio is oracle best, i.e., attains the highest measured TPS.
    Error bars in (a) and (b) show $\pm 1$ standard error (SE) for the corresponding metric.}
    \label{fig:llava16_swift_average_structure}
\end{figure}

\begin{figure}[!ht]
    \centering
    \subfigure[Acceptance rate.]{%
        \includegraphics[width=0.32536\linewidth]{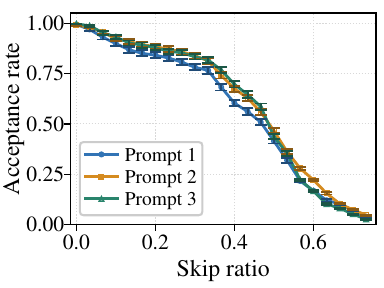}%
        \label{fig:llava16_swift_prompt_alpha}%
    }\hfill%
    \subfigure[Time per step.]{%
        \includegraphics[width=0.32536\linewidth]{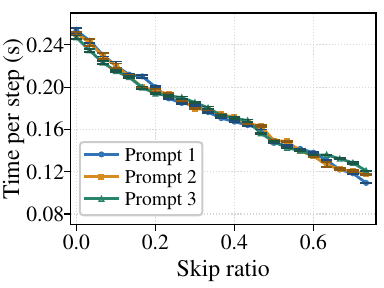}%
        \label{fig:llava16_swift_prompt_time}%
    }\hfill%
    \subfigure[TPS.]{%
        \includegraphics[width=0.32536\linewidth]{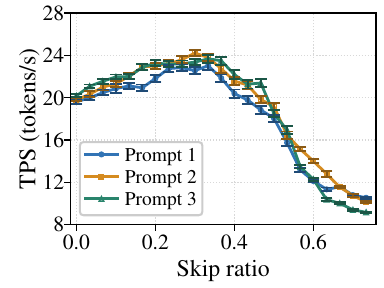}%
        \label{fig:llava16_swift_prompt_tps}%
    }
    \caption{Prompt-level statistics for three randomly selected Video-MME prompts with \textsc{SWIFT} on LLaVA-NeXT-Mistral-7B.
    The figures are for (a) acceptance rate, (b) time per speculative decoding step, and (c) TPS, respectively.
    Colors identify the same prompts across all three figures.
    Error bars show $\pm 1$ SE specific to each prompt and metric at each skip ratio.}
    \label{fig:llava16_swift_prompt_structure}
\end{figure}

\clearpage
\subsection{KnapSpec}
\label{app:single_peak_knapspec}

\subsubsection{Qwen2.5-VL-7B-Instruct}

% Sources: data/qwen25_knapspec_avg_alpha_t_TPS.xlsx and
% data/qwen25_knapspec_perPrompt_alpha_t_TPS.xlsx, sheet AIME25.
% All means and metric-specific SEs are used directly without smoothing.

For \textsc{KnapSpec}, the \emph{budget skip ratio} measures the normalized saved-latency budget used to construct a draft model. As described in Appendix~\ref{app:existing_draft_construction}, attention and MLP layers have separate costs based on their profiled latencies, and skipping a layer contributes its discretized cost to this budget. A larger budget skip ratio therefore corresponds to a smaller retained-latency budget $b$ and more aggressive computation reduction. Since layers have different costs, this ratio generally differs from the fraction of layers skipped.

\begin{figure}[!ht]
    \centering
    \subfigure[Acceptance rate and time.]{%
        \includegraphics[width=0.35084\linewidth]{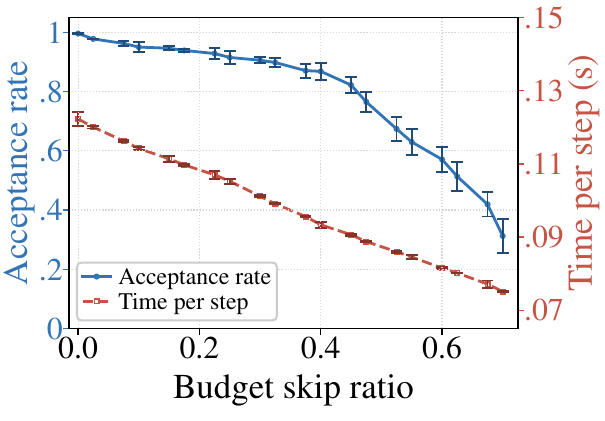}%
        \label{fig:knapspec_avg_alpha_time}%
    }\hfill%
    \subfigure[TPS.]{%
        \includegraphics[width=0.31262\linewidth]{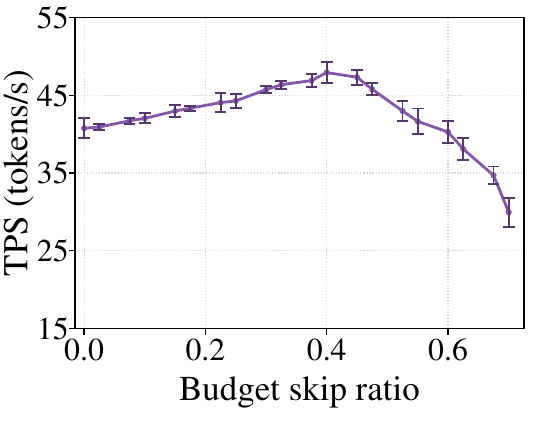}%
        \label{fig:knapspec_avg_tps}%
    }\hfill%
    \subfigure[Optimal arm distribution.]{%
        \includegraphics[width=0.31262\linewidth]{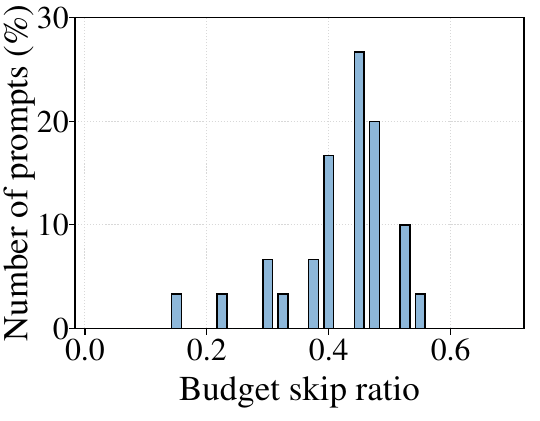}%
        \label{fig:knapspec_best_arm_distribution}%
    }
    \caption{Average statistics across AIME25 prompts with \textsc{KnapSpec} on Qwen2.5-VL-7B-Instruct.
    The horizontal axis is the budget skip ratio.
    (a) average acceptance rate and time per speculative decoding step.
    (b) Tokens per Second (TPS).
    (c) Percentage of prompts for which the arm at each budget skip ratio is oracle best, i.e., attains the highest measured TPS.
    Error bars in (a) and (b) show $\pm 1$ standard error (SE) for the corresponding metric.}
    \label{fig:knapspec_average_structure}
\end{figure}

\begin{figure}[!ht]
    \centering
    \subfigure[Acceptance rate.]{%
        \includegraphics[width=0.32536\linewidth]{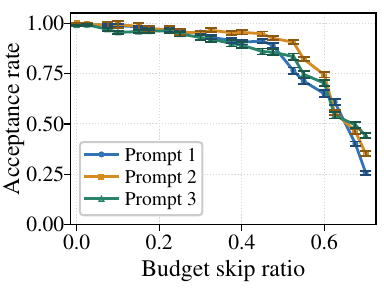}%
        \label{fig:knapspec_prompt_alpha}%
    }\hfill%
    \subfigure[Time per step.]{%
        \includegraphics[width=0.32536\linewidth]{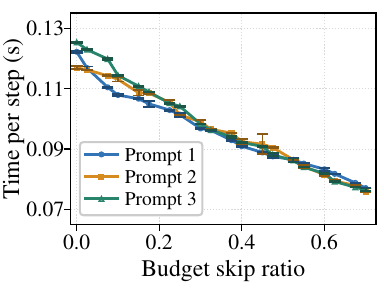}%
        \label{fig:knapspec_prompt_time}%
    }\hfill%
    \subfigure[TPS.]{%
        \includegraphics[width=0.32536\linewidth]{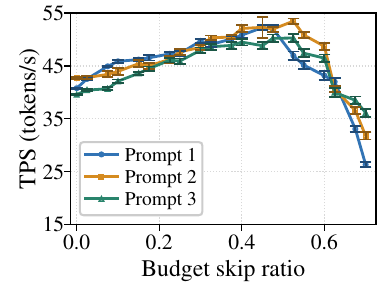}%
        \label{fig:knapspec_prompt_tps}%
    }
    \caption{Prompt-level statistics for three randomly selected AIME25 prompts with \textsc{KnapSpec} on Qwen2.5-VL-7B-Instruct.
    The figures are for (a) acceptance rate, (b) time per speculative decoding step, and (c) TPS, respectively.
    Colors identify the same prompts across all three figures.
    Error bars show $\pm 1$ SE specific to each prompt and metric at each budget skip ratio.}
    \label{fig:knapspec_prompt_structure}
\end{figure}

In Figures~\ref{fig:knapspec_average_structure} and~\ref{fig:knapspec_prompt_structure}, acceptance decreases particularly slowly at small budget skip ratios.
This is consistent with \textsc{KnapSpec}'s effective selection of layers that preserve similarity to the target model's hidden states under a latency budget.
Step time is reasonably linear, as latency budget is designed to align with model decoding time.
We remark that some small local timing deviations can arise from the differences between profiled layer costs and their execution costs as the context grows.

\clearpage
\subsubsection{LLaVA-NeXT-Mistral-7B}

\begin{figure}[!ht]
    \centering
    \subfigure[Acceptance rate and time.]{%
        \includegraphics[width=0.35084\linewidth]{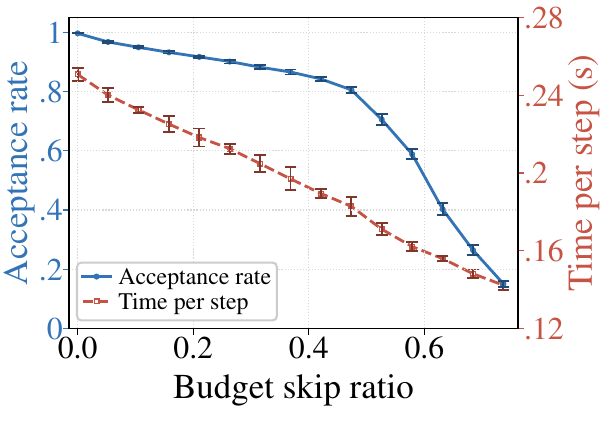}%
        \label{fig:llava16_knapspec_avg_alpha_time}%
    }\hfill%
    \subfigure[TPS.]{%
        \includegraphics[width=0.31262\linewidth]{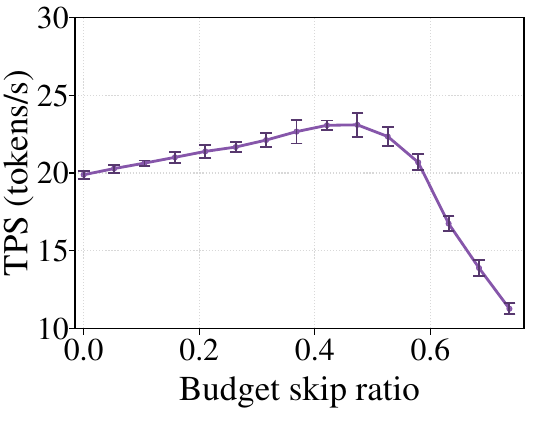}%
        \label{fig:llava16_knapspec_avg_tps}%
    }\hfill%
    \subfigure[Optimal arm distribution.]{%
        \includegraphics[width=0.31262\linewidth]{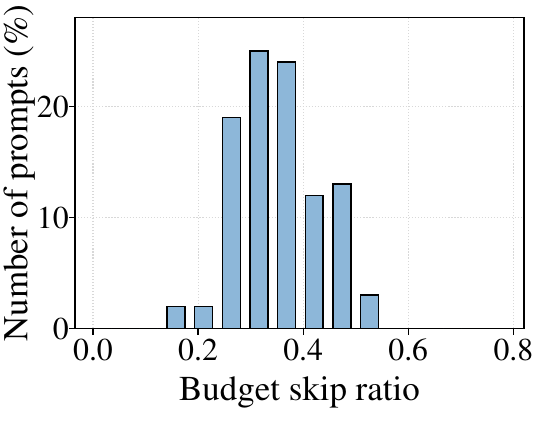}%
        \label{fig:llava16_knapspec_best_arm_distribution}%
    }
    \caption{Average statistics across Video-MME prompts with \textsc{KnapSpec} on LLaVA-NeXT-Mistral-7B.
    The horizontal axis is the budget skip ratio.
    (a) average acceptance rate and time per speculative decoding step.
    (b) Tokens per Second (TPS).
    (c) Percentage of prompts for which the arm at each budget skip ratio is oracle best, i.e., attains the highest measured TPS.
    Error bars in (a) and (b) show $\pm 1$ standard error (SE) for the corresponding metric.}
    \label{fig:llava16_knapspec_average_structure}
\end{figure}

\begin{figure}[!ht]
    \centering
    \subfigure[Acceptance rate.]{%
        \includegraphics[width=0.32536\linewidth]{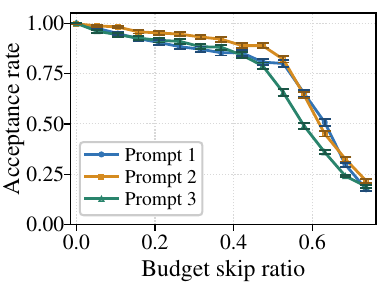}%
        \label{fig:llava16_knapspec_prompt_alpha}%
    }\hfill%
    \subfigure[Time per step.]{%
        \includegraphics[width=0.32536\linewidth]{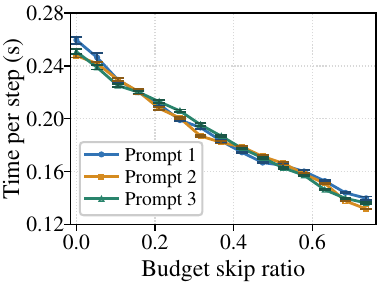}%
        \label{fig:llava16_knapspec_prompt_time}%
    }\hfill%
    \subfigure[TPS.]{%
        \includegraphics[width=0.32536\linewidth]{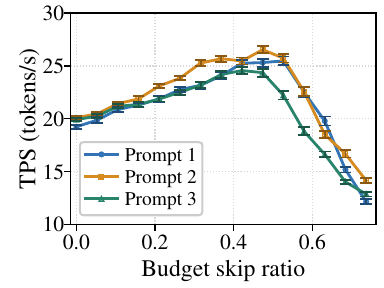}%
        \label{fig:llava16_knapspec_prompt_tps}%
    }
    \caption{Prompt-level statistics for three randomly selected Video-MME prompts with \textsc{KnapSpec} on LLaVA-NeXT-Mistral-7B.
    The figures are for (a) acceptance rate, (b) time per speculative decoding step, and (c) TPS, respectively.
    Colors identify the same prompts across all three figures.
    Error bars show $\pm 1$ SE specific to each prompt and metric at each budget ratio.}
    \label{fig:llava16_knapspec_prompt_structure}
\end{figure}

% Sources: data/llava16/llava16_knapspec_avg_alpha_t_TPS.xlsx and
% data/llava16/llava16_knapspec_perPrompt_alpha_t_TPS.xlsx, sheet Video-MME.
% All means and metric-specific SEs are used directly without smoothing.
Figures~\ref{fig:llava16_knapspec_average_structure} and~\ref{fig:llava16_knapspec_prompt_structure} show the same slow acceptance decline and approximately linear time reduction.

\clearpage
\subsection{SpecVLM}
\label{app:single_peak_specvlm}

\subsubsection{Qwen2.5-VL-7B-Instruct}

\begin{figure}[!ht]
    \centering
    \subfigure[Acceptance rate and time.]{%
        \includegraphics[width=0.35084\linewidth]{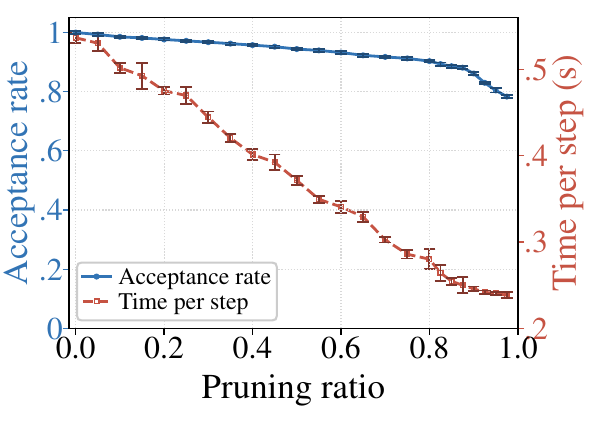}%
        \label{fig:specvlm_avg_alpha_time}%
    }\hfill%
    \subfigure[TPS.]{%
        \includegraphics[width=0.31262\linewidth]{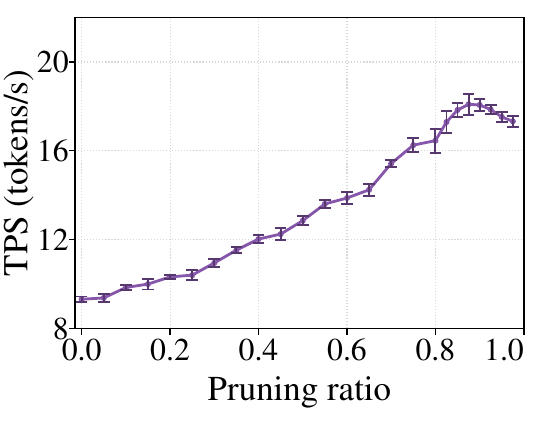}%
        \label{fig:specvlm_avg_tps}%
    }\hfill%
    \subfigure[Optimal arm distribution.]{%
        \includegraphics[width=0.31262\linewidth]{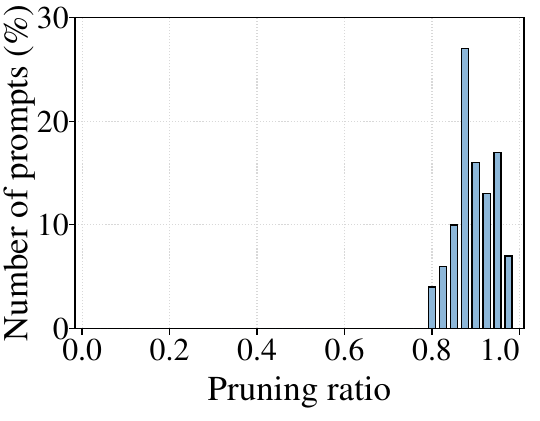}%
        \label{fig:specvlm_best_arm_distribution}%
    }
    \caption{Average statistics across Video-MME prompts with \textsc{SpecVLM} on Qwen2.5-VL-7B-Instruct.
    The horizontal axis in (c) is the pruning ratio.
    (a) average acceptance rate and time per speculative decoding step.
    (b) Tokens per Second (TPS).
    (c) Percentage of prompts for which the arm at each pruning ratio is oracle best, i.e., attains the highest measured TPS.
    Error bars in (a) and (b) show $\pm 1$ standard error (SE) for the corresponding metric.}
    \label{fig:specvlm_average_structure}
\end{figure}

\begin{figure}[!ht]
    \centering
    \subfigure[Acceptance rate.]{%
        \includegraphics[width=0.32536\linewidth]{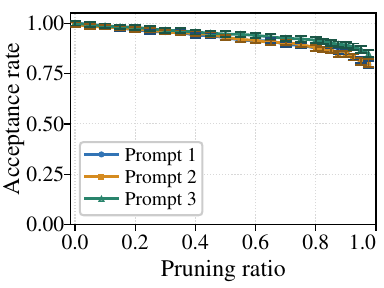}%
        \label{fig:specvlm_prompt_alpha}%
    }\hfill%
    \subfigure[Time per step.]{%
        \includegraphics[width=0.32536\linewidth]{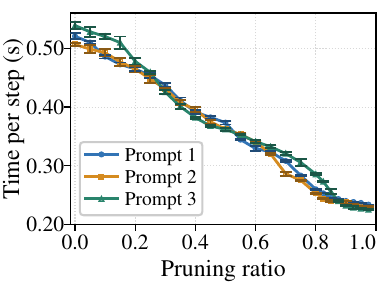}%
        \label{fig:specvlm_prompt_time}%
    }\hfill%
    \subfigure[TPS.]{%
        \includegraphics[width=0.32536\linewidth]{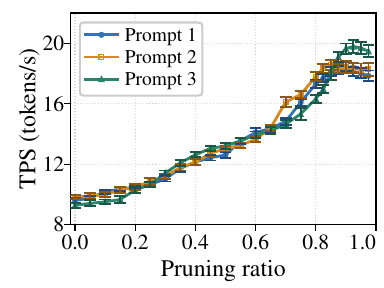}%
        \label{fig:specvlm_prompt_tps}%
    }
    \caption{Prompt-level statistics for three randomly selected Video-MME prompts with \textsc{SpecVLM} on Qwen2.5-VL-7B-Instruct.
    The figures are for (a) acceptance rate, (b) time per speculative decoding step, and (c) TPS, respectively.
    Colors identify the same prompts across all three figures.
    Error bars show $\pm 1$ SE specific to each prompt and metric at each pruning ratio.}
    \label{fig:specvlm_prompt_structure}
\end{figure}

% Sources: data/qwen25_specvlm_avg_alpha_t_TPS.xlsx and
% data/qwen25_specvlm_perPrompt_alpha_t_TPS.xlsx, sheet AIME25 (Video-MME prompts).
% All means and metric-specific SEs are used directly without smoothing.
\textsc{SpecVLM} varies visual-token pruning rather than layer skipping.
Figures~\ref{fig:specvlm_average_structure} and~\ref{fig:specvlm_prompt_structure} show that acceptance decreases slowly over most of this range, remaining around $0.90$ on average even at $80\%$ pruning.
This is consistent with retaining highly attended visual tokens and preserving spatial coverage.

Step time decreases approximately linearly at first and flattens slightly at high pruning ratios.
This is because un-pruned language-prompt tokens, generated history, and full-context target verification remain, limiting the total time savings as visual attention becomes a smaller part of the workload.

\clearpage
\subsubsection{LLaVA-NeXT-Mistral-7B}

\begin{figure}[!ht]
    \centering
    \subfigure[Acceptance rate and time.]{%
        \includegraphics[width=0.35084\linewidth]{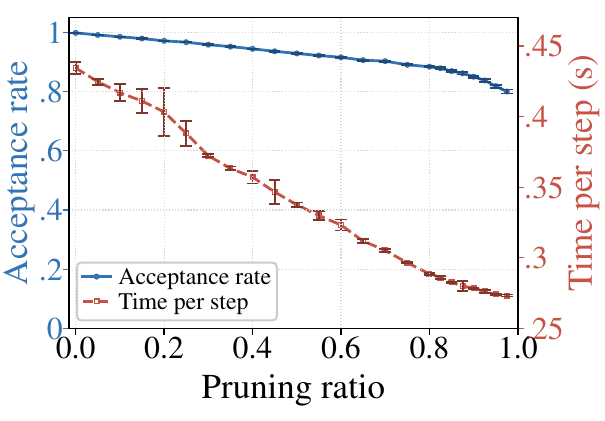}%
        \label{fig:llava16_specvlm_avg_alpha_time}%
    }\hfill%
    \subfigure[TPS.]{%
        \includegraphics[width=0.31262\linewidth]{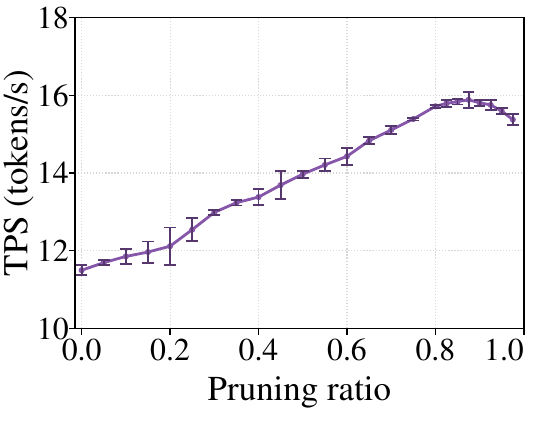}%
        \label{fig:llava16_specvlm_avg_tps}%
    }\hfill%
    \subfigure[Optimal arm distribution.]{%
        \includegraphics[width=0.31262\linewidth]{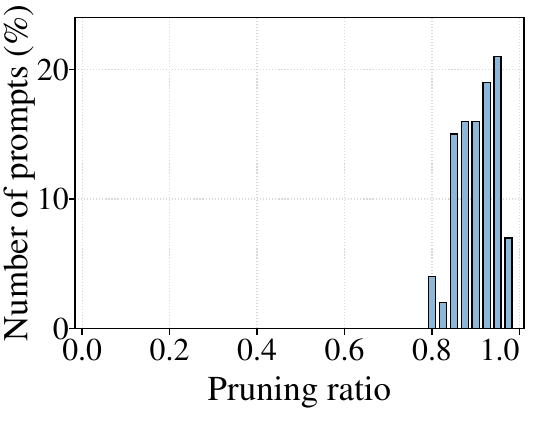}%
        \label{fig:llava16_specvlm_best_arm_distribution}%
    }
    \caption{Average statistics across Video-MME prompts with \textsc{SpecVLM} on LLaVA-NeXT-Mistral-7B.
    The horizontal axis is the pruning ratio.
    (a) average acceptance rate and time per speculative decoding step.
    (b) Tokens per Second (TPS).
    (c) Percentage of prompts for which the arm at each pruning ratio is oracle best, i.e., attains the highest measured TPS.
    Error bars in (a) and (b) show $\pm 1$ standard error (SE) for the corresponding metric.}
    \label{fig:llava16_specvlm_average_structure}
\end{figure}

\begin{figure}[!ht]
    \centering
    \subfigure[Acceptance rate.]{%
        \includegraphics[width=0.32536\linewidth]{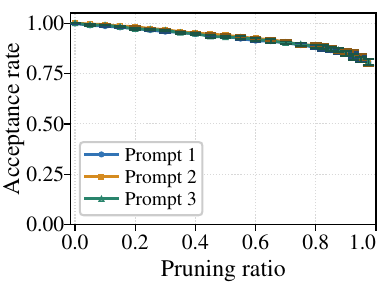}%
        \label{fig:llava16_specvlm_prompt_alpha}%
    }\hfill%
    \subfigure[Time per step.]{%
        \includegraphics[width=0.32536\linewidth]{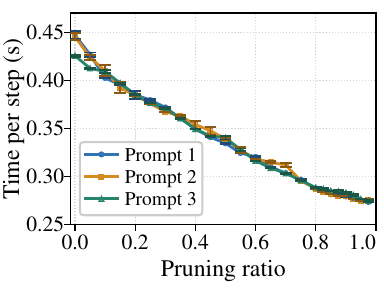}%
        \label{fig:llava16_specvlm_prompt_time}%
    }\hfill%
    \subfigure[TPS.]{%
        \includegraphics[width=0.32536\linewidth]{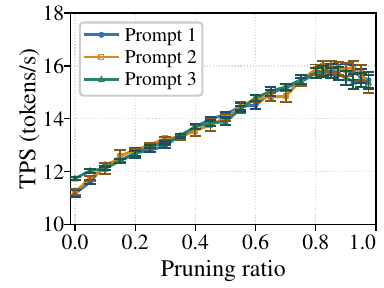}%
        \label{fig:llava16_specvlm_prompt_tps}%
    }
    \caption{Prompt-level statistics for three randomly selected Video-MME prompts with \textsc{SpecVLM} on LLaVA-NeXT-Mistral-7B.
    The figures are for (a) acceptance rate, (b) time per speculative decoding step, and (c) TPS, respectively.
    Colors identify the same prompts across all three figures.
    Error bars show $\pm 1$ SE specific to each prompt and metric at each pruning ratio.}
    \label{fig:llava16_specvlm_prompt_structure}
\end{figure}

% Sources: data/llava16/llava16_specvlm_avg_alpha_t_TPS.xlsx and
% data/llava16/llava16_specvlm_perPrompt_alpha_t_TPS.xlsx, sheet Video-MME.
% All means and metric-specific SEs are used directly without smoothing.
The LLaVA results in Figures~\ref{fig:llava16_specvlm_average_structure} and~\ref{fig:llava16_specvlm_prompt_structure} show the same slow acceptance decline and diminishing time savings at high pruning ratios.
At $80\%$ pruning, average acceptance remains approximately $0.88$, and the average TPS also peaks near $0.88$ pruning.
Both models therefore attain their best average throughput while retaining only a small fraction of visual tokens.

\section{Ablation Studies}
\label{app:ablation}

This appendix evaluates the robustness of \ourmethod{} to different generation settings.
Subsection~\ref{app:ablation_draft_length} examines the effect of draft length on decoding speedup.
Subsection~\ref{app:ablation_output_length} studies how output length affects the amortization of construction and exploration costs.
Subsection~\ref{app:ablation_input_size} evaluates sensitivity to the number of visual input tokens.
Subsection~\ref{app:ablation_reset} compares periodic resets of the arm-pull and reward history with resetting only between prompts.

\subsection{Draft Length}
\label{app:ablation_draft_length}

\begin{figure}[!ht]
    \centering
    \subfigure[\textsc{cosSim}.]{%
        \includegraphics[width=0.32\linewidth]{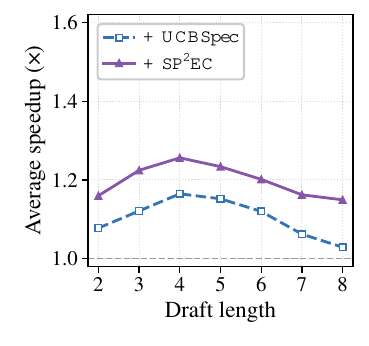}%
        \label{fig:ablation_draft_length_cossim}%
    }\hfill%
    \subfigure[\textsc{KnapSpec}.]{%
        \includegraphics[width=0.32\linewidth]{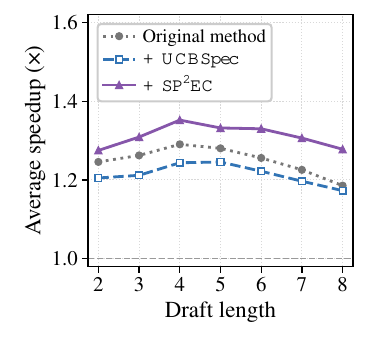}%
        \label{fig:ablation_draft_length_knapspec}%
    }\hfill%
    \subfigure[\textsc{SpecVLM}.]{%
        \includegraphics[width=0.32\linewidth]{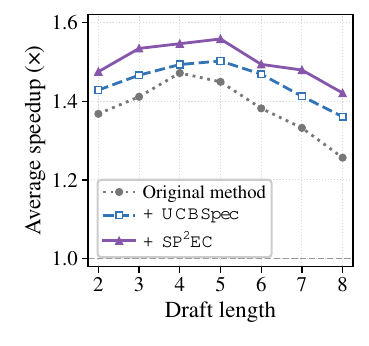}%
        \label{fig:ablation_draft_length_specvlm}%
    }
    \caption{Effect of draft length on average speedup with Qwen2.5-VL-7B-Instruct.
    The panels compare UCBSpec and SP$^2$EC applied to (a) \textsc{cosSim}, (b) \textsc{KnapSpec}, and (c) \textsc{SpecVLM}.
    The original methods are also shown for \textsc{KnapSpec} and \textsc{SpecVLM}.
    Speedup is normalized to autoregressive decoding, indicated by the dashed horizontal line at $1\times$.}
    \label{fig:ablation_draft_length}
\end{figure}

\begin{figure}[!ht]
    \centering
    \subfigure[\textsc{cosSim}.]{%
        \includegraphics[width=0.32\linewidth]{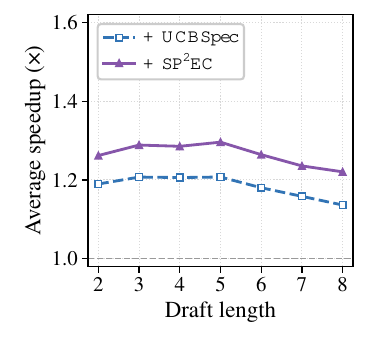}%
        \label{fig:llava16_ablation_draft_length_cossim}%
    }\hfill%
    \subfigure[\textsc{KnapSpec}.]{%
        \includegraphics[width=0.32\linewidth]{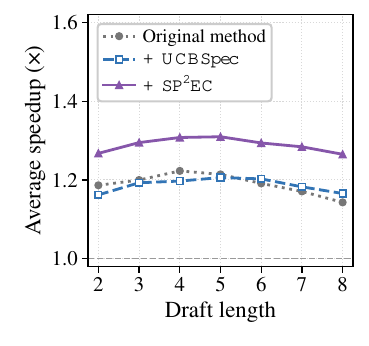}%
        \label{fig:llava16_ablation_draft_length_knapspec}%
    }\hfill%
    \subfigure[\textsc{SpecVLM}.]{%
        \includegraphics[width=0.32\linewidth]{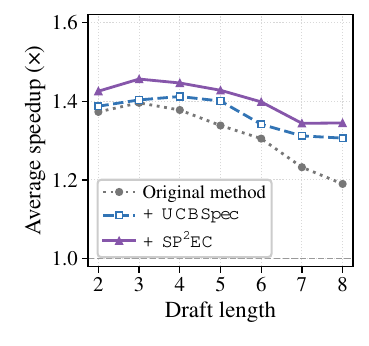}%
        \label{fig:llava16_ablation_draft_length_specvlm}%
    }
    \caption{Effect of draft length on average speedup with LLaVA-NeXT-Mistral-7B.
    The panels compare UCBSpec and SP$^2$EC applied to (a) \textsc{cosSim}, (b) \textsc{KnapSpec}, and (c) \textsc{SpecVLM}.
    The original methods are also shown for \textsc{KnapSpec} and \textsc{SpecVLM}.
    Speedup is normalized to autoregressive decoding, indicated by the dashed horizontal line at $1\times$.}
    \label{fig:llava16_ablation_draft_length}
\end{figure}

Figures~\ref{fig:ablation_draft_length} and~\ref{fig:llava16_ablation_draft_length} examine the effect of draft length $L$ on average speedup.
Increasing $L$ has two opposing effects: a longer draft allows the verifier to accept more tokens in one speculative round, but requires more sequential draft forward passes, increases verification cost, and wastes more computation when an early draft token is rejected.
Across both models, $L=4$--$6$ provides a good practical balance: all three constructions with \ourmethod{} remain within $5\%$ of their respective highest measured speedups throughout this range.
The exact optimum varies across constructions and models, while the shorter and longer drafts tested generally perform worse.
These results support $L=4$ as a robust default.

% Source: data/qwen25_ablation_draft_length.xlsx, sheet 'Average SpeedUp', A1:C64.
% Mapping: ucb_basic/sp2ec_basic correspond to cosSim with UCBSpec/SP2EC.
% The workbook supplies no original cosSim curve or uncertainty estimates.

% Source: data/llava16/llava16_ablation_draft_length.xlsx.

\clearpage
\subsection{Output Length}
\label{app:ablation_output_length}

\begin{figure}[!ht]
    \centering
    \subfigure[\textsc{cosSim}.]{%
        \includegraphics[width=0.32\linewidth]{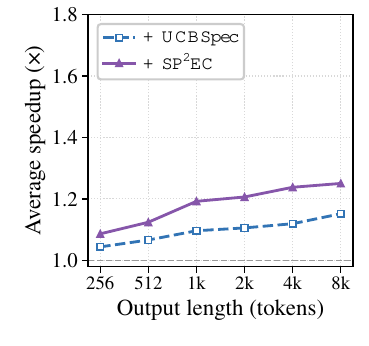}%
        \label{fig:ablation_output_length_cossim}%
    }\hfill%
    \subfigure[\textsc{KnapSpec}.]{%
        \includegraphics[width=0.32\linewidth]{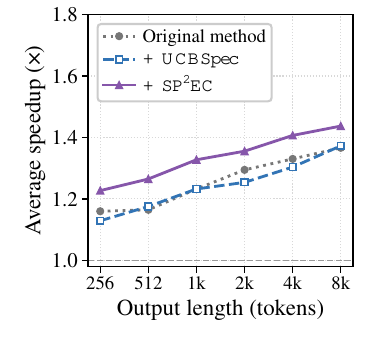}%
        \label{fig:ablation_output_length_knapspec}%
    }\hfill%
    \subfigure[\textsc{SpecVLM}.]{%
        \includegraphics[width=0.32\linewidth]{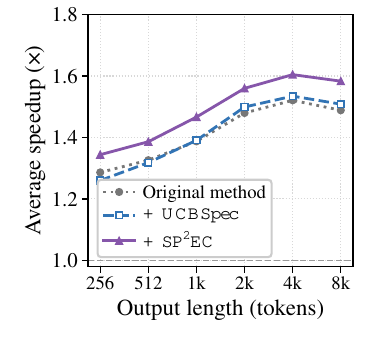}%
        \label{fig:ablation_output_length_specvlm}%
    }
    \caption{Effect of output length on average speedup with Qwen2.5-VL-7B-Instruct.
    The panels compare UCBSpec and SP$^2$EC applied to (a) \textsc{cosSim}, (b) \textsc{KnapSpec}, and (c) \textsc{SpecVLM}.
    The original methods are also shown for \textsc{KnapSpec} and \textsc{SpecVLM}.
    The horizontal axis shows output length in tokens on a base-2 logarithmic scale, with $1\mathrm{k}=1024$.
    Speedup is normalized to autoregressive decoding, indicated by the dashed horizontal line at $1\times$.}
    \label{fig:ablation_output_length}
\end{figure}

\begin{figure}[!ht]
    \centering
    \subfigure[\textsc{cosSim}.]{%
        \includegraphics[width=0.32\linewidth]{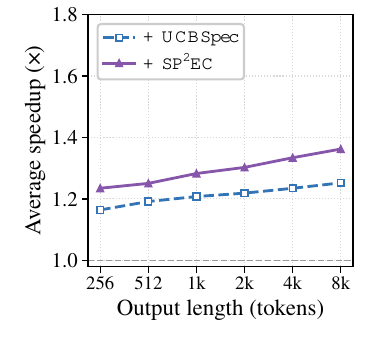}%
        \label{fig:llava16_ablation_output_length_cossim}%
    }\hfill%
    \subfigure[\textsc{KnapSpec}.]{%
        \includegraphics[width=0.32\linewidth]{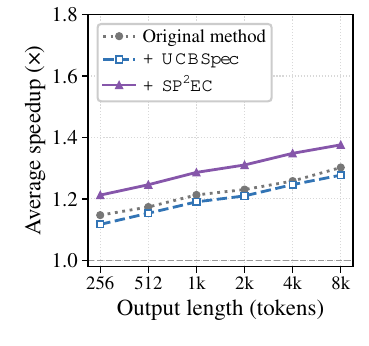}%
        \label{fig:llava16_ablation_output_length_knapspec}%
    }\hfill%
    \subfigure[\textsc{SpecVLM}.]{%
        \includegraphics[width=0.32\linewidth]{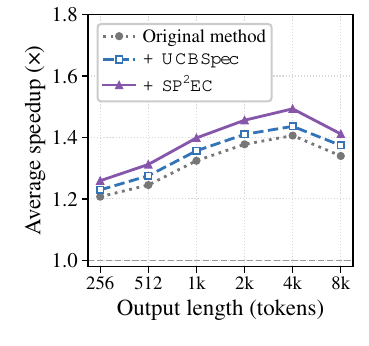}%
        \label{fig:llava16_ablation_output_length_specvlm}%
    }
    \caption{Effect of output length on average speedup with LLaVA-NeXT-Mistral-7B.
    The panels compare UCBSpec and SP$^2$EC applied to (a) \textsc{cosSim}, (b) \textsc{KnapSpec}, and (c) \textsc{SpecVLM}.
    The original methods are also shown for \textsc{KnapSpec} and \textsc{SpecVLM}.
    The horizontal axis shows output length in tokens on a base-2 logarithmic scale, with $1\mathrm{k}=1024$.
    Speedup is normalized to autoregressive decoding, indicated by the dashed horizontal line at $1\times$.}
    \label{fig:llava16_ablation_output_length}
\end{figure}

We modify the input prompts to elicit longer responses and evaluate output lengths from $256$ to $8192$ tokens in Figures~\ref{fig:ablation_output_length} and~\ref{fig:llava16_ablation_output_length}.
Speedup generally improves with a longer generation horizon, as it amortizes prompt-level arm-construction, warm-up, and initial exploration costs.
A longer horizon also allows more decoding steps to use well-performing arms after initial exploration.
Nevertheless, all six model--construction combinations with \ourmethod{} can provide considerable speedups even at $256$ tokens.

Forcing unnaturally long outputs can produce repetitive, simpler continuations, changing the workload and limiting how representative their speedups are of natural task responses.
In the main experiments, outputs are typically around $1000$--$2000$ tokens on the video and image datasets and below $10{,}000$ tokens on AIME25.
We therefore do not treat the longest forced outputs as the primary performance measure.

% Source: data/qwen25_ablation_output_length.xlsx, sheet 'Average SpeedUp', A1:C59.
% C38 stores '1.2504x', plotted as a speedup of 1.2504.
% All methods use the shared 256--8192-token range; 128-token observations are omitted.

% Source: data/llava16/llava16_ablation_output_length.xlsx.xlsx.

\clearpage
\subsection{Input Size}
\label{app:ablation_input_size}

\begin{figure}[!ht]
    \centering
    \subfigure[\textsc{cosSim}.]{%
        \includegraphics[width=0.32\linewidth]{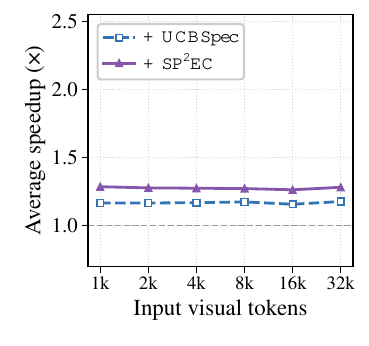}%
        \label{fig:ablation_input_size_cossim}%
    }\hfill%
    \subfigure[\textsc{KnapSpec}.]{%
        \includegraphics[width=0.32\linewidth]{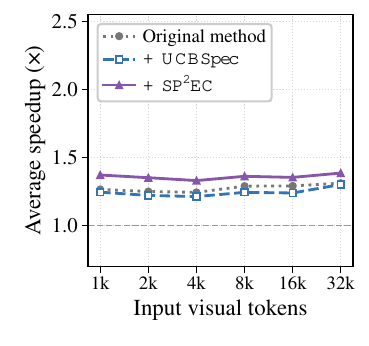}%
        \label{fig:ablation_input_size_knapspec}%
    }\hfill%
    \subfigure[\textsc{SpecVLM}.]{%
        \includegraphics[width=0.32\linewidth]{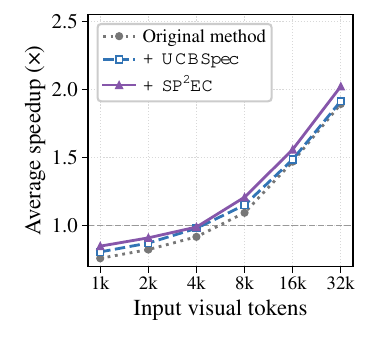}%
        \label{fig:ablation_input_size_specvlm}%
    }
    \caption{Effect of input size on average speedup with Qwen2.5-VL-7B-Instruct.
    The panels compare UCBSpec and SP$^2$EC applied to (a) \textsc{cosSim}, (b) \textsc{KnapSpec}, and (c) \textsc{SpecVLM}.
    The original methods are also shown for \textsc{KnapSpec} and \textsc{SpecVLM}.
    The horizontal axis shows the visual token budget on a base-2 logarithmic scale, with $1\mathrm{k}=1024$ tokens.
    Speedup is normalized to autoregressive decoding, indicated by the dashed horizontal line at $1\times$. AIME25 is excluded when running this comparison.}
    \label{fig:ablation_input_size}
\end{figure}

\begin{figure}[!ht]
    \centering
    \subfigure[\textsc{cosSim}.]{%
        \includegraphics[width=0.32\linewidth]{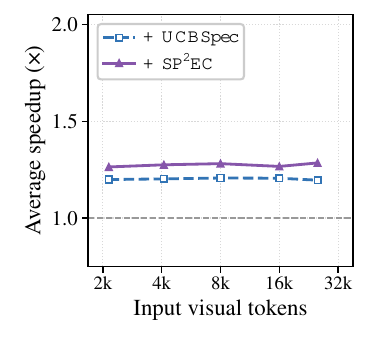}%
        \label{fig:llava16_ablation_input_size_cossim}%
    }\hfill%
    \subfigure[\textsc{KnapSpec}.]{%
        \includegraphics[width=0.32\linewidth]{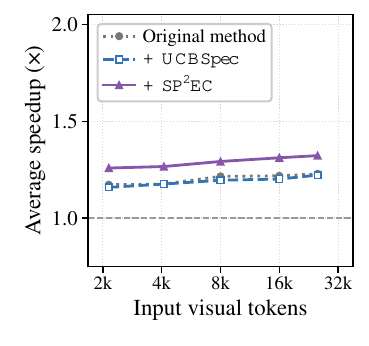}%
        \label{fig:llava16_ablation_input_size_knapspec}%
    }\hfill%
    \subfigure[\textsc{SpecVLM}.]{%
        \includegraphics[width=0.32\linewidth]{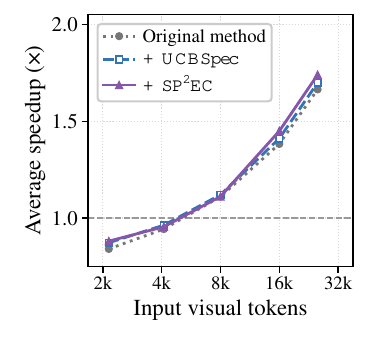}%
        \label{fig:llava16_ablation_input_size_specvlm}%
    }
    \caption{Effect of input size on average speedup with LLaVA-NeXT-Mistral-7B.
    The panels compare UCBSpec and SP$^2$EC applied to (a) \textsc{cosSim}, (b) \textsc{KnapSpec}, and (c) \textsc{SpecVLM}.
    The original methods are also shown for \textsc{KnapSpec} and \textsc{SpecVLM}.
    The horizontal axis shows the mean number of input video tokens on a base-2 logarithmic scale, with $1\mathrm{k}=1000$ tokens.
    Speedup is normalized to autoregressive decoding, indicated by the dashed horizontal line at $1\times$.}
    \label{fig:llava16_ablation_input_size}
\end{figure}

Figures~\ref{fig:ablation_input_size} and~\ref{fig:llava16_ablation_input_size} vary the visual input size.
The layer-skipping constructions, \textsc{cosSim} and \textsc{KnapSpec}, show relatively stable speedups over the tested ranges.

\textsc{SpecVLM} is substantially more sensitive to the input size because it reduces draft computation by pruning visual tokens.
Larger visual inputs provide more computation to remove, whereas short visual inputs leave limited savings to offset drafting and verification costs.
In our experiments, it requires $8000$ visual tokens to be able to exceed baseline autoregressive speed.

% Source: data/qwen25_ablation_input_size.xlsx, sheet 'Average SpeedUp', A1:C59.
% Input size is measured by the visual token budget, not the number of frames.
% All methods use the shared 1024--32768-token range; 65536-token observations are omitted.

% Source: data/llava16/llava16_ablation_input_size.xlsx.
% Points use the supplied mean input video-token counts: 2144, 4104, 8000, 16000, and 25128.

\clearpage
\subsection{Decoding Length and Resetting}
\label{app:ablation_reset}

\begin{table}[t]
\centering
\small
\setlength{\tabcolsep}{5pt}
\renewcommand{\arraystretch}{1.12}
\begin{tabular}{@{}llcc@{}}
\toprule
Model & Method
& \begin{tabular}[c]{@{}c@{}}Average SU\\Reset every 512\\speculative steps\end{tabular}
& \begin{tabular}[c]{@{}c@{}}Average SU\\No within-prompt\\reset\end{tabular} \\
\midrule
\multirow{4}{*}{Qwen2.5-VL-7B-Instruct}
& \textsc{cosSim} + \ourmethod{} & 1.26 & 1.25 \\
& \swift{} + \ourmethod{} & 1.22 & 1.22 \\
& \textsc{KnapSpec} + \ourmethod{} & 1.35 & 1.31 \\
& \textsc{SpecVLM} + \ourmethod{} & 1.56 & 1.55 \\
\midrule
\multirow{4}{*}{LLaVA-NeXT-Mistral-7B}
& \textsc{cosSim} + \ourmethod{} & 1.29 & 1.27 \\
& \swift{} + \ourmethod{} & 1.25 & 1.24 \\
& \textsc{KnapSpec} + \ourmethod{} & 1.31 & 1.28 \\
& \textsc{SpecVLM} + \ourmethod{} & 1.45 & 1.45 \\
\bottomrule
\end{tabular}
\vspace{1em}
\caption{
Average speedup (SU) with and without within-prompt resets.
The periodic-reset variant clears the online arm-pull and reward history every 512 speculative decoding steps, whereas the no-reset variant resets only when a new prompt starts.
}
\label{tab:decoding_length}
\end{table}

As generation proceeds, the growing context can change drafting and verification latency, challenging the fixed-context approximation used in Assumption~\ref{ass:affine_cost}.
To satisfy this assumption, by default, \ourmethod{} clears its online arm-pull and reward history every $512$ speculative decoding steps, which is widely adopted in piece-wise stationary bandits~\citep{cao2019nearly,hou2024almost}.
To examine the effect of periodic resetting, we compare this default with an ablation that resets only between prompts and retains the online history throughout each generation.
Resetting allows the optimizer to adapt to new contexts and latency, but also discards useful evidence and requires renewed exploration.

Table~\ref{tab:decoding_length} shows that periodic resetting improves average speedup marginally, by at most $0.04\times$ across the eight model--construction combinations.
These close results suggest that \ourmethod{} is robust to the within-prompt changes in the time-cost function induced by increasing context length.

\section{Extra Results}

This section provides some extra experimental results, on run time breakdown~(\ref{app:timing_breakdown}), comparison with EAGLE3~(\ref{app:eagle3_comparison}) and extra experiment results for \textsc{cosSim}~(\ref{legacy:cossim_qwen3}).

\subsection{Runtime Breakdown}
\label{app:timing_breakdown}

\begin{figure*}[t]
    \centering
    \includegraphics[width=\textwidth]{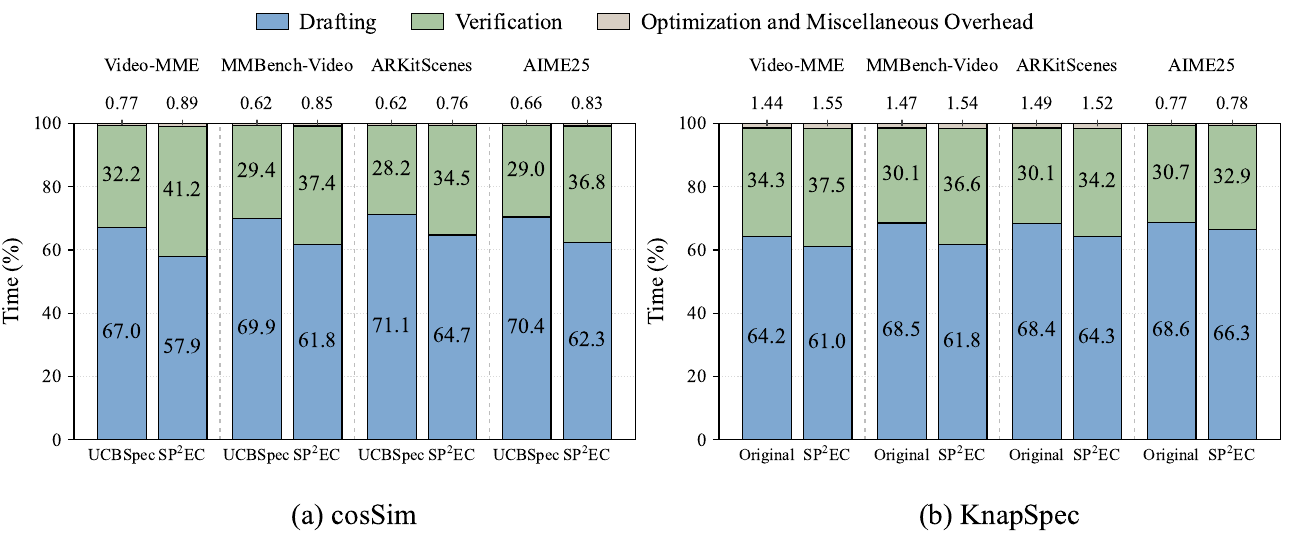}
    \caption{Runtime breakdown on Qwen2.5-VL-7B-Instruct with
    (a) \textsc{cosSim} and (b) \textsc{KnapSpec}.
    Each dataset compares the baseline (left) with \ourmethod{} (right).
    The baseline is \ucbspec{} for \textsc{cosSim} and the original method
    for \textsc{KnapSpec}. Numbers indicate percentages of recorded generation time.}
    \label{fig:timing_breakdown_cossim_knapspec}
\end{figure*}

\begin{figure*}[t]
    \centering
    \includegraphics[width=0.6\textwidth]{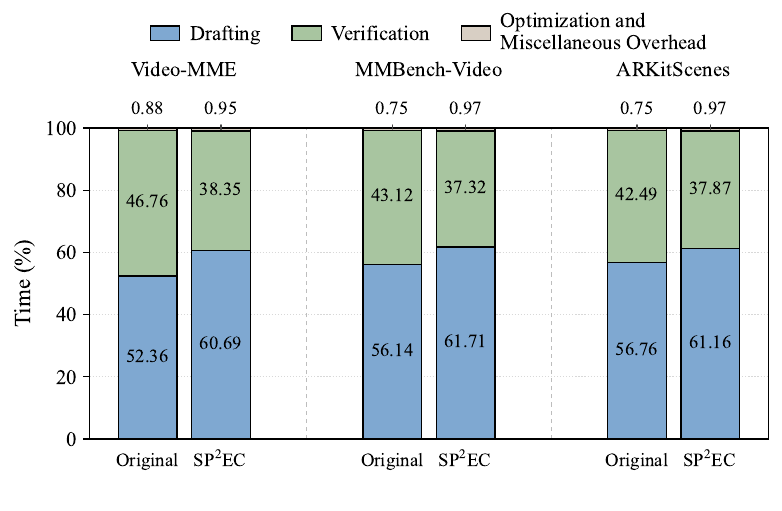}
    \caption{Runtime breakdown for \textsc{SpecVLM} on Qwen2.5-VL-7B-Instruct.
    Each dataset compares original \textsc{SpecVLM} (left) with
    \textsc{SpecVLM} augmented by \ourmethod{} (right).
    Numbers indicate percentages of recorded generation time.}
    \label{fig:timing_breakdown_specvlm}
\end{figure*}

Figures~\ref{fig:timing_breakdown_cossim_knapspec}
and~\ref{fig:timing_breakdown_specvlm} report the time spent on drafting,
verification (including prefill), and optimization and miscellaneous overhead for each method. The plotted fractions describe time allocation; they should be
interpreted together with the speedup results, since a larger fraction
does not necessarily imply a larger absolute cost.

For \textsc{cosSim} with \ourmethod{}, the recorded optimization and
miscellaneous overhead occupies only $0.76\%$--$0.89\%$ of generation time, smaller than that of  \textsc{KnapSpec}, showing that our \textsc{cosSim} is efficient to build.

Most importantly, the results show that \ourmethod{} targets overall token throughput rather than minimizing
drafting or verification cost individually.
For \textsc{cosSim} and \textsc{KnapSpec}, verification occupies a larger
fraction of runtime under \ourmethod{}, while the drafting fraction
decreases. Here, drafting-time
savings outweigh additional verification costs.
Whereas for SpecVLM, reduced verification and prefill costs contribute more substantially
to the speedup.
These different patterns reflect different sources of computational
savings across draft-model construction methods.

\subsection{Comparison with \eaglethree{}}
\label{app:eagle3_comparison}

\begin{table}[t]
\centering
\small
\setlength{\tabcolsep}{12pt}
\renewcommand{\arraystretch}{1.12}
\begin{tabular}{@{}lccc@{}}
\toprule
Method & Training-free & MAT $\uparrow$ & SU $\uparrow$ \\
\midrule
\textsc{cosSim} + \ourmethod{} & Yes & 4.5 & \textbf{1.31}$\times$ \\
\textsc{KnapSpec} + \ourmethod{} & Yes & 4.6 & 1.27$\times$ \\
\eaglethree{} & No & 1.9 & 1.20$\times$ \\
\bottomrule
\end{tabular}
\caption{Comparison with \eaglethree{} on MVBench. MAT denotes mean accepted tokens, including one verification bonus token; SU denotes speedup over vanilla autoregressive decoding. Training-free indicates that no auxiliary draft-model training is required.}
\label{tab:eagle3_comparison}
\end{table}

We compare \textsc{cosSim} and \textsc{KnapSpec} with \ourmethod{} against \eaglethree{}~\citep{li2025eagle3} on MVBench~\citep{mvbench2024}, with Qwen3-VL-4B-Instruct~\citep{qwen3vltechnicalreport}.
For \eaglethree{}, we use AngelSlim's publicly released \texttt{Qwen3-VL-4B-Instruct\_eagle3} draft-model checkpoint~\citep{AngelSlim2025}.

Compared with \eaglethree{}, \ourmethod{} is training-free: it does not require training or storing an auxiliary draft model, and it avoids dependence on the alignment between the draft model's training distribution and the test-time prompt distribution.
Empirically, results for \eaglethree{} also suggest that small draft models can be weak when the prompt contains mixed visual and textual context.
In vision-language tasks, prefilling and verification occupy a larger fraction of each speculative round, so the small number of valid draft tokens produced by a weak drafter cannot amortize these extra costs.
In contrast, \ourmethod{} obtains the draft model directly by skipping layers of the target model, which keeps the drafter better aligned with the target under visual and contextual distribution shifts.
As shown in Table~\ref{tab:eagle3_comparison}, \textsc{cosSim} and \textsc{KnapSpec} with \ourmethod{} achieve MATs of $4.5$ and $4.6$ and speedups of $1.31\times$ and $1.27\times$, respectively, compared with a MAT of $1.9$ and a speedup of $1.20\times$ for \eaglethree{}.

\subsection{Extra Experiment Results for \textsc{cosSim} + \ourmethod{}} \label{legacy:cossim_qwen3}

Tables~\ref{tab:cossim_qwen3vl4b} and~\ref{tab:cossim_llava15} report additional results with \textsc{cosSim} on Qwen3-VL-4B-Instruct and LLaVA-1.5-7B.
\ourmethod{} outperforms \ucbspec{} in speedups on all four benchmarks for both models, attaining average speedups of $1.19\times$ and $1.18\times$, respectively.

\begin{table*}[t]
\caption{Additional results on Qwen3-VL-4B-Instruct.
MAT denotes mean accepted tokens, and SU denotes speedup over vanilla
autoregressive decoding. Bold marks the higher speedup.}
\label{tab:cossim_qwen3vl4b}
\centering
\small
\setlength{\tabcolsep}{3.5pt}
\renewcommand{\arraystretch}{1.12}
\resizebox{\textwidth}{!}{
\begin{tabular}{l*{10}{c}}
\toprule
\multirow{2}{*}{Method}
& \multicolumn{2}{c}{Video-MME}
& \multicolumn{2}{c}{MMBench-Video}
& \multicolumn{2}{c}{ARKitScenes}
& \multicolumn{2}{c}{WritingBench}
& \multicolumn{2}{c}{Average} \\
\cmidrule(lr){2-3}
\cmidrule(lr){4-5}
\cmidrule(lr){6-7}
\cmidrule(lr){8-9}
\cmidrule(lr){10-11}
& MAT $\uparrow$ & SU $\uparrow$
& MAT $\uparrow$ & SU $\uparrow$
& MAT $\uparrow$ & SU $\uparrow$
& MAT $\uparrow$ & SU $\uparrow$
& MAT $\uparrow$ & SU $\uparrow$ \\
\midrule
\textsc{cosSim} + \ucbspec{}
& 3.98 & 1.13$\times$
& 3.64 & 1.10$\times$
& 3.50 & 1.09$\times$
& 3.65 & 1.08$\times$
& 3.69 & 1.10$\times$ \\
\rowcolor{sp2ecrow}
\textsc{cosSim} + \ourmethod{}
& 3.88 & \textbf{1.22}$\times$
& 3.57 & \textbf{1.21}$\times$
& 3.33 & \textbf{1.16}$\times$
& 3.69 & \textbf{1.17}$\times$
& 3.62 & \textbf{1.19}$\times$ \\
\bottomrule
\end{tabular}
}
\end{table*}

\begin{table*}[t]
\caption{Additional results on LLaVA-1.5-7B.
MAT denotes mean accepted tokens, and SU denotes speedup over vanilla
autoregressive decoding. Bold marks the higher speedup.}
\label{tab:cossim_llava15}
\centering
\small
\setlength{\tabcolsep}{3.5pt}
\renewcommand{\arraystretch}{1.12}
\resizebox{\textwidth}{!}{
\begin{tabular}{l*{10}{c}}
\toprule
\multirow{2}{*}{Method}
& \multicolumn{2}{c}{Video-MME}
& \multicolumn{2}{c}{MMBench-Video}
& \multicolumn{2}{c}{ARKitScenes}
& \multicolumn{2}{c}{WritingBench}
& \multicolumn{2}{c}{Average} \\
\cmidrule(lr){2-3}
\cmidrule(lr){4-5}
\cmidrule(lr){6-7}
\cmidrule(lr){8-9}
\cmidrule(lr){10-11}
& MAT $\uparrow$ & SU $\uparrow$
& MAT $\uparrow$ & SU $\uparrow$
& MAT $\uparrow$ & SU $\uparrow$
& MAT $\uparrow$ & SU $\uparrow$
& MAT $\uparrow$ & SU $\uparrow$ \\
\midrule
\textsc{cosSim} + \ucbspec{}
& 4.23 & 1.16$\times$
& 3.35 & 1.10$\times$
& 3.40 & 1.16$\times$
& 3.64 & 1.06$\times$
& 3.66 & 1.12$\times$ \\
\rowcolor{sp2ecrow}
\textsc{cosSim} + \ourmethod{}
& 4.19 & \textbf{1.24}$\times$
& 3.36 & \textbf{1.17}$\times$
& 3.43 & \textbf{1.21}$\times$
& 3.65 & \textbf{1.10}$\times$
& 3.66 & \textbf{1.18}$\times$ \\
\bottomrule
\end{tabular}
}
\end{table*}

%%%%%%%%%%%%%%%%%%%%%%%%%%%%%%%%%%%%%%%%%%%%%%%%%%%%%%%%%%%%

\newpage
\section*{NeurIPS Paper Checklist}

\begin{enumerate}

\item {\bf Claims}
    \item[] Question: Do the main claims made in the abstract and introduction accurately reflect the paper's contributions and scope?
    \item[] Answer: \answerYes{} % Replace by \answerYes{}, \answerNo{}, or \answerNA{}.
    \item[] Justification: The claims made in the abstract and introduction are consistent with the contributions presented in the paper. They accurately reflect the theoretical and empirical findings, remain within the demonstrated scope, and acknowledge relevant assumptions and limitations.

\item {\bf Limitations}
    \item[] Question: Does the paper discuss the limitations of the work performed by the authors?
    \item[] Answer: \answerYes{} % Replace by \answerYes{}, \answerNo{}, or \answerNA{}.
    \item[] Justification: We provide a discussion of limitations in the conclusion.

\item {\bf Theory assumptions and proofs}
    \item[] Question: For each theoretical result, does the paper provide the full set of assumptions and a complete (and correct) proof?
    \item[] Answer: \answerYes{} % Replace by \answerYes{}, \answerNo{}, or \answerNA{}.
    \item[] Justification: Assumptions and correct proofs are provided, either in the main text or the appendix.

    \item {\bf Experimental result reproducibility}
    \item[] Question: Does the paper fully disclose all the information needed to reproduce the main experimental results of the paper to the extent that it affects the main claims and/or conclusions of the paper (regardless of whether the code and data are provided or not)?
    \item[] Answer: \answerYes{} % Replace by \answerYes{}, \answerNo{}, or \answerNA{}.
    \item[] Justification: The paper provides complete details on architecture, test setup, and experimental configurations, enabling reproduction of the main results and supporting the core claims.

\item {\bf Open access to data and code}
    \item[] Question: Does the paper provide open access to the data and code, with sufficient instructions to faithfully reproduce the main experimental results, as described in supplemental material?
    \item[] Answer: \answerYes{} % Replace by \answerYes{}, \answerNo{}, or \answerNA{}.
    \item[] Justification: The paper provides detailed descriptions of each experiment, and a link to the code is included in the experiment section.

\item {\bf Experimental setting/details}
    \item[] Question: Does the paper specify all the training and test details (e.g., data splits, hyperparameters, how they were chosen, type of optimizer) necessary to understand the results?
    \item[] Answer: \answerYes{} % Replace by \answerYes{}, \answerNo{}, or \answerNA{}.
    \item[] Justification: The paper provides test and evaluation details, either in the main text or appendix.

\item {\bf Experiment statistical significance}
    \item[] Question: Does the paper report error bars suitably and correctly defined or other appropriate information about the statistical significance of the experiments?
    \item[] Answer: \answerYes{} % Replace by \answerYes{}, \answerNo{}, or \answerNA{}.
    \item[] Justification: We reported error bars.

\item {\bf Experiments compute resources}
    \item[] Question: For each experiment, does the paper provide sufficient information on the computer resources (type of compute workers, memory, time of execution) needed to reproduce the experiments?
    \item[] Answer: \answerYes{} % Replace by \answerYes{}, \answerNo{}, or \answerNA{}.
    \item[] Justification: Mentioned in appendix.
    
\item {\bf Code of ethics}
    \item[] Question: Does the research conducted in the paper conform, in every respect, with the NeurIPS Code of Ethics \url{https://neurips.cc/public/EthicsGuidelines}?
    \item[] Answer: \answerYes{} % Replace by \answerYes{}, \answerNo{}, or \answerNA{}.
    \item[] Justification: The research complies with the NeurIPS Code of Ethics, with no involvement of sensitive data, human subjects, or foreseeable risks. Ethical standards are upheld throughout.

\item {\bf Broader impacts}
    \item[] Question: Does the paper discuss both potential positive societal impacts and negative societal impacts of the work performed?
    \item[] Answer: \answerNA{} % Replace by \answerYes{}, \answerNo{}, or \answerNA{}.
    \item[] Justification: The paper focuses on theoretical analysis and verification of proposed method, and the authors believe it does not pose any societal impact, either positive or negative.
    
\item {\bf Safeguards}
    \item[] Question: Does the paper describe safeguards that have been put in place for responsible release of data or models that have a high risk for misuse (e.g., pre-trained language models, image generators, or scraped datasets)?
    \item[] Answer: \answerNA{} % Replace by \answerYes{}, \answerNo{}, or \answerNA{}.
    \item[] Justification: No such data or models are released, and the described methods do not pose risks of malicious use.

\item {\bf Licenses for existing assets}
    \item[] Question: Are the creators or original owners of assets (e.g., code, data, models), used in the paper, properly credited and are the license and terms of use explicitly mentioned and properly respected?
    \item[] Answer: \answerYes{} % Replace by \answerYes{}, \answerNo{}, or \answerNA{}.
    \item[] Justification: We use standard and open sourced libraries  under permissive licenses.

\item {\bf New assets}
    \item[] Question: Are new assets introduced in the paper well documented and is the documentation provided alongside the assets?
    \item[] Answer: \answerYes{} % Replace by \answerYes{}, \answerNo{}, or \answerNA{}.
    \item[] Justification: Link to new code is provided.

\item {\bf Crowdsourcing and research with human subjects}
    \item[] Question: For crowdsourcing experiments and research with human subjects, does the paper include the full text of instructions given to participants and screenshots, if applicable, as well as details about compensation (if any)? 
    \item[] Answer: \answerNA{} % Replace by \answerYes{}, \answerNo{}, or \answerNA{}.
    \item[] Justification: No crowdsourcing or human subjects are involved in this paper.

\item {\bf Institutional review board (IRB) approvals or equivalent for research with human subjects}
    \item[] Question: Does the paper describe potential risks incurred by study participants, whether such risks were disclosed to the subjects, and whether Institutional Review Board (IRB) approvals (or an equivalent approval/review based on the requirements of your country or institution) were obtained?
    \item[] Answer: \answerNA{} % Replace by \answerYes{}, \answerNo{}, or \answerNA{}.
    \item[] Justification: No human subjects are involved in this paper.

\item {\bf Declaration of LLM usage}
    \item[] Question: Does the paper describe the usage of LLMs if it is an important, original, or non-standard component of the core methods in this research? Note that if the LLM is used only for writing, editing, or formatting purposes and does \emph{not} impact the core methodology, scientific rigor, or originality of the research, declaration is not required.
    %this research? 
    \item[] Answer: \answerNA{} % Replace by \answerYes{}, \answerNo{}, or \answerNA{}.
    \item[] Justification: The LLM is only used for grammar checking, editing and formatting purposes.

\end{enumerate}

\end{document}